\documentclass[times,review,10pt]{elsarticle}

\usepackage{amssymb}
\usepackage{amsmath}
\usepackage{booktabs}
\usepackage{multirow}
\usepackage{hyperref}
\usepackage{makecell}
\usepackage[table]{xcolor}
\usepackage{colortbl}
\usepackage{pifont}
\usepackage{graphicx}
\usepackage{algorithmic}
\usepackage{algorithm}

\newcommand{\xmark}{{\color[HTML]{C00000}\ding{55}}}   
\newcommand{\cmark}{{\color[HTML]{70AD47}\ding{51}}} 
\definecolor{mygreen}{HTML}{70AD47}
\definecolor{myred}{HTML}{C00000}

\newcommand{\mathsmallish}[1]{%
  \mathchoice
    {\scalebox{0.9}{$\displaystyle #1$}}
    {\scalebox{0.9}{$\textstyle #1$}}
    {\scalebox{0.9}{$\scriptstyle #1$}}
    {\scalebox{0.9}{$\scriptscriptstyle #1$}}
}
\newcommand{\mathtiny}[1]{%
  \mathchoice
    {\scalebox{0.75}{$\displaystyle #1$}}
    {\scalebox{0.75}{$\textstyle #1$}}
    {\scalebox{0.75}{$\scriptstyle #1$}}
    {\scalebox{0.75}{$\scriptscriptstyle #1$}}
}

\newcommand{\mathttt}[1]{%
  \mathchoice
    {\scalebox{0.6}{$\displaystyle #1$}}
    {\scalebox{0.6}{$\textstyle #1$}}
    {\scalebox{0.6}{$\scriptstyle #1$}}
    {\scalebox{0.6}{$\scriptscriptstyle #1$}}
}

\DeclareMathOperator*{\argmax}{argmax}

\journal{Pattern Recognition}

\begin{document}

\begin{frontmatter}


\title{UniADC: A Unified Framework for Anomaly Detection and Classification}

\author[1]{Ximiao Zhang}
\ead{2024010482@bupt.cn}

\author[2]{Min Xu}
\ead{xumin@cnu.edu.cn}

\author[1]{Zheng Zhang}
\ead{zhangzheng@bupt.edu.cn}

\author[3,4]{Yap-Peng Tan}
\ead{yp.t@vinuni.edu.vn}

\author[1]{Xiuzhuang Zhou\corref{cor1}}
\ead{xiuzhuang.zhou@bupt.edu.cn}

\affiliation[1]{organization={School of Intelligent Engineering and Automation},
             addressline={Beijing University of Posts and Telecommunications},
             city={Beijing},
             postcode={100876},
             country={China}}
\affiliation[2]{organization={College of Information and Engineering},
             addressline={Capital Normal University},
             city={Beijing},
             postcode={100048},
             country={China}}
\affiliation[3]{addressline={VinUniversity},
             city={Hanoi},
             postcode={12426},
             country={Vietnam}}
\affiliation[4]{addressline={Nanyang Technological University},
             city={Singapore},
             postcode={639798},
             country={Singapore}}   
\cortext[cor1]{Corresponding author.}

\begin{abstract}

In this paper, we introduce a novel task termed unified anomaly detection and classification, which aims to simultaneously detect anomalous regions in images and identify their specific categories. Existing methods typically treat anomaly detection and classification as separate tasks, thereby neglecting their inherent correlations and limiting information sharing, which results in suboptimal performance. To address this, we propose UniADC, a model designed to effectively perform both tasks with only a few or even no anomaly images. Specifically, UniADC consists of two key components: a training-free Controllable Inpainting Network and an Implicit-Normal Discriminator. The inpainting network can synthesize anomaly images of specific categories by repainting normal regions guided by anomaly priors, and can also repaint few-shot anomaly samples to augment the available anomaly data. The implicit-normal discriminator addresses the severe challenge of the imbalance between normal and anomalous pixel distributions by implicitly modeling the normal state, achieving precise anomaly detection and classification by aligning fine-grained image features with anomaly-category embeddings. We conduct extensive experiments on four anomaly detection and classification datasets, including MVTec-FS, MTD, WFDD and Real-IAD, and the results demonstrate that UniADC consistently outperforms existing methods in anomaly detection, localization, and classification. The code is available at \url{https://github.com/cnulab/UniADC}.

\end{abstract}


\begin{keyword}
Anomaly detection \sep Anomaly classification \sep Anomaly synthesis 
\end{keyword}

\end{frontmatter}

\section{Introduction}

Image anomaly detection aims to train models to detect and localize anomalous regions within images. It has numerous practical applications and has attracted growing research interest in recent years. Anomaly classification further categorizes detected anomalies, which aids in assessing anomaly severity and facilitating root-cause analysis. For instance, severe anomalies like scratches and breakage significantly compromise product quality, necessitating immediate production halts for inspection. Conversely, minor anomalies like stains or foreign matter have a negligible impact and are typically not classified as genuine defects. Existing studies \cite{lyu2025mvrec, huang2025anomalyncd} typically treat anomaly classification as a downstream task of anomaly detection, in which the detected anomalous regions are cropped into sub-images and then fed into a separate classification model. This two-stage pipeline ignores the inherent connection between anomaly detection and classification, and hinders information sharing across the two tasks. Moreover, missed and over detections during the detection stage, as well as the difficulty in selecting appropriate anomaly score thresholds, pose a series of challenges for subsequent classification. As a result, such methods often suffer from high computational complexity and suboptimal performance, limiting their practicality in real-world scenarios.

\begin{figure}[t]
  \centering
   \includegraphics[width=\linewidth]{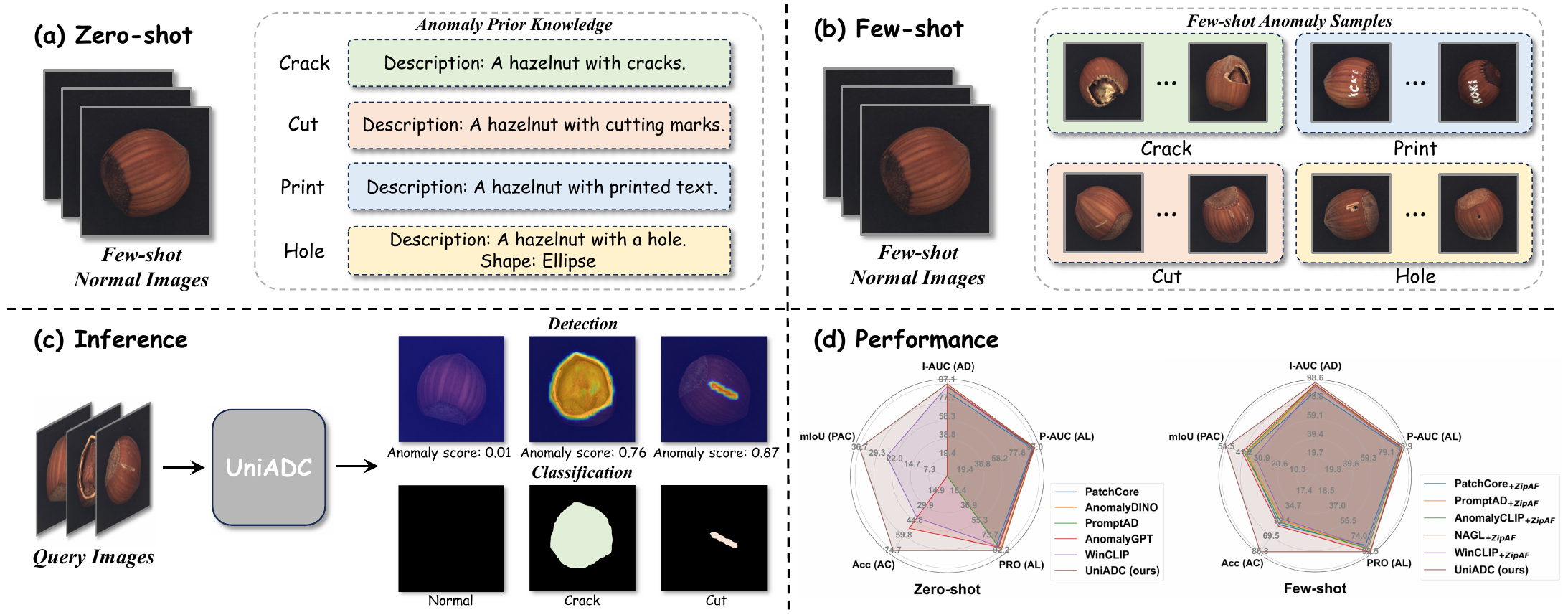}
   \caption{Task settings for unified anomaly detection and classification. \textbf{(a) Zero-shot}: training with a few normal samples and prior knowledge for each anomaly category. \textbf{(b) Few-shot}: training with a few normal samples and few-shot anomaly samples per category. \textbf{(c) Inference}: UniADC predicts anomaly scores and categories for query images. \textbf{(d) Performance}: UniADC demonstrates superior performance over existing methods in Anomaly Detection (AD), Anomaly Localization (AL), Anomaly Classification (AC), and Pixel-level Anomaly Classification (PAC).}
   \label{fig:fig1}
\end{figure}

In this paper, we integrate anomaly detection and classification tasks and address them with a unified model. We propose zero-shot and few-shot settings for unified anomaly detection and classification, both extending the widely studied few-shot anomaly detection task \cite{jeong2023winclip, li2024promptad, gu2024anomalygpt}. Few-shot anomaly detection provides only a small number of normal samples for model training, thereby reducing data collection costs, and has been widely applied in industrial \cite{fan2026towards, zhang2025dc} and medical \cite{zhang2024mediclip} fields. This setting is highly aligned with anomaly classification tasks in which available data is limited. Building upon this, the zero-shot anomaly detection and classification task does not provide additional anomaly samples for model training, but instead offers prior knowledge about anomaly categories, such as textual descriptions or shape information. This setting simulates real-world scenarios where collecting anomalous data is challenging. Differently, the few-shot anomaly detection and classification task provides a small number of anomaly samples for each category during model training, simulating practical scenarios where anomalous data is limited. The training and inference workflow for anomaly detection and classification is shown in Fig. \ref{fig:fig1}(a--c).

To address the above tasks, we propose UniADC, a unified model for anomaly detection and classification, which tackles the challenge of scarce anomalous data through controllable image inpainting. It consists of two key components: a training-free controllable inpainting network and an implicit-normal discriminator. The controllable inpainting network consists of a pre-trained latent diffusion model \cite{rombach2022high} and an inpainting control network \cite{zhang2023adding}, enabling controllable inpainting guided by either anomaly priors or anomaly samples. Specifically, anomaly prior-guided controllable inpainting synthesizes category-specific anomalous images by repainting normal regions according to the provided prior knowledge of anomalies, whereas anomaly sample-guided controllable inpainting refines and repaints few-shot anomalous samples to enhance data diversity. In addition, we propose a category consistency selection strategy to filter synthetic anomaly samples that are highly consistent with the target category. The implicit-normal discriminator is trained on these synthetic anomaly samples and mitigates performance degradation caused by pixel-level distribution imbalance by disentangling the normal state from various anomaly concepts, thereby enabling robust anomaly detection and classification. Through the aforementioned innovative designs, UniADC comprehensively outperforms existing methods in anomaly detection, localization, and classification, as shown in Fig. \ref{fig:fig1}(d), particularly strengthening the reliability of defect categorization. Our main contributions are summarized as follows:

\begin{itemize}
\item We introduce the task of unified anomaly detection and classification, which has broad application prospects yet remains underexplored. To this end, we propose UniADC, which enables accurate anomaly detection and classification under both zero-shot and few-shot settings.
\item We propose a training-free controllable inpainting network that can generate category-specific anomaly samples conditioned on either anomaly prior or few-shot anomaly images. This enables broad applicability across various anomaly detection and classification tasks, serving as an effective alternative to existing anomaly synthesis methods.
\item We propose an implicit-normal discriminator that mitigates performance degradation arising from the long-tail effect of anomaly data by implicitly modeling normal concepts, enabling it to effectively align fine-grained image features with anomaly-category embeddings for unified anomaly detection and classification. 
\item Extensive experiments on the MVTec-FS \cite{lyu2025mvrec}, MTD \cite{huang2020surface}, WFDD \cite{chen2024unified}, and Real-IAD \cite{wang2024real} datasets demonstrate the effectiveness of UniADC and highlight its potential for real-world anomaly detection and classification applications.
\end{itemize}

\section{Related Work}

\subsection{Anomaly Detection}

Image anomaly detection has gained widespread attention in recent years, leading to extensive research. Due to the scarcity of anomaly samples in real-world scenarios, most methods \cite{roth2022towards, zhang2025towards,zhu2026real,li2026prototype} follow an unsupervised paradigm, training solely on normal data and identifying anomalies that deviate from normal patterns during inference. However, in practical production scenarios, anomaly information is not entirely out of reach. Consequently, recent studies have explored integrating anomaly priors or limited anomaly samples into the detection pipeline to improve reliability. For example, PromptAD \cite{li2024promptad} employs textual descriptions of each anomaly category to improve detection accuracy. AnomalyGPT \cite{gu2024anomalygpt} predefines the expected appearance of query images and potential anomaly types to enhance the reasoning capability of the anomaly question answering model. AnoGen \cite{gui2024few} integrates size priors of different anomaly types into the anomaly synthesis pipeline to ensure class consistency of the synthesized anomaly samples. BGAD \cite{yao2023explicit} and NAGL \cite{wang2025normalabnormal} leverage a small number of anomaly samples to refine the model's decision boundary, thereby bolstering its discriminative capability for more accurate detection. Motivated by these advances and real-world requirements, we systematically integrate diverse anomaly priors and limited anomaly samples into a unified anomaly detection and classification pipeline, and propose UniADC, a general framework that functionally extends existing methods.

\subsection{Anomaly Classification}

Anomaly classification aims to identify the categories of detected anomalies, which remains highly challenging due to the scarcity of anomaly samples. Despite its broad range of applications, this field has not yet been extensively explored. ZipAF \cite{lyu2025mvrec} employs the AlphaCLIP model \cite{sun2024alpha} to extract region-contextual anomaly features and adopts a zero-initialized projection to align query features with cached anomaly features for anomaly classification. However, it does not incorporate the anomaly detection process, which may result in degraded performance in practical scenarios. HypDFS \cite{li2026hyperbolic} follows the setting of ZipAF \cite{lyu2025mvrec} and models defect categories in hyperbolic space. AnomalyNCD \cite{huang2025anomalyncd} integrates novel class discovery methods into the anomaly detection pipeline for anomaly detection and clustering. MultiADS \cite{sadikaj2025multiads} focuses on zero-shot multi-type anomaly detection, training the model to learn cross-domain anomaly priors for defect classification. However, due to the lack of targeted training, this method struggles to identify domain-specific anomaly types, such as highly similar defects within a specific product \cite{song2013noise, huang2020surface, zhan2022fabric}. A recent work, PG-SFD \cite{luo2026dual}, shares a motivation similar to ours by jointly considering anomaly detection and classification rather than treating them as two independent stages. However, PG-SFD \cite{luo2026dual} mainly focuses on semi-supervised scenarios where anomaly annotations are available and adopts a multi-stage prototype-guided reconstruction paradigm, enabling the prediction of image-level anomaly categories. In contrast, UniADC targets low-data regimes with few or even no anomaly images and addresses this problem through controllable anomaly inpainting and implicit-normal discrimination, enabling precise pixel-level anomaly classification, which makes its technical route and practical focus fundamentally different.

\subsection{Anomaly Synthesis}

Existing anomaly synthesis methods can generally be classified into zero-shot and few-shot approaches, depending on the availability of real anomaly data during the synthesis process. Zero-shot approaches generate anomaly samples relying on predefined data augmentation rules \cite{li2021cutpaste, schluter2022natural,wang2025mmfnet}, noise injection \cite{zhang2024realnet}, or textual descriptions \cite{sun2025unseen, jiang2025anomagic}, without access to real anomaly samples. However, the synthesized anomalies may exhibit a distribution shift from real-world cases, limiting their effectiveness. In contrast, few-shot approaches \cite{duan2023few, hu2024anomalydiffusion, jin2025dual} aim to enhance data diversity by augmenting a limited number of real anomaly samples. 
Despite significant progress in anomaly synthesis techniques, existing research remains primarily focused on anomaly detection. The category controllability and accuracy of synthesized anomaly samples still require further validation to ensure their feasibility for more challenging anomaly classification tasks.

\section{Method}

\begin{figure*}[t]
\centering
\includegraphics[width=\linewidth]{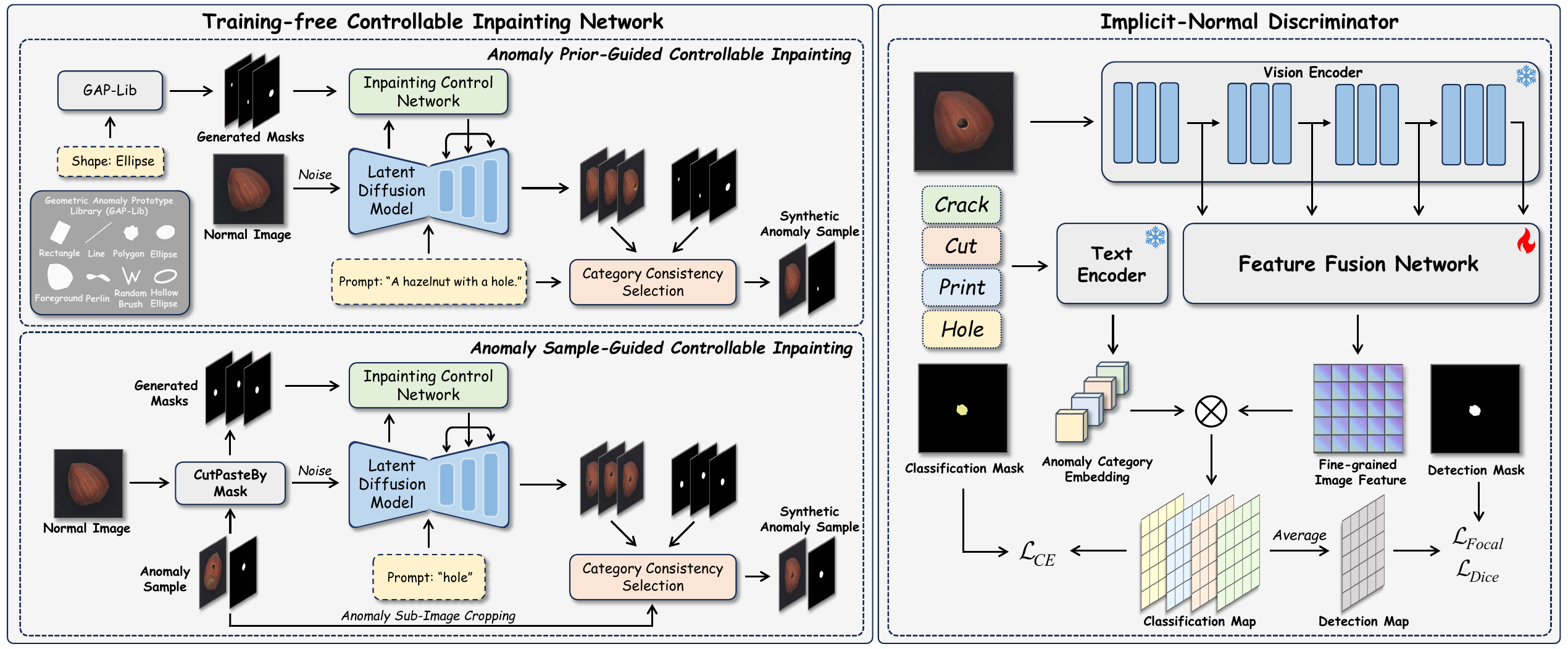}
\caption{Overview of the proposed UniADC pipeline, which consists of a training-free controllable inpainting network and an implicit-normal discriminator. The controllable inpainting network supports two modes: anomaly prior-guided controllable inpainting and anomaly sample-guided controllable inpainting, enabling the generation of category-specific anomaly samples under different settings. The implicit-normal discriminator aligns fine-grained image features with anomaly category embeddings for accurate anomaly detection and classification.}
\label{fig:fig2}
\end{figure*}

In this section, we present our proposed anomaly detection and classification model, UniADC, with its overall pipeline illustrated in Fig. \ref{fig:fig2}. We first provide the problem definition, then describe the main functionalities and implementation details of the controllable inpainting network, including anomaly prior-guided controllable inpainting and anomaly sample-guided controllable inpainting. Finally, we introduce the implicit-normal discriminator of UniADC.

\subsection{Problem Definition}

In few-shot anomaly detection, the model is provided with a support set of normal samples $\mathcal{D}^{n}=\{ X_{1},X_{2},\cdots,X_{K_n} \}$, where $K_n$ denotes the number of normal samples per image class. Given a query image, the model predicts an image-level anomaly detection score $I_{d} \in [0,1]$ and a pixel-level anomaly map $S_{d} \in [0,1]^{H \times W}$, where $H$ and $W$ denote the image height and width, respectively. The zero-shot and few-shot anomaly detection and classification tasks respectively provide additional anomaly prior knowledge and an anomaly support set for model training. Specifically, the anomaly support set is represented as $ \mathcal{D}^{a}= \bigcup_{y \in \{1,2,\cdots,Y\}}{\{ (X_{1}^{y}, M_{1}^{y}), \:(X_{2}^{y},M_{2}^{y}),\cdots,\:(X_{K_a}^{y},M_{K_a}^{y})\}}$, where $Y$ denotes the number of anomaly categories, $K_a$ is the number of samples per anomaly category, and $M$ denotes the binary anomaly mask. The model is further trained to predict both an image-level classification result $I_{c} \in \{0,1,\cdots,Y\}$ and a pixel-level classification result $S_{c} \in \{0,1,\cdots,Y\}^{H \times W}$, where 0 indicates the normal class, and 1 to $Y$ represent the $Y$ anomaly categories.

It is worth noting that the task of unified anomaly detection and classification significantly extends the functional scope of traditional image anomaly detection. Given that the objective has evolved from simple "anomaly discovery" to "anomaly identification and categorization", it is natural for this task to adopt distinct task formulations compared to standard anomaly detection. In standard settings, few-shot typically implies training solely on few-shot normal data. However, in a classification context, we introduce anomaly prior knowledge for zero-shot learning and a few anomaly samples for few-shot learning to define category boundaries. These adjustments represent necessary evolutions driven by increased task complexity, aiming to better simulate real-world requirements for fine-grained defect recognition.

\subsection{Anomaly Prior-Guided Controllable Inpainting}

\begin{figure}[t]
\centering
\includegraphics[width=0.8\linewidth]{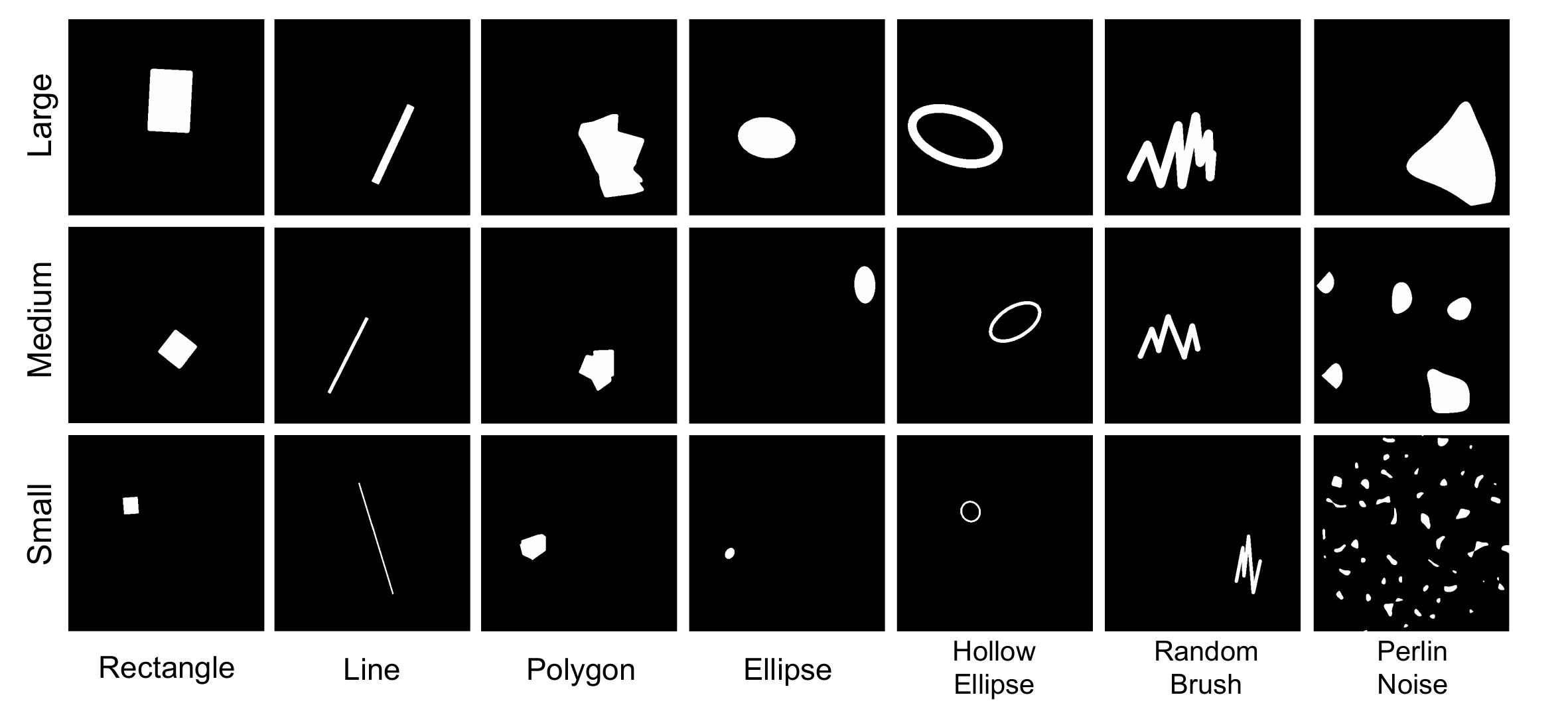}
\caption{Examples of anomaly masks generated from the GAP-Lib.}
\label{fig:fig3}
\vspace{-0.4cm}
\end{figure}

This module synthesizes category-specific anomaly samples based on provided anomaly priors, making it suitable for zero-shot anomaly detection and classification tasks. The anomaly prior includes the shape, size, and textual description of each anomaly category. Shape and size are used to generate the anomaly mask, while the textual description serves as a prompt to guide the synthesis of the desired anomalous appearance. Specifically, to map various anomaly shapes and sizes into explicit spatial constraints, we constructed a Geometric Anomaly Prototype Library (GAP-Lib). It comprises eight types of candidate masks, including \textit{Rectangle, Line, Polygon, Ellipse, Hollow Ellipse, Random Brush, Perlin Noise, and Foreground Mask}. The first seven types are used to simulate various local anomalous regions, with each type defined in three sizes: \textit{Large, Medium, and Small}, as illustrated in Fig. \ref{fig:fig3}. Each mask size corresponds to a specific size range. \textit{Foreground Mask} uses the object's foreground as a mask to synthesize global anomalies. We use specific mask types for mask generation based on the given anomaly shape and size. For example, we specify the shape of the anomaly category "Hole" as Ellipse, which will generate elliptical masks of arbitrary sizes. When both shape and size priors are unavailable for an anomaly category, all mask types are used for mask generation. 

For a given normal image $X$ and prior knowledge from anomaly category $y$, we first generate an anomaly mask $M^{y}$ utilizing the proposed GAP-Lib. Then, the latent diffusion model encodes the image $X$ and the text prompt $p^y$ into latent variables $z=\mathbb{E}_{v}(X)$ and $z^p=\mathbb{E}_{t}(p^y)$, where $\mathbb{E}_v(\cdot)$ and $\mathbb{E}_t(\cdot)$ denote the vision and text encoders, respectively. We define the number of forward diffusion steps as $T'=T \cdot \gamma$, where $T$ denotes the total number of diffusion steps in the original latent diffusion model, and $\gamma \in (0,1]$ controls the noise strength. A larger $\gamma$ injects more noise into the latent variable, resulting in synthesized anomalies that deviate further from the original image distribution and exhibit more visually prominent anomalous patterns. To ensure spatial consistency between synthesized anomalies and the anomaly mask, we use an inpainting control network to precisely control the locations of the synthesized anomalies. Specifically, at each time step $t \in \{1,2,\cdots,T'\}$, the inpainting control network $\phi$ predicts the conditioning variable $z^m=\phi(z_t,t,z,M^y)$ of noisy latent $z_t$ for controlling the denoising process:
\begin{equation}
\label{eq:equ1}
z_{t-1}= \Psi(z_t,t,z^p,z^m),
\end{equation}
where $\Psi$ denotes the latent diffusion model, and the conditioning variables $z^p$ and $z^m$ respectively control the semantics and spatial positions of the synthesized anomalies. The inpainting control network ensures that the denoised latent representation $z_0$ remains consistent with the original latent $z$ outside the masked region, while repainting only the masked region to align with the prompt embedding $z^p$. Finally, the denoised latent representation $z_0$ is decoded by decoder $\mathbb{D}(\cdot)$ into the anomaly image $X^y=\mathbb{D}(z_0)$. In our implementation, we adopt Stable Diffusion v1.5 \cite{rombach2022high} as the latent diffusion model and integrate BrushNet \cite{ju2024brushnet} as a plug-and-play control network for inpainting. This configuration enables UniADC to achieve highly accurate control over anomaly placement and image-mask consistency, bypassing the need for costly post-processing or mask refinement steps common in other anomaly synthesis methods \cite{jin2025dual, sun2025unseen, jiang2025anomagic}.

However, the above synthesis process does not ensure the semantic alignment between anomaly masks and their corresponding textual descriptions, which may lead to inconsistencies between the synthesized anomaly samples and their intended categories. For example, in the case of \textit{"a broken transistor lead"}, the synthesized anomaly should be located in the lead region, rather than randomly elsewhere. To this end, we propose a Category Consistency Selection (CCS) strategy to further filter synthetic anomalous images with high category consistency, thereby eliminating noisy samples introduced by anomaly synthesis. Given an anomaly category $y$, we first generate a mini-batch of synthesized samples using diverse anomaly masks, denoted as $\mathcal{S}^y=\{(X_b^y, M_b^y)\}_{b=1}^{B}$, where $B$ is the mini-batch size. We then adopt the AlphaCLIP model \cite{sun2024alpha} to evaluate the semantic consistency between the synthesized image-mask pairs and the textual description of the anomaly category. For the synthetic anomaly sample $(X^y_b, M^y_b)$, we calculate its category matching score as:
\begin{equation}
\label{eq:equ2}
\mathcal{P}_{b} = \frac{\exp (\langle \psi_v(X^y_b, M^y_b) , \psi_t(p^y) \rangle)}{\sum_{ \mathsmallish{\mathbf{y} \in \{1,2,\cdots,Y\}} }\exp(  \langle \psi_v(X^y_b, M^y_b) , \psi_t(p^\mathbf{y}) \rangle)},
\end{equation}
where $\psi_v$ and $\psi_t$ denote the vision and text encoders of AlphaCLIP, respectively, and $\langle \cdot, \cdot \rangle$ denotes the inner product. The AlphaCLIP model aligns the image region features guided by the masks with the text embeddings. If the synthetic anomalous region fails to match the corresponding anomaly description, such as being placed in the wrong location, the matching score decreases. In the end, we select the synthetic anomaly sample with the highest matching score in the mini-batch $\mathcal{S}^y$ for subsequent discriminator training.

\begin{table}[t]
  \centering
   \renewcommand\arraystretch{0.78}
  \caption{Comparison between UniADC and alternative anomaly synthesis methods. Note: "training-free" specifically characterizes the anomaly synthesis stage, independent of the subsequent training of the anomaly detection model.}
   \resizebox{0.75\linewidth}{!}{
    \begin{tabular}{c|ccccc}
    \toprule
    Method & \multicolumn{1}{c}{Training-free} & \multicolumn{1}{c}{\makecell{ Zero-shot \\ Generation}} & \multicolumn{1}{c}{\makecell{ Few-shot \\ Generation}} & \multicolumn{1}{c}{\makecell{ Diversity \\ Mask}} & \multicolumn{1}{c}{\makecell{ Controllable \\ Anomaly Types}} \\
    \midrule
    DR\AE M \cite{zavrtanik2021draem} & \cmark     & \cmark     & \xmark     & \xmark     & \xmark \\
    RealNet \cite{zhang2024realnet} & \xmark     & \cmark     & \xmark     & \xmark     & \xmark \\
    AnoGen \cite{gui2024few} & \xmark     & \xmark     & \cmark     & \xmark     & \cmark \\
    AnomalyDiff \cite{hu2024anomalydiffusion} & \xmark     & \xmark     & \cmark     & \cmark     & \cmark \\
    DualAnoDiff \cite{jin2025dual} & \xmark     & \xmark     & \cmark     & \cmark     & \cmark \\
    AnomalyAny \cite{sun2025unseen} & \cmark     & \cmark     & \xmark     & \cmark     & \cmark \\
    UniADC (ours) & \cmark     & \cmark     & \cmark     & \cmark     & \cmark \\
    \bottomrule
    \end{tabular}%
   }
  \label{tab:table1}%
\end{table}%

\subsection{Anomaly Sample-Guided Controllable Inpainting}

Anomaly sample-guided controllable inpainting aims to enrich the diversity of anomalies by repainting few-shot anomaly samples, making it well-suited for few-shot anomaly detection and classification tasks. Given a normal image $X$ and an anomalous sample $(X^y, M^y)$, we first crop the anomalous region from $X^y$ based on the mask $M^y$, and paste it at a random location on the normal image $X$, yielding a preliminary synthetic anomaly image $\hat{X}^{y}$ and its corresponding mask $\hat{M}^{y}$. To increase the shape variation, we apply identical data augmentations, such as affine transformations, to both the anomalous region and its mask. However, at this stage, the synthesized anomalies still suffer from limited diversity and suboptimal visual coherence. To improve diversity and image quality, we use the latent diffusion model along with an inpainting control network to repaint the pasted regions in $\hat{X}^{y}$. The repainting process is guided by simple text prompts (e.g., the name of the anomaly category) and follows the same diffusion and denoising procedures as described in anomaly prior-guided inpainting. The degree of diversity and deviation from the original anomaly is controlled by the noise factor $\gamma$. To ensure semantic consistency, we perform category consistency selection by computing the Structural Similarity Index (SSIM) \cite{wang2004image} between the original and repainted anomalous sub-images. These sub-images include both the anomalous region and a fixed amount of surrounding normal context, thereby validating the semantic consistency of the anomaly's spatial placement. We compute the SSIM-based category matching score for each sample and retain the highest-scoring synthetic anomaly sample for discriminator training. Table \ref{tab:table1} compares UniADC with other anomaly synthesis methods. To the best of our knowledge, UniADC is the first method capable of synthesizing anomalies under both zero-shot and few-shot settings. Moreover, it supports both diverse mask generation and fine-grained control over anomaly categories, thereby significantly enhancing its applicability across a wide range of anomaly detection and classification tasks.

\subsection{Implicit-Normal Discriminator}

Beyond the challenge of scarce anomalous samples, another core difficulty in anomaly detection and classification lies in the severe distribution imbalance between normal pixels and various categories of anomalous pixels. Conventional discriminative anomaly detection models \cite{zhou2024anomalyclip, li2024promptad, sadikaj2025multiads, nazer2025defect} typically align image patch features with class embeddings corresponding to "normal" and "anomaly" concepts through adapter fine-tuning or learnable text prompts. However, for multi-class anomaly classification, the long-tail effect of the data is significantly exacerbated. This causes such methods to inevitably overfit to the massive volume of normal pixels, ultimately leading to performance degradation in both anomaly detection and classification. To mitigate this, we propose the Implicit-Normal Discriminator (IND). Our key insight is to eliminate the explicit category embedding for the normal state. Instead of treating "normal" as a distinct class, we identify normal regions by the absence of all potential anomaly types. This exclusion-based logic forces the model to focus exclusively on the manifold of diverse anomaly concepts, thereby substantially alleviating the bias induced by imbalanced data distributions. Specifically, the proposed implicit-normal discriminator is trained using synthetic anomalous samples to align fine-grained image features with anomaly category embeddings, thereby enabling unified anomaly detection and classification. First, we represent each synthetic anomalous sample as a 4-tuple $(X, y, M_d, M_c)$, where $X$ is the synthesized anomalous image, $y$ is the anomaly category label, and $M_d$ and $M_c$ respectively denote the anomaly detection mask and classification mask, for a pixel location $(i, j)$ satisfying:
\begin{equation}
\label{eq:equ3}
M_{c}(i,j)=\left\{
\begin{aligned}
    y, & \qquad M_{d}(i,j)=1, \\
    0, & \qquad M_{d}(i,j)=0. \\
\end{aligned}
\right.\end{equation}

\begin{figure}[t]
\centering
\includegraphics[width=0.7\linewidth]{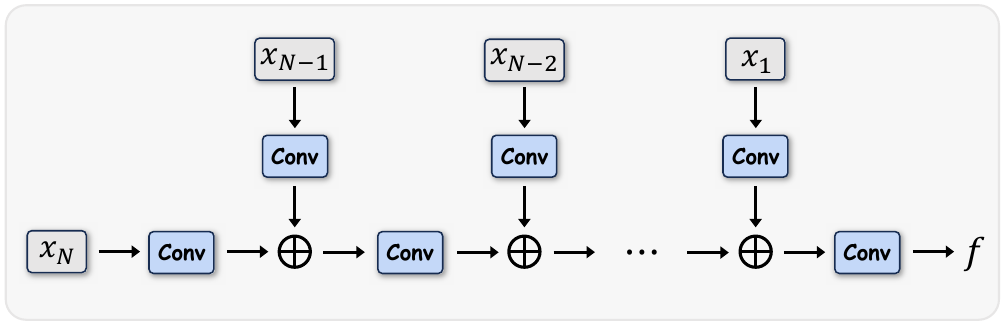}
\caption{Architecture of the Feature Fusion Network.}
\label{fig:fig4}
\end{figure}
\noindent We use a pre-trained vision encoder to extract multi-scale features $ \{x_1,x_2,\cdots,x_N\} = \varphi(X)$, and fuse these features into a fine-grained representation through a Feature Fusion Network (FFN) $f =\theta(x_1,x_2,\cdots,x_N)$, where $f \in \mathbb{R}^{H' \times W' \times C}$. As illustrated in Fig. \ref{fig:fig4}, the FFN progressively aggregates intermediate-layer features from high to low semantic levels via a fully convolutional architecture. Concurrently, a text encoder embeds anomaly category names into a semantic space to obtain embeddings $\{g_1, g_2, \cdots, g_Y\}$, where $g_y \in \mathbb{R}^{C}$. The similarity between the anomaly category embedding $g_y$ and the feature map $f$ at a spatial location $(h,w)$ is computed as: 
\begin{equation}
\label{eq:equ4}
s_{y}(h,w) = \sigma( \langle f(h,w),g_y \rangle/\epsilon),
\end{equation}
where $\sigma(\cdot)$ denotes the Sigmoid function, and $\epsilon$ is a learnable scaling parameter. We compute the similarity score at each spatial location and upsample the resulting similarity matrix to the original image resolution, yielding a classification map $S_y \in (0,1)^{H \times W}$, which indicates the likelihood of anomaly category $y$ at each pixel. By computing classification maps for all categories, we obtain a set of classification maps $\{S_1, S_2, \cdots, S_Y\}$. The pixel-level anomaly detection map $S_d$ is then computed as the average of these classification maps: $S_d= \frac{1}{Y} \Sigma_{y=1}^{Y}S_y $. The maximum value of $S_d$ is used as the image-level anomaly detection score $I_d$. Then, the pixel-level anomaly classification result at a pixel location $(i, j)$ is:
\begin{equation}
\label{eq:equ5}
S_{c}(i,j)=\left\{
\begin{aligned}
     \argmax_{ \mathsmallish{ y \in \{1,2,\cdots,Y\} } } S_y(i,j) ,  \qquad S_d(i,j) \geq \tau, \\
      0,\qquad\qquad  \qquad S_d(i,j) < \tau, \\
\end{aligned}
\right.\end{equation}
where $\tau$ represents the anomaly score threshold used to distinguish between normal and anomalous states. We use the anomaly category with the largest pixel area in $S_c$ as the image-level anomaly classification result $I_c$. If all pixels in $S_c$ are classified as normal, then $I_c$ is set to 0. We optimize the feature fusion network using Binary Focal loss \cite{lin2017focal} and Dice loss \cite{milletari2016v} for anomaly detection, and  Cross-Entropy loss for anomaly classification. The overall loss function is:
\begin{equation}
\label{eq:equ6}
\begin{aligned}
\mathcal{L} &=  \mathcal{L}_{Focal}(S_d,M_{d}) + \mathcal{L}_{Dice}(S_d,M_{d})  \\
              & +\lambda\mathcal{L}_{CE}([S_1, S_2, \cdots, S_Y],M_{c}),
\end{aligned}
\end{equation}
where $[\cdot, \cdot]$ denotes the concatenation operation, and $\lambda$ is a weight hyperparameter. The normal regions in $M_c$ are ignored during the computation of $\mathcal{L}_{CE}$. By excluding the normal category, our model is forced to refine the boundaries between various anomaly concepts, thereby enhancing its anomaly recognition performance and classification reliability.

\section{Experiment}

To validate the effectiveness of our proposed UniADC framework, we conducted a comprehensive evaluation of its anomaly detection and classification performance on the MVTec-FS \cite{lyu2025mvrec}, MTD \cite{huang2020surface}, WFDD \cite{chen2024unified}, and Real-IAD \cite{wang2024real} datasets. This section provides a detailed account of the experimental setup, main experimental results and analysis, as well as a systematic ablation study.

\begin{figure}[t]
\centering
\includegraphics[width=\linewidth]{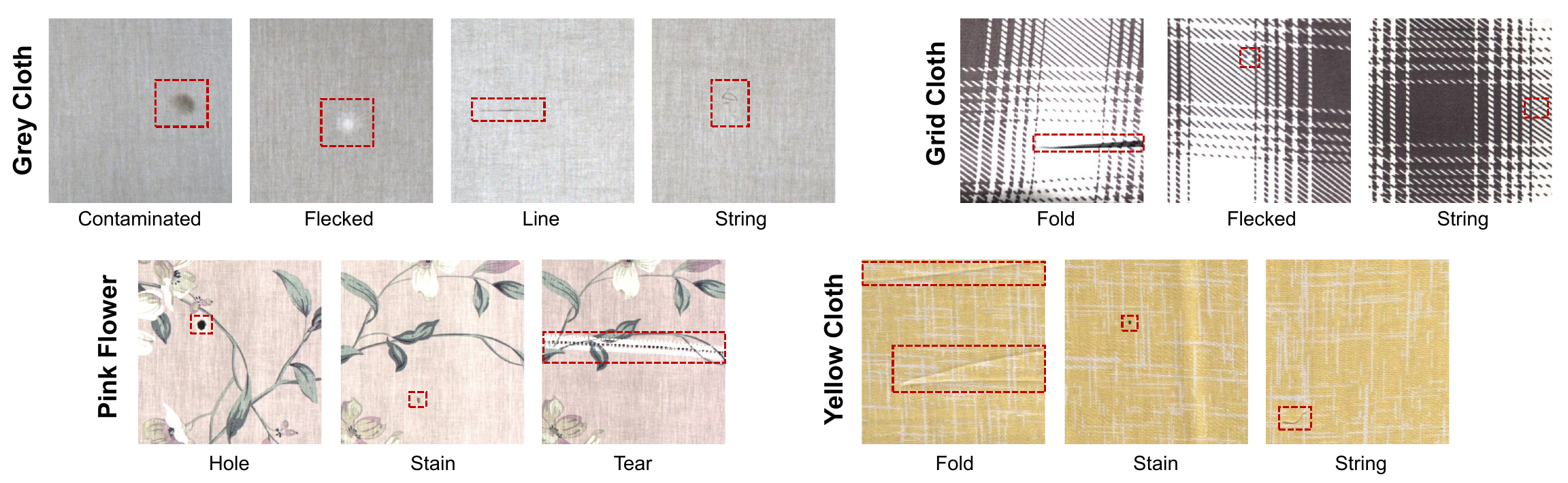}
\caption{Examples of different anomaly categories in the WFDD dataset, with anomalous regions marked by red bounding boxes.}
\label{fig:fig5}
\end{figure}

\subsection{Experimental Setup}

\begin{table*}[t]
\centering
 \renewcommand\arraystretch{0.75}
 \addtolength{\tabcolsep}{-2pt}
 \caption{Comparison of UniADC with alternative methods under zero-shot anomaly detection and classification settings on the MVTec-FS, MTD, and WFDD datasets. $\mathcal{y}$ denotes methods that use anomaly priors.}
 \resizebox{\linewidth}{!}{
    \begin{tabular}{c|c|ccccc|ccccc|ccccc}
    \toprule
    \multicolumn{1}{c|}{\multirow{2}[1]{*}{$ \;\; K_n \;\; $}} & \multirow{2}[1]{*}{Method} & \multicolumn{5}{c|}{MVTec-FS \cite{lyu2025mvrec}} & \multicolumn{5}{c|}{MTD \cite{huang2020surface}}     & \multicolumn{5}{c}{WFDD \cite{chen2024unified}} \\
    \cmidrule{3-17}          & \multicolumn{1}{c|}{} & \multicolumn{1}{c}{I-AUC} & \multicolumn{1}{c}{P-AUC} & \multicolumn{1}{c}{PRO} & \multicolumn{1}{c}{Acc} & \multicolumn{1}{c|}{mIoU} & \multicolumn{1}{c}{I-AUC} & \multicolumn{1}{c}{P-AUC} & \multicolumn{1}{c}{PRO} & \multicolumn{1}{c}{Acc} & \multicolumn{1}{c|}{mIoU} & \multicolumn{1}{c}{I-AUC} & \multicolumn{1}{c}{P-AUC} & \multicolumn{1}{c}{PRO} & \multicolumn{1}{c}{Acc} & \multicolumn{1}{c}{mIoU} \\
    \midrule
    \multirow{6}[6]{*}{1} & InCTRL \cite{zhu2024toward} & 92.36 &   -    &  -     &   -    &      - & 70.34 &     -  &   -    &      - &    -   & 96.66 &    -   &    -   &     -  & - \\
          & PatchCore \cite{roth2022towards} & 84.76 & 93.77 & 82.44 &   -    &   -    & 68.22 & 73.44 & 59.50  &   -    &   -    & 84.81 & 96.07 & 70.62 &   -    & - \\
          & AnomalyDINO \cite{damm2025anomalydino} & 96.12	& \textbf{96.54} & 89.04 &   -    &   -    & 85.32 & 78.01 & 74.11  &   -    &   -    & 97.42  & 98.86 & 87.88 &   -    & - \\
          & $\mathcal{y}$ PromptAD \cite{li2024promptad} & 91.80  & 95.09 & 87.06 &    -   &   -    & 86.27 & 71.70  & 70.61 &    -   &   -    & 96.90  & 97.10  & 86.72 &   -    & - \\
          & $\mathcal{y}$ AnomalyGPT \cite{gu2024anomalygpt} & 93.48 & 95.92 & 86.60  & 45.85 &   -    & 71.91 & 68.41 & 58.67 & 27.57 &    -   & 96.89 & 97.72 & 85.69 & 45.76 &  -\\
          & $\mathcal{y}$ WinCLIP \cite{jeong2023winclip} & 93.20  & 94.43 & 86.47 & 40.75 & 25.17 & 77.53 & 69.26 & 57.58 & 28.69 & 15.10  & 95.72 & 93.94 & 78.98 & 34.64 & 27.53 \\
          & \cellcolor{gray!15}$\mathcal{y}$ \textbf{UniADC{\fontsize{7}{7}\selectfont \texttt{(CLIP)}}}  & \cellcolor{gray!15}95.03 & \cellcolor{gray!15}96.33 & \cellcolor{gray!15}88.75 & \cellcolor{gray!15}66.07 & \cellcolor{gray!15}32.90  & \cellcolor{gray!15}86.19 & \cellcolor{gray!15}80.80 & \cellcolor{gray!15}77.69 & \cellcolor{gray!15}55.82 &  \cellcolor{gray!15}27.92  &  \cellcolor{gray!15}97.01  & \cellcolor{gray!15}99.00  &  \cellcolor{gray!15}87.03  & \cellcolor{gray!15}86.70 &  \cellcolor{gray!15}45.55	\\
          & \cellcolor{gray!15}$\mathcal{y}$ \textbf{UniADC{\fontsize{7}{7}\selectfont \texttt{(DINO)}}}  & \cellcolor{gray!15}\textbf{96.37} & \cellcolor{gray!15}96.11 & \cellcolor{gray!15}\textbf{89.16} & \cellcolor{gray!15}\textbf{68.30} & \cellcolor{gray!15}\textbf{35.06}  & \cellcolor{gray!15}\textbf{90.09} & \cellcolor{gray!15}\textbf{82.65} & \cellcolor{gray!15}\textbf{79.98} & \cellcolor{gray!15}\textbf{59.22} &  \cellcolor{gray!15}\textbf{29.58}  &  \cellcolor{gray!15}\textbf{98.08}   & \cellcolor{gray!15}\textbf{99.46}  &  \cellcolor{gray!15}\textbf{92.13}  & \cellcolor{gray!15}\textbf{88.88} & \cellcolor{gray!15}\textbf{52.17} \\
    \midrule
    \multirow{6}[6]{*}{2} & InCTRL \cite{zhu2024toward} & 93.01 &   -    &   -    &   -    &    -   & 72.07 &   -    &    -   &    -   &    -   & 97.36 &  -     &    -   &   -    & - \\
          & PatchCore \cite{roth2022towards} & 88.49 & 94.43 & 84.65 &   -    &   -    & 69.90  & 75.07 & 60.52 &    -   &   -    & 88.06 & 96.35 & 71.45 &    -   & - \\
          & AnomalyDINO \cite{damm2025anomalydino} & 96.87	& \textbf{97.14} & 90.96 &   -    &   -    & 86.32 & 79.27 & 76.82 &    -   &   -    & 97.72	& 99.13 &  89.03 &    -   & - \\
          & $\mathcal{y}$ PromptAD \cite{li2024promptad} & 93.95 & 95.42 & 87.93 &    -   &    -   & 87.06 & 73.53 & 69.91 &   -    &  -     & 97.15 & 97.20  & 86.81 &    -   & - \\
          & $\mathcal{y}$ AnomalyGPT \cite{gu2024anomalygpt} & 94.91 & 96.24 & 87.97 & 52.65 &    -   & 72.57 & 70.20  & 60.12 & 32.56 &    -   & 97.48 & 97.78 & 85.75 & 47.39 & - \\
          & $\mathcal{y}$ WinCLIP \cite{jeong2023winclip} & 94.37 & 94.60  & 86.95 & 41.93 & 25.34 & 78.07 & 71.57 & 57.70  & 29.90  & 15.28 & 96.50  & 94.16 & 80.19 & 35.83 & 27.62 \\
          &  \cellcolor{gray!15}$\mathcal{y}$ \textbf{UniADC{\fontsize{7}{7}\selectfont \texttt{(CLIP)}}}  &   \cellcolor{gray!15}95.27   &  \cellcolor{gray!15}96.53  & \cellcolor{gray!15}88.52  & \cellcolor{gray!15}71.64  & \cellcolor{gray!15}35.12  &  \cellcolor{gray!15}90.31 &  \cellcolor{gray!15}82.92 & \cellcolor{gray!15}78.99  &  \cellcolor{gray!15}62.87 &  \cellcolor{gray!15}30.56 & \cellcolor{gray!15}97.55  &   \cellcolor{gray!15}98.98 & \cellcolor{gray!15}87.14  & \cellcolor{gray!15}\textbf{89.48}  & \cellcolor{gray!15}48.15 \\
          &  \cellcolor{gray!15}$\mathcal{y}$ \textbf{UniADC{\fontsize{7}{7}\selectfont \texttt{(DINO)}}}  &   \cellcolor{gray!15}\textbf{97.09}   &  \cellcolor{gray!15}97.04  & \cellcolor{gray!15}\textbf{92.15}  & \cellcolor{gray!15}\textbf{74.74}  & \cellcolor{gray!15}\textbf{36.66}  &  \cellcolor{gray!15}\textbf{92.30} &  \cellcolor{gray!15}\textbf{83.52} &  \cellcolor{gray!15}\textbf{81.22} &  \cellcolor{gray!15}\textbf{64.15} &  \cellcolor{gray!15}\textbf{31.85} & \cellcolor{gray!15}\textbf{98.21}  &   \cellcolor{gray!15}\textbf{99.46} & \cellcolor{gray!15}\textbf{92.35}  & \cellcolor{gray!15}89.22 &  \cellcolor{gray!15}\textbf{53.38} \\
    \midrule
    \multirow{6}[6]{*}{4} & InCTRL \cite{zhu2024toward} & 93.62 &    -   &   -    &     -  &    -   &   73.23    &  -     &   -    &   -    &     -  & 97.38      &    -   &      - &   -    & - \\
          & PatchCore \cite{roth2022towards} & 90.75 & 95.28 & 86.56 &    -   &   -    &   71.40	&   75.35    &  59.67     &    -   &   -    &  88.26	&  	97.21     &   71.41    &   -    &  - \\
          & AnomalyDINO \cite{damm2025anomalydino} & 97.50	& 97.30 & 92.21 &    -   &   -    &  89.68	&  83.11  &  78.61  &    -   &   -    & 98.15 &  \textbf{99.43}   &  91.54  &   -    &  - \\
          & $\mathcal{y}$ PromptAD \cite{li2024promptad} & 94.91 & 95.92 & 89.86 &     -  &     -  &    88.16	&   73.82    &    71.49   &    -   &     -   &   96.79	&   97.53    &  87.80     &    -   & - \\
          & $\mathcal{y}$ AnomalyGPT \cite{gu2024anomalygpt} & 96.10  & 96.42 & 91.09 & 56.59 &   -    &   74.38	&   68.68    &   59.92    &  42.48     &    -   &    98.00	&   97.92    &   86.35    &  48.24     & -  \\
          & $\mathcal{y}$ WinCLIP \cite{jeong2023winclip} & 95.17 & 94.98 & 87.67 & 42.69 & 25.70  &   78.90   &    69.84   &    58.11   &    29.91   &  16.17    &    96.74   &   94.58 &  80.28     &   34.99    & 27.79 \\
          &  \cellcolor{gray!15}$\mathcal{y}$ \textbf{UniADC{\fontsize{7}{7}\selectfont \texttt{(CLIP)}}}  &  \cellcolor{gray!15}96.18 & \cellcolor{gray!15}96.47  & \cellcolor{gray!15}91.17  &  \cellcolor{gray!15}73.33 & \cellcolor{gray!15}35.14  &  \cellcolor{gray!15}90.88 &  \cellcolor{gray!15}84.88  &  \cellcolor{gray!15}79.81  &  \cellcolor{gray!15}63.98   & \cellcolor{gray!15}31.29  & \cellcolor{gray!15}98.15   &  \cellcolor{gray!15}99.09 &   \cellcolor{gray!15}88.29  & \cellcolor{gray!15}90.85 &  \cellcolor{gray!15}48.24 \\
           &  \cellcolor{gray!15}$\mathcal{y}$ \textbf{UniADC{\fontsize{7}{7}\selectfont \texttt{(DINO)}}}  & \cellcolor{gray!15}\textbf{97.65}  & \cellcolor{gray!15}\textbf{97.36} & \cellcolor{gray!15}\textbf{92.68}  &  \cellcolor{gray!15}\textbf{76.35} & \cellcolor{gray!15}\textbf{37.23}  &  \cellcolor{gray!15}\textbf{93.04} &  \cellcolor{gray!15}\textbf{85.82}  &  \cellcolor{gray!15}\textbf{81.43}  &   \cellcolor{gray!15}\textbf{66.03}  & \cellcolor{gray!15}\textbf{33.75}  &  \cellcolor{gray!15}\textbf{98.60}  & \cellcolor{gray!15}99.39  &  \cellcolor{gray!15}\textbf{94.76}  & \cellcolor{gray!15}\textbf{91.06} &  \cellcolor{gray!15}\textbf{55.44} \\
    \bottomrule
    \end{tabular}%
    }
  \label{tab:table2}%
\end{table*}%

\textit{1) Datasets.} We conduct extensive evaluations on four anomaly detection and classification datasets, including MVTec-FS \cite{lyu2025mvrec}, MTD \cite{huang2020surface}, WFDD \cite{chen2024unified}, and Real-IAD \cite{wang2024real}. The MVTec-FS dataset is introduced by ZipAF \cite{lyu2025mvrec} as an extension of the MVTec-AD benchmark \cite{bergmann2019mvtec} for anomaly classification. It contains 15 types of industrial products, each with an average of four anomaly categories. The MTD dataset \cite{huang2020surface} consists of 1,344 images of magnetic tiles, including five types of anomalies. The subtle differences among these anomalies make the classification task particularly challenging. The WFDD dataset \cite{chen2024unified} comprises 4,101 images across four fabric types: Grey Cloth, Grid Cloth, Pink Flower, and Yellow Cloth. Since it was originally designed for anomaly detection, it lacked a detailed categorization of anomaly types. To facilitate our proposed anomaly detection and classification task, we reclassified the dataset by grouping identical defects, resulting in an average of three anomaly categories per fabric. Representative examples for each category are illustrated in Fig. \ref{fig:fig5}. The Real-IAD dataset \cite{wang2024real} comprises 151,050 multi-view images across 30 different products, with an average of four defect categories per product. Some of these products contain subtle structural anomalies that are difficult to detect and classify. We adopt the single-view experimental setting, following recent works \cite{lee2024continuous, zhang2025towards}. Other widely-used anomaly detection datasets, such as VisA \cite{zou2022spot}, are excluded from this work due to the lack of anomaly category labels.

\begin{table*}[t]
  \centering
     \renewcommand\arraystretch{0.7}
     \addtolength{\tabcolsep}{-2pt}
   \caption{Comparison of UniADC with alternative methods under few-shot anomaly detection and classification settings on the MVTec-FS, MTD, and WFDD datasets.}
   \resizebox{1.0\linewidth}{!}{
    \begin{tabular}{c|c|c|ccccc|ccccc|ccccc}
    \toprule
    \multicolumn{1}{c|}{\multirow{2}[1]{*}{$K_n$}} & \multicolumn{1}{c|}{\multirow{2}[1]{*}{$K_a$}} & \multirow{2}[1]{*}{Method} & \multicolumn{5}{c|}{MVTec-FS \cite{lyu2025mvrec}} & \multicolumn{5}{c|}{MTD \cite{huang2020surface}}     & \multicolumn{5}{c}{WFDD \cite{chen2024unified}} \\
\cmidrule{4-18}          &       & \multicolumn{1}{c|}{} & \multicolumn{1}{c}{I-AUC} & \multicolumn{1}{c}{P-AUC} & \multicolumn{1}{c}{PRO} & \multicolumn{1}{c}{Acc} & \multicolumn{1}{c|}{mIoU} & \multicolumn{1}{c}{I-AUC} & \multicolumn{1}{c}{P-AUC} & \multicolumn{1}{c}{PRO} & \multicolumn{1}{c}{Acc} & \multicolumn{1}{c|}{mIoU} & \multicolumn{1}{c}{I-AUC} & \multicolumn{1}{c}{P-AUC} & \multicolumn{1}{c}{PRO} & \multicolumn{1}{c}{Acc} & \multicolumn{1}{c}{mIoU} \\
    \midrule
    \multirow{6}[6]{*}{1} & \multirow{6}[6]{*}{1} & PatchCore{\fontsize{8}{8}\selectfont +ZipAF} \cite{roth2022towards} & 84.76 & 93.77 & 82.44 & 52.10  & 37.30  & 68.22 & 73.44 & 59.50  & 28.26 & 20.33 & 84.81 & 96.07 & 70.62 & 61.75 & 39.10 \\
          &       & WinCLIP{\fontsize{8}{8}\selectfont +ZipAF} \cite{jeong2023winclip} & 93.20  & 94.43 & 86.47 & 49.60  & 34.39 & 77.53 & 69.26 & 57.58 & 33.78 & 24.23 & 95.72 & 93.94 & 78.98 & 59.59 & 37.30 \\
          &       & PromptAD{\fontsize{8}{8}\selectfont +ZipAF} \cite{li2024promptad} & 92.74 & 95.72 & 88.06 & 49.12 & 39.57 & 86.85 & 72.30  & 70.87 & 31.79 & 28.92 & 97.47 & 97.51 & 86.23 & 64.11 & 44.60 \\
          &       & AnomalyCLIP{\fontsize{8}{8}\selectfont +ZipAF} \cite{zhou2024anomalyclip} & 95.59 & 96.00    & 87.98 & 52.60  & 40.25 & 76.18 & 75.98 & 69.54 & 25.61 & 21.78 & 93.76 & 98.59 & 87.58 & 76.48 & 45.99
          \\
           &       & NAGL{\fontsize{8}{8}\selectfont +ZipAF} \cite{wang2025normalabnormal} & 95.60	&  96.51  & \textbf{92.79} & 55.92  & 42.00	& 84.98	& 75.45 & 75.77 & 36.37 & 30.89 & 97.20	& 98.93  & 92.40 & 71.82 & 	45.32\\
           &     & SegDINO \cite{yang2025segdino}  & 79.19	&  92.54  & 73.15 & 53.85  & 25.26 & 63.89	& 73.27 & 50.66 & 40.22 & 20.53 & 81.02	& 91.96  & 70.81 & 59.96 & 28.02 \\
          &       & \cellcolor{gray!15}\textbf{UniADC{\fontsize{7}{7}\selectfont \texttt{(CLIP)}}} & \cellcolor{gray!15}97.69 & \cellcolor{gray!15}98.53  & \cellcolor{gray!15}89.72  & \cellcolor{gray!15}84.72  & \cellcolor{gray!15}48.62 & \cellcolor{gray!15}88.49  & \cellcolor{gray!15}82.41 & \cellcolor{gray!15}77.41 & \cellcolor{gray!15}60.81  & \cellcolor{gray!15}\textbf{34.80} & \cellcolor{gray!15}98.50 & \cellcolor{gray!15}98.83 &  \cellcolor{gray!15}92.96 & \cellcolor{gray!15}93.51 &  \cellcolor{gray!15}48.76	\\
          &       & \cellcolor{gray!15}\textbf{UniADC{\fontsize{7}{7}\selectfont \texttt{(DINO)}}} & \cellcolor{gray!15}\textbf{98.42} & \cellcolor{gray!15}\textbf{98.96}  & \cellcolor{gray!15}92.26  & \cellcolor{gray!15}\textbf{86.74}  & \cellcolor{gray!15}\textbf{51.28} & \cellcolor{gray!15}\textbf{91.41} &  \cellcolor{gray!15}\textbf{82.88} & \cellcolor{gray!15}\textbf{80.07} & \cellcolor{gray!15}\textbf{62.47} & \cellcolor{gray!15}32.21 &  \cellcolor{gray!15}\textbf{99.85} & \cellcolor{gray!15}\textbf{99.37} & \cellcolor{gray!15}\textbf{94.07}  & \cellcolor{gray!15}\textbf{96.10} & \cellcolor{gray!15}\textbf{50.53} \\
    \midrule
    \multirow{10}[24]{*}{2} & \multirow{6}[6]{*}{1} & PatchCore{\fontsize{8}{8}\selectfont +ZipAF} \cite{roth2022towards} & 88.49 & 94.43 & 84.65 & 56.16 & 39.63 & 69.90  & 75.07 & 60.52 & 30.15 & 21.03 & 88.06 & 96.35 & 71.45 & 63.82 & 39.71 \\
          &       & WinCLIP{\fontsize{8}{8}\selectfont +ZipAF} \cite{jeong2023winclip} & 94.37 & 94.60  & 86.95 & 49.99 & 34.77 & 78.07 & 71.57 & 57.70  & 33.31 & 24.54 & 96.50  & 94.16 & 80.19 & 60.21 & 37.28 \\
          &       & PromptAD{\fontsize{8}{8}\selectfont +ZipAF} \cite{li2024promptad} & 94.58 & 95.66 & 88.93 & 51.40  & 40.09 & 87.41 & 74.10  & 71.12 & 36.09 & 29.78 & 97.52 & 97.74 & 87.01 & 67.17 & 44.51 \\
          &       & AnomalyCLIP{\fontsize{8}{8}\selectfont +ZipAF} \cite{zhou2024anomalyclip} & 95.94 & 96.12 & 88.77 & 53.80  & 40.90  & 76.42 & 76.11 & 70.62 & 26.57 & 21.17 & 94.96 & 98.75 & 87.80  & 77.99 & 46.49 \\
           &       & NAGL{\fontsize{8}{8}\selectfont +ZipAF} \cite{wang2025normalabnormal} & 96.41	&   96.56 & \textbf{92.96} & 57.78 & 42.34	& 84.92	& 77.71 & 77.72 & 39.75 & 31.49 &  97.47&  98.89 & 92.60 & 72.40 & 47.19 \\  &     & SegDINO \cite{yang2025segdino}  & 79.31	& 89.89   & 68.12 & 53.88  & 25.40 & 63.51 & 75.36 & 51.11 & 39.45 & 18.82 & 78.47	& 89.59  & 69.43 & 58.17 & 26.64\\

          &       & \cellcolor{gray!15}\textbf{UniADC{\fontsize{7}{7}\selectfont \texttt{(CLIP)}}} & \cellcolor{gray!15}98.35 & \cellcolor{gray!15}98.63 & \cellcolor{gray!15}90.64 & \cellcolor{gray!15}85.20 & \cellcolor{gray!15}48.93 & \cellcolor{gray!15}90.70 & \cellcolor{gray!15}84.49 & \cellcolor{gray!15}80.71 & \cellcolor{gray!15}63.78 & \cellcolor{gray!15}\textbf{35.58} & \cellcolor{gray!15}98.97 & \cellcolor{gray!15}99.16 & \cellcolor{gray!15}\textbf{94.64} & \cellcolor{gray!15}93.66 & \cellcolor{gray!15}49.71\\
          &       & \cellcolor{gray!15}\textbf{UniADC{\fontsize{7}{7}\selectfont \texttt{(DINO)}}} & \cellcolor{gray!15}\textbf{98.56} & \cellcolor{gray!15}\textbf{98.90} & \cellcolor{gray!15}92.48 & \cellcolor{gray!15}\textbf{86.85} & \cellcolor{gray!15}\textbf{51.49} & \cellcolor{gray!15}\textbf{92.57} & \cellcolor{gray!15}\textbf{86.79} & \cellcolor{gray!15}\textbf{83.17} & \cellcolor{gray!15}\textbf{65.10} & \cellcolor{gray!15}34.33 & \cellcolor{gray!15}\textbf{99.87} & \cellcolor{gray!15}\textbf{99.48} & \cellcolor{gray!15}94.12 & \cellcolor{gray!15}\textbf{97.22} & \cellcolor{gray!15}\textbf{51.78} \\
\cmidrule{2-18}          & \multirow{6}[6]{*}{2} & PatchCore{\fontsize{8}{8}\selectfont +ZipAF} \cite{roth2022towards} & 88.49 & 94.43 & 84.65 & 58.25 & 40.32 &   69.90    &   75.07   &  60.52     &   34.71    &   21.78  &    88.06   &   96.35    &   71.45    &  64.52 & 41.29\\
          &       & WinCLIP{\fontsize{8}{8}\selectfont +ZipAF} \cite{jeong2023winclip} & 94.37 & 94.60  & 86.95 & 51.76 & 35.25 &   78.07    &     71.57   &   57.70    &  37.36     &    27.14   &   96.50    &   94.16    &  80.19     &    61.64   &  37.69\\
          &       & PromptAD{\fontsize{8}{8}\selectfont +ZipAF} \cite{li2024promptad} & 95.16 & 95.60  & 89.03 & 52.66 & 40.93 &   88.46    &   74.53    &     71.91  &   43.89    &    31.16   & 97.75 &  98.20 & 86.81 & 67.91 & 45.25\\
          &       & AnomalyCLIP{\fontsize{8}{8}\selectfont +ZipAF} \cite{zhou2024anomalyclip} & 95.93 & 96.47 & 89.12 & 52.93 & 42.49 &    76.72   &   76.32    &   73.90    &   35.57    &  24.81     &   95.25    &    98.73   &   90.38    &    81.17   & 47.74 \\
          &       & NAGL{\fontsize{8}{8}\selectfont +ZipAF} \cite{wang2025normalabnormal} & 96.74	& 97.17  & 92.80 & 61.33 & 45.30 & 85.77& 80.46 & 78.18 & 47.21 & 35.01 & 97.56	& 98.90  & 92.90 & 76.49 & 48.85 \\
           &     & SegDINO \cite{yang2025segdino}  & 84.42	&  91.98  & 74.50 & 58.49  & 30.21 & 67.65	& 77.72 & 67.54 & 42.17 & 21.29 & 85.26	& 92.66  & 74.99 & 59.79 & 30.28 \\

          &       & \cellcolor{gray!15}\textbf{UniADC{\fontsize{7}{7}\selectfont \texttt{(CLIP)}}} & \cellcolor{gray!15}98.75 & \cellcolor{gray!15}98.88 & \cellcolor{gray!15}93.31 & \cellcolor{gray!15}\textbf{89.88} & \cellcolor{gray!15}\textbf{54.93} & \cellcolor{gray!15}93.15  &  \cellcolor{gray!15}86.20  & \cellcolor{gray!15}83.36  &  \cellcolor{gray!15}\textbf{71.34}  &  \cellcolor{gray!15}\textbf{37.05}   &    \cellcolor{gray!15}99.14 &  \cellcolor{gray!15}99.34 &  \cellcolor{gray!15}\textbf{94.85}  &  \cellcolor{gray!15}94.90 & \cellcolor{gray!15}51.06 \\
          &       & \cellcolor{gray!15}\textbf{UniADC{\fontsize{7}{7}\selectfont \texttt{(DINO)}}} & \cellcolor{gray!15}\textbf{99.05} & \cellcolor{gray!15}\textbf{99.10} & \cellcolor{gray!15}\textbf{93.55} & \cellcolor{gray!15}88.72 & \cellcolor{gray!15}54.21 & \cellcolor{gray!15}\textbf{93.85} &  \cellcolor{gray!15}\textbf{89.31}  &  \cellcolor{gray!15}\textbf{85.78} &  \cellcolor{gray!15}70.87  &  \cellcolor{gray!15}35.46   &    \cellcolor{gray!15}\textbf{99.93} & \cellcolor{gray!15}\textbf{99.49}  & \cellcolor{gray!15}94.19  & \cellcolor{gray!15}\textbf{97.76}  &  \cellcolor{gray!15}\textbf{51.91} \\
    \midrule
    \multirow{6}[6]{*}{4} & \multirow{6}[6]{*}{1} & PatchCore{\fontsize{8}{8}\selectfont +ZipAF} \cite{roth2022towards} & 90.75 & 95.28 & 86.56 & 60.32 & 39.80  &  71.40  &  75.35 &  59.67 &   31.58   &  21.92   &  88.26   &   97.21 &   71.41    &   64.17    & 42.93 \\
              &       & WinCLIP{\fontsize{8}{8}\selectfont +ZipAF} \cite{jeong2023winclip} & 95.17 & 94.98 & 87.67 & 53.75 & 35.98 &    78.90   &   69.84   &   58.11   &   33.89    &   28.32   &   96.74   &   94.58    &  80.28    &  63.59    & 37.59 \\
          &       & PromptAD{\fontsize{8}{8}\selectfont +ZipAF} \cite{li2024promptad} & 96.49 & 96.10  & 90.02 & 55.21 & 41.66 &  87.78  &   73.82  &  71.49   &  38.36  &  30.13  &  97.57  &   97.83   &  87.80   &   70.59   & 44.87 \\
          &       & AnomalyCLIP{\fontsize{8}{8}\selectfont +ZipAF} \cite{zhou2024anomalyclip} & 95.98 & 96.67 & 89.77 & 55.07 & 40.83 &  76.84   &   76.22    & 	71.11   &    29.58   &  22.37    &  95.22    &   98.74   & 90.17   &   78.88   &  46.97 \\
          &       & NAGL{\fontsize{8}{8}\selectfont +ZipAF} \cite{wang2025normalabnormal} & 97.14	& 96.75 & 93.24 & 59.49 & 42.52 & 85.71	& 79.49  & 79.15 & 42.42 & 32.07 & 97.94 &  99.07 &  92.72 &  75.30 &  49.40 \\
           &     & SegDINO \cite{yang2025segdino}  & 82.33	&  90.95  & 72.00	 & 52.74  & 25.11 & 64.88	& 71.73 & 51.33 & 39.99 & 19.38 & 78.40	& 90.87  & 71.62 & 57.44 & 26.36\\
          &       & \cellcolor{gray!15}\textbf{UniADC{\fontsize{7}{7}\selectfont \texttt{(CLIP)}}} & \cellcolor{gray!15}98.57 & \cellcolor{gray!15}98.55 & \cellcolor{gray!15}91.42 & \cellcolor{gray!15}85.57 & \cellcolor{gray!15}50.01 & \cellcolor{gray!15}92.74 & \cellcolor{gray!15}84.98 & \cellcolor{gray!15}82.15 & \cellcolor{gray!15}64.40  & \cellcolor{gray!15}\textbf{35.86}  & \cellcolor{gray!15}99.35  &   \cellcolor{gray!15}99.19 & \cellcolor{gray!15}94.48 & \cellcolor{gray!15}93.91  & \cellcolor{gray!15}49.94 \\
          &       & \cellcolor{gray!15}\textbf{UniADC{\fontsize{7}{7}\selectfont \texttt{(DINO)}}} & \cellcolor{gray!15}\textbf{98.70} & \cellcolor{gray!15}\textbf{98.91} & \cellcolor{gray!15}\textbf{93.35} & \cellcolor{gray!15}\textbf{86.83} & \cellcolor{gray!15}\textbf{52.37} & \cellcolor{gray!15}\textbf{93.32} & \cellcolor{gray!15}\textbf{87.12} & \cellcolor{gray!15}\textbf{83.83} & \cellcolor{gray!15}\textbf{66.59} &  \cellcolor{gray!15}34.87 &  \cellcolor{gray!15}\textbf{99.87} &   \cellcolor{gray!15}\textbf{99.45} & \cellcolor{gray!15}\textbf{94.59} &  \cellcolor{gray!15}\textbf{97.67} & \cellcolor{gray!15}\textbf{51.80} \\
    \bottomrule
    \end{tabular}%
    }
  \label{tab:table3}%
\end{table*}%

\textit{2) Implementation Details.} We adopt a consistent experimental setup for both anomaly prior-guided and anomaly sample-guided inpainting. Specifically, we use the DDIM sampler \cite{song2021denoising} with 1,000 original diffusion steps. The noise factor $\gamma$ is uniformly sampled between 0.4 and 0.6, and the number of accelerated sampling steps is set to 10. For mask generation, we leverage BiRefNet \cite{zheng2024birefnet} for binary segmentation to ensure that the mask is located within the foreground region. We set the mini-batch size to 32 for category consistency selection, and generate 16 samples with a resolution of $512 \times 512$ per anomaly category for discriminator training. For the implicit-normal discriminator, we adopt two experimental settings, namely \textbf{UniADC{\fontsize{7}{7}\selectfont \texttt{(CLIP)}}} and \textbf{UniADC{\fontsize{7}{7}\selectfont \texttt{(DINO)}}}, which use the CLIP ViT-L/14 \cite{radford2021learning} and the DINOv3-based \texttt{dino.txt} \cite{simeoni2025dinov3} as the vision-language backbone of the discriminator, respectively. We set both the anomaly score threshold $\tau$ and the loss weight $\lambda$ to 0.5. The list of anomaly priors is provided in the \textit{Supplementary Material}.

\textit{3) Metrics.} We use the Area Under the Receiver Operating Characteristic Curve (I-AUC) metric to evaluate image-level anomaly detection performance. For pixel-level anomaly location, we use Pixel-AUROC (P-AUC) and Per Region Overlap (PRO) \cite{bergmann2020uninformed} metrics. In addition, we report the Top-1 classification accuracy (Acc) and Mean Intersection over Union (mIoU) to evaluate image-level and pixel-level anomaly classification performance, respectively. For the "combined" category in MVTec-FS dataset \cite{lyu2025mvrec}, we only evaluate its anomaly detection performance and exclude it from the anomaly classification metrics, as in previous works \cite{lyu2025mvrec,huang2025anomalyncd}.

\textit{4) Baselines.} In the zero-shot setting, we adopt InCTRL \cite{zhu2024toward}, PatchCore \cite{roth2022towards}, AnomalyDINO \cite{damm2025anomalydino}, PromptAD \cite{li2024promptad}, AnomalyGPT \cite{gu2024anomalygpt}, and WinCLIP \cite{jeong2023winclip} as baseline methods. For AnomalyGPT \cite{gu2024anomalygpt}, we cast classification as a single-choice QA task by prompting the model with candidate anomaly types. For WinCLIP \cite{jeong2023winclip}, we provide it with anomaly category descriptions and compute the similarity between patch features and anomaly category embeddings, obtaining the anomaly classification results. In the few-shot setting, we combine anomaly detection methods such as PatchCore \cite{roth2022towards}, WinCLIP \cite{jeong2023winclip}, PromptAD \cite{li2024promptad}, AnomalyCLIP \cite{zhou2024anomalyclip}, and NAGL \cite{wang2025normalabnormal} with the anomaly classification method ZipAF \cite{lyu2025mvrec} as our baseline methods. We employ the threshold selection strategy proposed in AnomalyNCD \cite{huang2025anomalyncd} to determine the anomaly score threshold, and fine-tune PromptAD \cite{li2024promptad} and AnomalyCLIP \cite{zhou2024anomalyclip} using few-shot anomaly samples to enhance their anomaly detection performance. Additionally, we compared SegDINO \cite{yang2025segdino}, a DINO-based semantic segmentation model, assessing its applicability to this task. To ensure a fair comparison, the visual backbones of CLIP- and DINO-based methods are kept consistent with \textbf{UniADC{\fontsize{7}{7}\selectfont \texttt{(CLIP)}}} and \textbf{UniADC{\fontsize{7}{7}\selectfont \texttt{(DINO)}}}, respectively.

\begin{table}[t]
  \centering
  \renewcommand\arraystretch{0.75}
  \caption{Comparison of UniADC with alternative methods in the few-shot anomaly detection and classification setting on the Real-IAD dataset ($K_n=2$, $K_a=1$).}
     \resizebox{0.6\linewidth}{!}{
    \begin{tabular}{c|ccc|cc}
    \toprule
     \multirow{2}[1]{*}{Method} & \multicolumn{3}{c|}{Detection} & \multicolumn{2}{c}{Classification} \\
    \cmidrule{2-6}    \multicolumn{1}{c|}{} & \multicolumn{1}{c}{I-AUC} & \multicolumn{1}{c}{P-AUC} & \multicolumn{1}{c|}{PRO} & \multicolumn{1}{c}{Acc} & \multicolumn{1}{c}{mIoU} \\
    \midrule
    WinCLIP{\fontsize{8}{8}\selectfont +ZipAF} \cite{jeong2023winclip} & 82.57 & 94.87 & 80.48 & 51.96 & 28.18 \\
    PromptAD{\fontsize{8}{8}\selectfont +ZipAF} \cite{li2024promptad} & 84.28 & 96.21 & 85.68 & 54.92 & 31.78 \\
    AnomalyCLIP{\fontsize{8}{8}\selectfont +ZipAF} \cite{zhou2024anomalyclip} & 83.47 & 95.98 & 86.09 & 52.73 & 32.03 \\
    NAGL{\fontsize{8}{8}\selectfont +ZipAF} \cite{wang2025normalabnormal} & 87.38 & 97.83 & 90.24 & 58.80  & 34.45 \\
    SegDINO \cite{yang2025segdino} & 62.16 & 86.05 & 78.21 & 54.40  & 22.34 \\
    \cellcolor{gray!15}\textbf{UniADC{\fontsize{7}{7}\selectfont \texttt{(CLIP)}}} & \cellcolor{gray!15}87.59 & \cellcolor{gray!15}97.39 & \cellcolor{gray!15}89.71 & \cellcolor{gray!15}70.24 & \cellcolor{gray!15}34.49 \\
    \cellcolor{gray!15}\textbf{UniADC{\fontsize{7}{7}\selectfont \texttt{(DINO)}}} & \cellcolor{gray!15}\textbf{89.17} & \cellcolor{gray!15}\textbf{98.29} & \cellcolor{gray!15}\textbf{90.34} & \cellcolor{gray!15}\textbf{70.44} & \cellcolor{gray!15}\textbf{35.57} \\
    \bottomrule
    \end{tabular}%
  \label{tab:table4}%
  }
\end{table}%

\begin{figure*}[h]
\centering
\includegraphics[width=\linewidth]{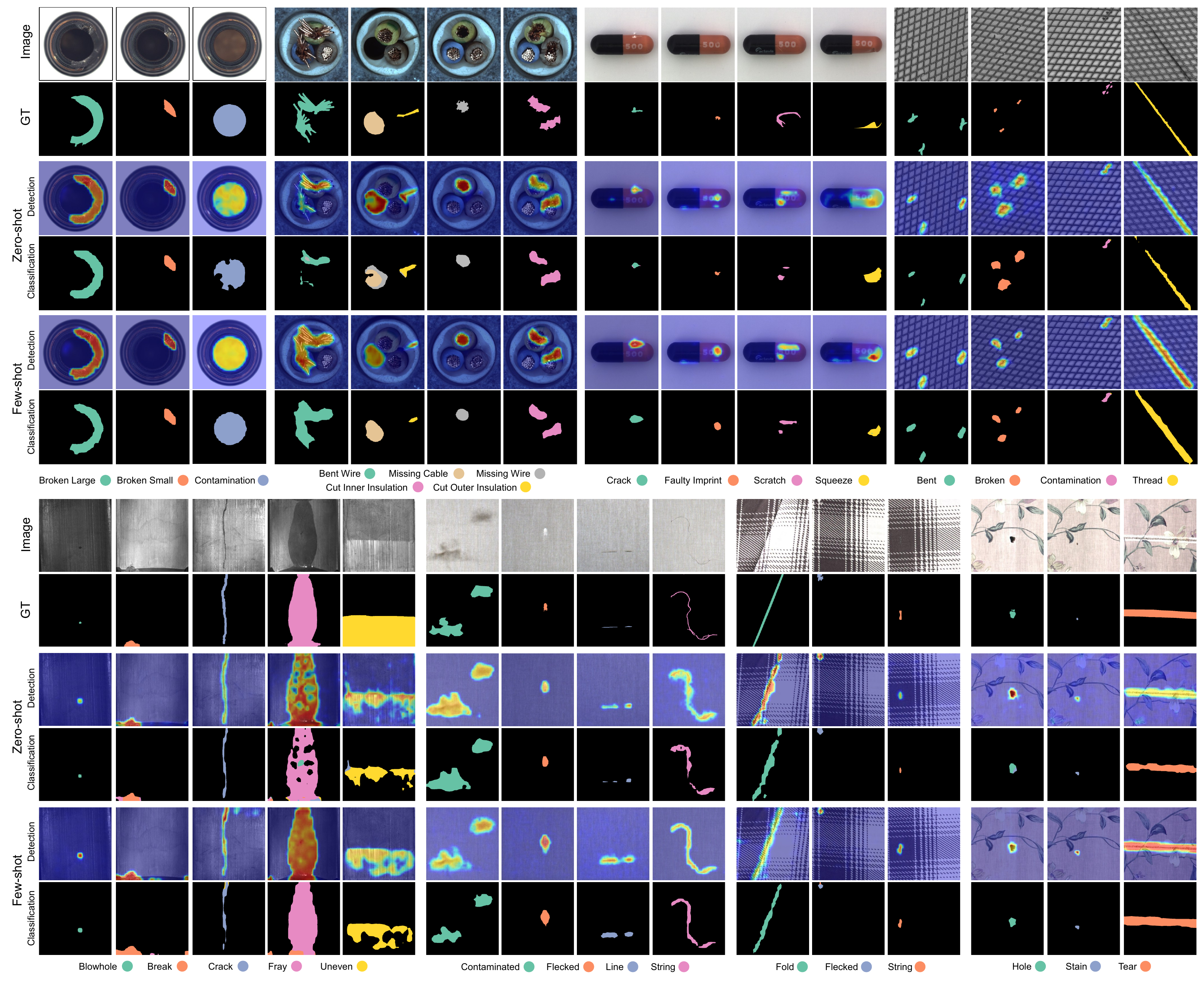}
\caption{Qualitative results of UniADC under $\text{zero-shot}_{\:(K_n = 2)}$ and $\text{few-shot}_{\:(K_n = 2,\,K_a = 1)}$ settings.}
\label{fig:fig6}
\end{figure*}

\subsection{Experimental Results}

\textit{1) Anomaly Detection and Classification Results.} Tables \ref{tab:table2} and \ref{tab:table3} report the anomaly detection and classification results under zero-shot and few-shot settings on the MVTec-FS, MTD, and WFDD datasets, respectively. In the zero-shot setting, most baselines struggle to identify specific anomaly categories due to the absence of real anomaly samples. Although some methods incorporate Vision-Language Models (VLMs) or Large Language Models (LLMs) with anomaly priors, they still fail to achieve satisfactory performance. In contrast, UniADC fully exploits the relevance between anomaly detection and classification, thereby achieving promising performance in both tasks. Compared to other methods, it achieves approximately a 20\% improvement in classification accuracy and a 10\% improvement in mIoU. Moreover, \textbf{UniADC{\fontsize{7}{7}\selectfont \texttt{(DINO)}}} achieves superior detection and classification accuracy over \textbf{UniADC{\fontsize{7}{7}\selectfont \texttt{(CLIP)}}}, highlighting the remarkable performance of DINOv3 \cite{simeoni2025dinov3} in extracting fine-grained image features. In the few-shot setting, UniADC delivers substantial performance gains, even with only one anomaly sample per category, demonstrating its low cost and effective data utilization. Furthermore, it markedly outperforms other two-stage methods, showcasing the benefit of unifying anomaly detection and classification. Notably, SegDINO \cite{yang2025segdino} underperforms in anomaly classification due to severe pixel-level class imbalance, and further fails to leverage additional normal samples effectively. Table \ref{tab:table4} reports the performance of UniADC and competing methods on the Real-IAD dataset \cite{wang2024real} under the few-shot setting. On the more challenging Real-IAD dataset, UniADC achieves state-of-the-art performance across all metrics, surpassing the second-best method by more than 10\% in classification accuracy. Qualitative results in Fig. \ref{fig:fig6} further illustrate its ability to accurately localize and classify anomalous regions under various settings, emphasizing its strong generalization capability and practical applicability. 

\begin{table}[t]
  \centering
  \renewcommand\arraystretch{0.75}
  \caption{Comparison of UniADC with alternative methods on the MVTec-FS dataset using full-shot normal samples, with $K_a=1$.}
  \resizebox{0.6\linewidth}{!}{
    \small
    \begin{tabular}{c|ccc|cc}
    \toprule
    \multirow{2}[1]{*}{Method} & \multicolumn{3}{c|}{Detection} & \multicolumn{2}{c}{Classification} \\
\cmidrule{2-6}    \multicolumn{1}{c|}{} & \multicolumn{1}{c}{I-AUC} & \multicolumn{1}{c}{P-AUC} & \multicolumn{1}{c|}{PRO} & \multicolumn{1}{c}{Acc} & \multicolumn{1}{c}{mIoU} \\
    \midrule
    PatchCore{\fontsize{7}{7}\selectfont +ZipAF} \cite{roth2022towards} & 98.77 & 98.76 & \textbf{95.17} & 69.50  & 39.60 \\
    RD4AD{\fontsize{7}{7}\selectfont +ZipAF} \cite{deng2022anomaly} & 98.89 & 98.54 & 94.23 & 62.23 & 39.76 \\
    RealNet{\fontsize{7}{7}\selectfont +ZipAF} \cite{zhang2024realnet} & \textbf{99.60}  & \textbf{99.02} & 93.71 & 67.20  & 43.24 \\
    BGAD{\fontsize{7}{7}\selectfont +ZipAF} \cite{yao2023explicit} &   98.80   &  98.24   &   92.33   &  65.62  &  41.93 \\
    \cellcolor{gray!15}\textbf{UniADC{\fontsize{7}{7}\selectfont \texttt{(CLIP)}}} & \cellcolor{gray!15}98.85 & \cellcolor{gray!15}98.73 & \cellcolor{gray!15}93.76 & \cellcolor{gray!15}87.01 & \cellcolor{gray!15}51.59 \\
    \cellcolor{gray!15}\textbf{UniADC{\fontsize{7}{7}\selectfont \texttt{(DINO)}}} & \cellcolor{gray!15}99.41 & \cellcolor{gray!15}98.95 & \cellcolor{gray!15}94.06 & \cellcolor{gray!15}\textbf{87.86} & \cellcolor{gray!15}\textbf{51.93} \\
    \bottomrule
    \end{tabular}%
    }
  \label{tab:table5}%
\end{table}%

\begin{table*}[!t]
  \centering
     \renewcommand\arraystretch{0.7}
     \caption{Comparison of UniADC with other anomaly synthesis methods on the MVTec-FS dataset, using DINOv3 as the vision backbone.}
     \resizebox{0.77\linewidth}{!}{
     \small
      \begin{tabular}{c|c|ccc|cc|cc}
    \toprule
    \multicolumn{1}{c|}{\multirow{2}[1]{*}{Setting}} & \multirow{2}[1]{*}{Method} & \multicolumn{3}{c|}{Detection} & \multicolumn{2}{c|}{Classification} & \multicolumn{2}{c}{Image Quality } \\
    \cmidrule{3-9}          & \multicolumn{1}{c|}{} & \multicolumn{1}{c}{I-AUC} & \multicolumn{1}{c}{P-AUC} & \multicolumn{1}{c|}{PRO} & \multicolumn{1}{c}{Acc} & \multicolumn{1}{c|}{mIoU} & \multicolumn{1}{c}{IS $\uparrow$} & \multicolumn{1}{c}{IC-L $\uparrow$} \\
    \midrule
    \multicolumn{1}{c|}{\multirow{6}[1]{*}{\makecell{Zero-shot \\ $\mathsmallish{(K_n=2)}$}}} & CutPaste \cite{li2021cutpaste} & 94.15  & 92.60 & 78.31 & \multicolumn{1}{c}{-} & \multicolumn{1}{c|}{-} & 1.35  & 0.11 \\
          & NSA \cite{schluter2022natural}  & 95.06    & 93.70 & 77.20 & \multicolumn{1}{c}{-} & \multicolumn{1}{c|}{-} & 1.50   & 0.14 \\
          & DR\AE M \cite{zavrtanik2021draem} & 95.90  & 95.71  & 84.48 & \multicolumn{1}{c}{-} & \multicolumn{1}{c|}{-} & 1.53  & 0.16 \\
          & RealNet \cite{zhang2024realnet} & 95.83 & 96.11 & 85.76 & \multicolumn{1}{c}{-} & \multicolumn{1}{c|}{-} & 1.58  & 0.14 \\
          & AnomalyAny \cite{sun2025unseen} & 96.08 & 95.78 & 88.15 & 54.07 & 25.46	 & 1.65  & 0.18 \\
          & \cellcolor{gray!15}\textbf{UniADC} & \cellcolor{gray!15}\textbf{97.09}  & \cellcolor{gray!15}\textbf{97.04} & \cellcolor{gray!15}\textbf{92.15} & \cellcolor{gray!15}\textbf{74.74} & \cellcolor{gray!15}\textbf{36.66} & \cellcolor{gray!15}\textbf{1.70} & \cellcolor{gray!15}\textbf{0.21} \\
    \midrule
    \multicolumn{1}{c|}{\multirow{6}[1]{*}{\makecell{Few-shot \\ { $\mathsmallish{(K_n=2,K_a=1)}$}}}} & \textit{w/o Synthesis} & 96.28 & 97.45 & 84.84 & 77.40 & 42.23 & \multicolumn{1}{c}{-} & \multicolumn{1}{c}{-} \\
          & \textit{CutPasteByMask} & 97.38 & 98.00 & 88.96 & 82.41 & 47.07 & 1.35  &  0.07 \\
          & AnoGen \cite{gui2024few} & 97.69  & 98.52 & 88.79  & 85.74 & 50.34 & 1.39  & 0.15 \\
          & DualAnoDiff \cite{jin2025dual} & 96.50 & 98.47 & 89.62 & 85.61& 50.71 & 1.48  & \textbf{0.26} \\
          & AnomalyDiffusion \cite{hu2024anomalydiffusion} & 97.16 & 98.21 & 90.16  & 83.07 & 50.82 & 1.50   & 0.21 \\
          & \cellcolor{gray!15}\textbf{UniADC} & \cellcolor{gray!15}\textbf{98.56} & \cellcolor{gray!15}\textbf{98.90} & \cellcolor{gray!15}\textbf{92.48} & \cellcolor{gray!15}\textbf{86.85} & \cellcolor{gray!15}\textbf{51.49} & \cellcolor{gray!15}\textbf{1.75}  & \cellcolor{gray!15}0.24 \\
    \bottomrule
    \end{tabular}%
    }
  \label{tab:table6}%
\end{table*}%

\textit{2) Experimental Results with Full-shot Normal Samples.} In addition, UniADC can be extended to the setting of full-shot normal samples, with the results provided in Table \ref{tab:table5}. We use the unsupervised anomaly detection methods PatchCore \cite{roth2022towards}, RD4AD \cite{deng2022anomaly}, and RealNet \cite{zhang2024realnet}, as well as the semi-supervised method BGAD \cite{yao2023explicit}, in combination with the anomaly classification method ZipAF \cite{lyu2025mvrec} as our baselines. When sufficient normal samples are available, UniADC can achieve anomaly detection performance comparable to mainstream full-shot methods. Moreover, it significantly improves anomaly classification performance, addressing the limitations of existing approaches in this aspect and demonstrating its unique practical value.

\begin{figure*}[t]
\centering
\includegraphics[width=\linewidth]{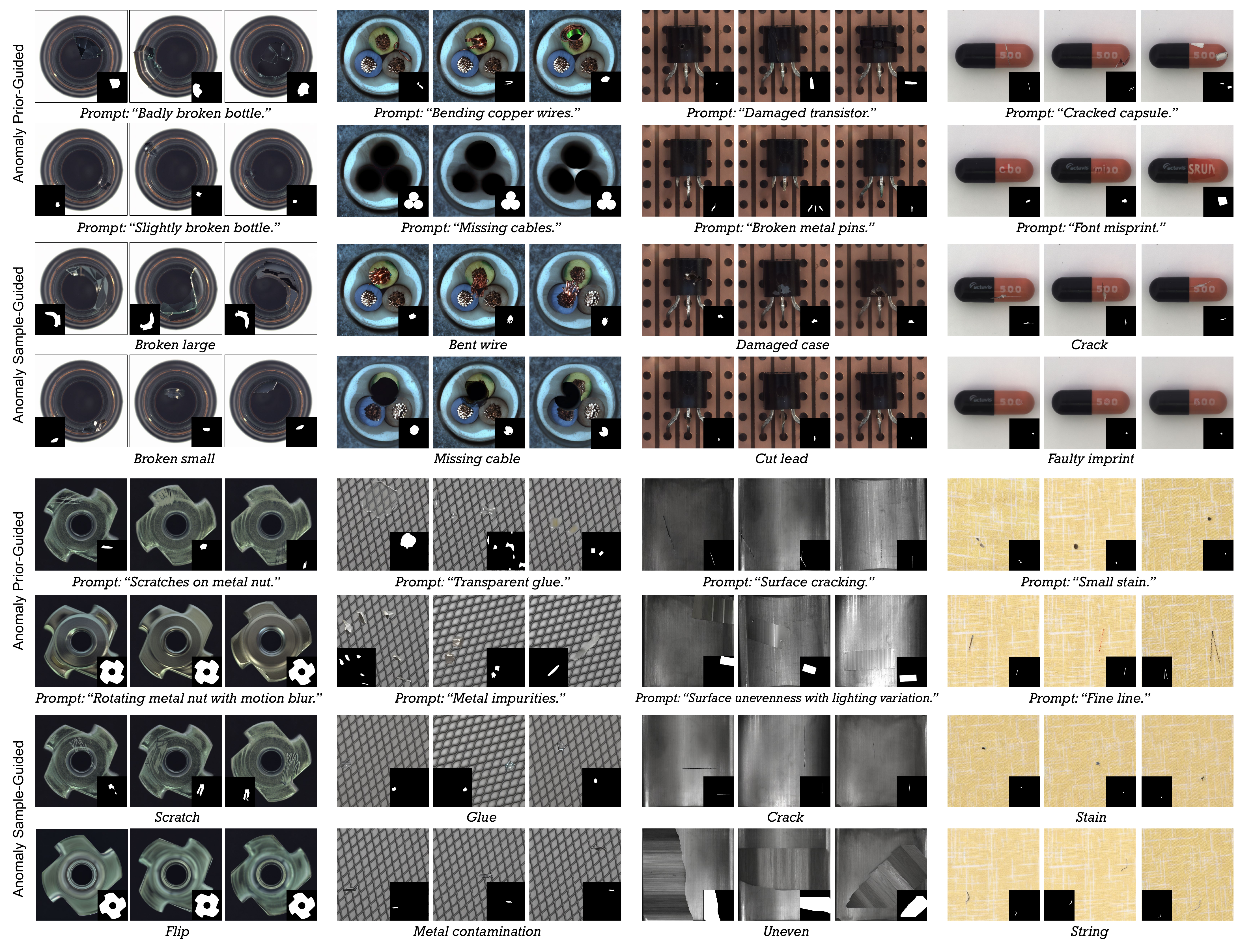}
\caption{Examples of synthetic anomaly samples generated by UniADC under the guidance of anomaly prior and anomaly sample.}
\label{fig:fig7}
\end{figure*}

\textit{3) Experimental Results on Anomaly Synthesis.} Table \ref{tab:table6} presents the performance of UniADC and other anomaly synthesis methods in terms of anomaly detection, classification, and the quality of synthetic anomaly images. For the zero-shot setting, we use CutPaste \cite{li2021cutpaste}, NSA \cite{schluter2022natural}, DR\AE M \cite{zavrtanik2021draem}, RealNet \cite{zhang2024realnet}, and AnomalyAny \cite{sun2025unseen} as baseline methods. For the few-shot setting, we use AnoGen \cite{gui2024few}, DualAnoDiff \cite{jin2025dual}, AnomalyDiffusion \cite{hu2024anomalydiffusion}, as well as the methods without anomaly synthesis (\textit{w/o Synthesis}) and CutPaste based on masks (\textit{CutPasteByMask}) as our baseline methods. We use the synthetic anomaly samples from the above methods to train the implicit-normal discriminator and evaluate their quality using the IS and IC-LPIPS metrics, as in previous works \cite{hu2024anomalydiffusion, jin2025dual}. In both settings, UniADC achieves satisfactory performance and outperforms other methods in multiple aspects. Compared to \textit{w/o Synthesis} and \textit{CutPasteByMask}, UniADC shows significant performance improvements, demonstrating the necessity of anomaly sample augmentation and image inpainting. Fig. \ref{fig:fig7} presents synthetic anomaly samples generated by UniADC. By incorporating an inpainting control network and reliable category consistency selection strategies, UniADC can generate high-quality anomaly samples that are mask-consistent, category-aligned, and visually diverse. These properties enable UniADC to generalize well across a wide range of anomaly detection and classification scenarios.

\begin{table}[t]
   \centering
  \Large
  \caption{Comparison of UniADC performance under closed-set and open-set settings with DINOv3 as the vision backbone.}
  \renewcommand\arraystretch{0.75}
\resizebox{0.65\linewidth}{!}{
    \begin{tabular}{c|ccc|ccccc}
    \toprule
    \multicolumn{9}{c}{Hazelnut} \\
    \midrule
    \multirow{2}[1]{*}{Setting} & \multicolumn{1}{c}{\multirow{2}[1]{*}{I-AUC}} & \multicolumn{1}{c}{\multirow{2}[1]{*}{P-AUC}} & \multicolumn{1}{c|}{\multirow{2}[1]{*}{PRO}} & \multicolumn{5}{c}{Acc} \\
\cmidrule{5-9}    \multicolumn{1}{c|}{} &       &       &       & \multicolumn{1}{c}{Normal } & \multicolumn{1}{c}{Crack} & \multicolumn{1}{c}{Cut } & \multicolumn{1}{c}{Hole } & \multicolumn{1}{c}{AVG} \\
    \midrule
    Closed-set & \textbf{99.63} & 97.53 & 91.29 & \textbf{97.50}  & \textbf{55.56} & \textbf{62.50}  & \textbf{88.89} & \textbf{86.36} \\
    \cellcolor{gray!15}Open-set & \cellcolor{gray!15}99.41 & \cellcolor{gray!15}\textbf{99.09} & \cellcolor{gray!15}\textbf{95.21} & \cellcolor{gray!15}95.00   & \cellcolor{gray!15}44.44 & \cellcolor{gray!15}\textbf{62.50}  & \cellcolor{gray!15}\textbf{88.89} & \cellcolor{gray!15}83.33 \\
    \midrule
    \multicolumn{9}{c}{Carpet} \\
    \midrule
    \multirow{2}[1]{*}{Setting} & \multicolumn{1}{c}{\multirow{2}[1]{*}{I-AUC}} & \multicolumn{1}{c}{\multirow{2}[1]{*}{P-AUC}} & \multicolumn{1}{c|}{\multirow{2}[1]{*}{PRO}} & \multicolumn{5}{c}{Acc} \\
\cmidrule{5-9}    \multicolumn{1}{c|}{} &       &       &       & \multicolumn{1}{c}{Normal} & \multicolumn{1}{c}{Color} & \multicolumn{1}{c}{Cut} & \multicolumn{1}{c}{Hole} & \multicolumn{1}{c}{AVG} \\
    \midrule
    Closed-set & \textbf{99.66} & \textbf{98.71} & \textbf{97.98} & 92.86 & \textbf{88.89} & 62.50  & \textbf{50.00}    & 81.13 \\
    \cellcolor{gray!15}Open-set & \cellcolor{gray!15}99.32 & \cellcolor{gray!15}98.57 & \cellcolor{gray!15}97.34 & \cellcolor{gray!15}\textbf{96.43} & \cellcolor{gray!15}77.78 & \cellcolor{gray!15}\textbf{75.00}   & \cellcolor{gray!15}\textbf{50.00}    & \cellcolor{gray!15}\textbf{83.02} \\
    \bottomrule
    \end{tabular}%
    }
  \label{tab:table7}%
\end{table}%

\begin{figure}[t]
  \centering
   \includegraphics[width=\linewidth]{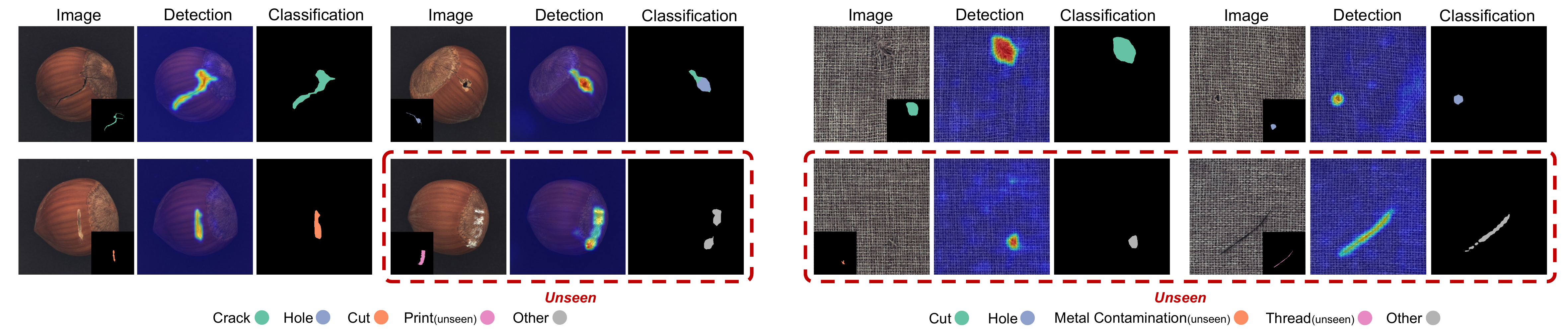}
   \caption{Qualitative results of UniADC under the setting of open-set anomaly detection and classification.}
   \label{fig:fig8}
\end{figure}

\textit{4) Open-set Anomaly Detection and Classification.} For open-set anomaly detection, a model is trained on only a subset of anomaly types and is expected to generalize to detect unseen anomalies \cite{ding2022catching}. We extend this setting to anomaly detection and classification, where a model is designed to detect both seen and unseen anomalies while ensuring correct classification of the seen anomalies. We propose a simple yet effective approach to adapt UniADC for open-set anomaly detection and classification tasks. Specifically, during training, we use unconstrained anomaly synthesis methods (e.g., DR\AE M \cite{zavrtanik2021draem}) to generate class-agnostic anomalous images and align them with an additional learnable anomaly embedding. During inference, the learnable embedding is used to match all unseen anomaly types and categorize them as "Other". 

We conducted experiments on the \textit{Hazelnut} and \textit{Carpet} classes of the MVTec-FS dataset \cite{lyu2025mvrec} to validate the effectiveness of our proposed approach. For the \textit{Hazelnut} class, \textit{"Crack"}, \textit{"Cut"}, and \textit{"Hole"} are used as seen anomalies, while \textit{"Print"} is designated as the unseen anomaly. For the \textit{Carpet} class, \textit{"Metal Contamination"} and \textit{"Thread"} are treated as unseen anomalies (external foreign objects). We adopt the zero-shot setting ($K_n=2$), where no prior knowledge about the unseen anomalies is provided during training. Table \ref{tab:table7} presents the anomaly detection and classification results of UniADC under both open-set and closed-set settings. In the open-set setting, UniADC achieves anomaly detection and classification performance comparable to the closed-set counterpart, demonstrating strong generalization to unseen anomalies without degrading the classification accuracy of seen ones. Fig. \ref{fig:fig8} presents qualitative results of UniADC, showing that it ensures accurate detection and classification of seen anomalies, while aligning unseen anomalies with the learnable anomaly embedding to identify them as the "Other" type.

\begin{table}[t]
  \centering
  \Large
  \addtolength{\tabcolsep}{-1pt}
  \caption{Comparison of computational efficiency between UniADC and other methods.}
  \renewcommand\arraystretch{1.}
  \resizebox{0.8\linewidth}{!}{
    \begin{tabular}{cccccc}
    \toprule
    \multicolumn{6}{c}{CLIP-based} \\
    \midrule
    Method & \multicolumn{1}{c}{WinCLIP \cite{jeong2023winclip}} & \multicolumn{1}{c}{WinCLIP{\fontsize{10}{10}\selectfont +ZipAF} \cite{jeong2023winclip}} & \multicolumn{1}{c}{AnomalyCLIP \cite{zhou2024anomalyclip}} & \multicolumn{1}{c}{AnomalyCLIP{\fontsize{10}{10}\selectfont +ZipAF} \cite{zhou2024anomalyclip}} & \multicolumn{1}{c}{\cellcolor{gray!15}\textbf{UniADC{\fontsize{10}{10}\selectfont \texttt{(CLIP)}}}} \\
    Images/s & 2.12  & 1.91  & 25.43 & 16.94 & \cellcolor{gray!15}\textbf{28.49} \\
    \midrule
    \multicolumn{6}{c}{DINO-based} \\
    \midrule
    Method & \multicolumn{1}{c}{AnomalyDINO \cite{damm2025anomalydino} } & \multicolumn{1}{c}{AnomalyDINO{\fontsize{10}{10}\selectfont +ZipAF} \cite{damm2025anomalydino} } & \multicolumn{1}{c}{NAGL \cite{wang2025normalabnormal}} & \multicolumn{1}{c}{NAGL{\fontsize{10}{10}\selectfont +ZipAF} \cite{wang2025normalabnormal}} & \multicolumn{1}{c}{\cellcolor{gray!15}\textbf{UniADC{\fontsize{10}{10}\selectfont \texttt{(DINO)}}}} \\
    Images/s & 21.55 & 12.95 & 38.48 & 27.33 & \cellcolor{gray!15}\textbf{44.67} \\
    \bottomrule
    \end{tabular}%
    }
  \label{tab:table8}%
\end{table}%

\begin{table}[t]
  \centering
  \Large
  \renewcommand\arraystretch{0.8}  
  \caption{Ablation results of UniADC on the MVTec-FS dataset. $\mathcal{L}_{CE}$ denotes the Cross-Entropy loss, $IND$ denotes the Implicit-Normal Discriminator, and $CCS$ refers to the Category Consistency Selection strategy.}
   \resizebox{0.8\linewidth}{!}{
    \begin{tabular}{c|c|ccccc}
    \toprule
    \multicolumn{1}{c|}{Setting}& \multicolumn{1}{c|}{Component} & \multicolumn{1}{c}{I-AUC} & \multicolumn{1}{c}{P-AUC} & \multicolumn{1}{c}{PRO} & \multicolumn{1}{c}{Acc} & \multicolumn{1}{c}{mIoU} \\
    \midrule
    \multicolumn{1}{c|}{\multirow{3}[6]{*}{\makecell{Zero-shot \\ $\mathtiny{(K_n=2)}$}}} & w/o $\mathcal{L}_{CE} $ & \textbf{97.76}\textcolor{mygreen}{$\mathttt{(0.67\uparrow)}$} & 96.23\textcolor{myred}{$\mathttt{(0.81\downarrow)}$} & 87.10\textcolor{myred}{$\mathttt{(5.05\downarrow)}$} & \multicolumn{1}{c}{-} & \multicolumn{1}{c}{-} \\
          & w/o $IND$ & 96.72\textcolor{myred}{$\mathttt{(0.37\downarrow)}$} & 95.70\textcolor{myred}{$\mathttt{(1.34\downarrow)}$} & 91.44\textcolor{myred}{$\mathttt{(0.71\downarrow)}$} & 55.91\textcolor{myred}{$\mathttt{(18.83\downarrow)}$} & 24.99\textcolor{myred}{$\mathttt{(11.67\downarrow)}$} \\
          & w/o $CCS$ & 95.19\textcolor{myred}{$\mathttt{(1.90\downarrow)}$} & 96.02\textcolor{myred}{$\mathttt{(1.02\downarrow)}$} & 89.26\textcolor{myred}{$\mathttt{(2.89\downarrow)}$} & 66.16\textcolor{myred}{$\mathttt{(8.58\downarrow)}$} & 33.15\textcolor{myred}{$\mathttt{(3.51\downarrow)}$} \\
          & \cellcolor{gray!15}\textbf{UniADC} & \cellcolor{gray!15}97.09 & \cellcolor{gray!15}\textbf{97.04} & \cellcolor{gray!15}\textbf{92.15} & \cellcolor{gray!15}\textbf{74.74} & \cellcolor{gray!15}\textbf{36.66} \\
    \midrule
    \multicolumn{1}{c|}{\multirow{3}[6]{*}{\makecell{Few-shot \\ { $\mathtiny{(K_n=2,K_a=1)}$}}}} & w/o $\mathcal{L}_{CE}$ & 97.76\textcolor{myred}{$\mathttt{(0.80\downarrow)}$} & 98.33\textcolor{myred}{$\mathttt{(0.57\downarrow)}$} & 86.26\textcolor{myred}{$\mathttt{(6.22\downarrow)}$} & \multicolumn{1}{c}{-} & \multicolumn{1}{c}{-} \\
          &  w/o $IND$ & 97.27\textcolor{myred}{$\mathttt{(1.29\downarrow)}$} &  96.92\textcolor{myred}{$\mathttt{(1.98\downarrow)}$}	& 92.37\textcolor{myred}{$\mathttt{(0.11\downarrow)}$} & 60.04\textcolor{myred}{$\mathttt{(26.81\downarrow)}$} & 30.33\textcolor{myred}{$\mathttt{(21.16\downarrow)}$} \\
          &  w/o $CCS$ & 96.84\textcolor{myred}{$\mathttt{(1.72\downarrow)}$} &  98.08\textcolor{myred}{$\mathttt{(0.82\downarrow)}$}	& 88.56\textcolor{myred}{$\mathttt{(3.92\downarrow)}$} & 79.91\textcolor{myred}{$\mathttt{(6.94\downarrow)}$} & 45.37\textcolor{myred}{$\mathttt{(6.12\downarrow)}$} \\
          & \cellcolor{gray!15}\textbf{UniADC} & \cellcolor{gray!15}\textbf{98.56} & \cellcolor{gray!15}\textbf{98.90} & \cellcolor{gray!15}\textbf{92.48} & \cellcolor{gray!15}\textbf{86.85} & \cellcolor{gray!15}\textbf{51.49} \\
    \bottomrule
    \end{tabular}%
    }
  \label{tab:table9}%
\end{table}%

\textit{5) Computational Efficiency Analysis.} Table \ref{tab:table8} presents the inference speed measurements for UniADC and other competing methods. All approaches were implemented in PyTorch framework and evaluated on a single NVIDIA RTX 4090 GPU. By seamlessly integrating anomaly detection and classification into a unified architecture, UniADC completes both tasks within a single forward pass, eliminating redundant feature extraction and yielding a significant advantage in inference efficiency. During the training phase, UniADC takes approximately 35 seconds to synthesize an anomaly image. The average training time of UniADC (including anomaly sample synthesis and discriminator training) is about 1 GPU hour. These characteristics ensure that UniADC satisfies the practical requirements for high-speed inference and rapid product iteration in industrial applications.

\textit{6)} More qualitative results and per-class metrics can be found in the \textit{Supplementary Material}.

\subsection{Ablation Study}

We conducted extensive experiments to validate the effectiveness of each component in UniADC and its hyperparameter sensitivity. By default, all ablation experiments use DINOv3 \cite{simeoni2025dinov3} as the backbone of the discriminator. 

\begin{table}[t]
  \centering
  \renewcommand\arraystretch{0.8}
  \caption{Ablation study of anomaly prior on the MVTec-FS dataset with $K_n=2$.}
  \resizebox{0.7\linewidth}{!}{
    \begin{tabular}{ccc|ccccc}
    \toprule
    Description & Shape & Size  & \multicolumn{1}{c}{I-AUC} & \multicolumn{1}{c}{P-AUC} & \multicolumn{1}{c}{PRO} & \multicolumn{1}{c}{Acc} & \multicolumn{1}{c}{mIoU} \\
    \midrule
    \xmark     & \xmark     & \xmark     & 96.04 &  95.82  & 88.22 & 65.31 & 31.76 \\
    \cmark   & \multicolumn{1}{c}{\xmark} & \multicolumn{1}{c|}{\xmark} & 96.75 & 96.24 & 91.03 & 69.58 & 34.54 \\
    \xmark     & \cmark     & \xmark     & 96.28 &  96.00   & 89.86 & 66.02 & 32.56\\
    \xmark     & \xmark     & \cmark     & 96.49 &  96.29   & 90.69 & 67.08 & 33.26 \\
    \cmark     & \cmark     & \multicolumn{1}{c|}{\xmark} & 96.97 & 96.56 & 91.73 & 70.84 & 34.79 \\
    \cmark     & \multicolumn{1}{c}{\xmark} & \cmark     & \textbf{97.12} & 96.80  & 91.58  & 72.69 & 35.17 \\
    \cellcolor{gray!20}\cmark     & \cellcolor{gray!20}\cmark     & \cellcolor{gray!20}\cmark     &   \cellcolor{gray!20}97.09  & \cellcolor{gray!20}\textbf{97.04} & \cellcolor{gray!20}\textbf{92.15} & \cellcolor{gray!20}\textbf{74.74} & \cellcolor{gray!20}\textbf{36.66} \\
    \bottomrule
    \end{tabular}%
    }
  \label{tab:table10}%
\end{table}%

\textit{1) Ablation Study on UniADC Components.} Table \ref{tab:table9} investigates the impact of various UniADC components on its performance. When the anomaly classification loss is removed, UniADC degenerates into a standard anomaly detection model. In the zero-shot setting, image-level anomaly detection performance improves slightly, while anomaly localization performance degrades. In the few-shot setting, all metrics decline, indicating that incorporating the anomaly classification task does not compromise anomaly detection but rather enhances the robustness and accuracy of anomaly detection. In the absence of the implicit-normal discriminator, we follow \cite{sadikaj2025multiads} to directly align image patch features with the normal class and $Y$ anomaly classes (totaling $Y+1$ classes), while employing Focal Loss \cite{lin2017focal} to mitigate pixel distribution imbalance. However, the model experiences a substantial performance decline, with anomaly classification accuracy dropping by approximately 20\%. This significant degradation underscores the necessity of implicitly modeling the normal class. Moreover, UniADC exhibits consistent performance improvement across all settings with the integration of category consistency selection, highlighting its critical role in enhancing the quality of synthesized anomaly samples.

\textit{2) Ablation Study on Anomaly Prior.} Table \ref{tab:table10} investigates the impact of different types of anomaly priors on the performance of UniADC. The experimental results show that anomaly descriptions have the most significant impact on the performance of UniADC, especially in anomaly classification. For most anomaly categories, the anomaly description is a synonym of its name or a combination of its name and the detection object, such as \textit{``Foreign matter in the bottle”}. For a few anomaly categories, we introduce additional anomaly information, such as the color of the anomalous region, to aid in anomaly classification. A comprehensive list of these descriptions is provided in the \textit{Supplementary Material}. When anomaly descriptions are unavailable, we simply use anomaly category names as prompts. The introduction of anomaly descriptions enables UniADC to achieve improvements of 4.27\% in classification accuracy and 2.78\% in mIoU. Compared to anomaly descriptions, shape and size priors are generally difficult to apply to certain types of anomalies. For example, it is hard to predetermine the shapes and sizes of anomaly types such as \textit{"Scratch"} and \textit{"Crack"}. Moreover, in our experiments, the size prior yields greater performance improvements for UniADC than the shape prior. When all three types of anomaly priors are applied, UniADC achieves the best overall performance.

\begin{table}[t]
  \centering
  \Large
   \renewcommand\arraystretch{0.7}
   \caption{Ablation study of the noise factor on the MVTec-FS dataset.}
  \resizebox{0.7\linewidth}{!}{
   \footnotesize
    \begin{tabular}{c|c|ccccc}
    \toprule
    \multicolumn{1}{c|}{Setting} & $\gamma$ & \multicolumn{1}{c}{I-AUC} & \multicolumn{1}{c}{P-AUC} & \multicolumn{1}{c}{PRO} & \multicolumn{1}{c}{Acc} & \multicolumn{1}{c}{mIoU} \\
    \midrule
    \multicolumn{1}{c|}{\multirow{5}[1]{*}{\makecell{Zero-shot \\ $\mathtiny{(K_n=2)}$}}} & $(0, 0.2]$ & 93.19 & 95.76 & 85.43 & 61.46 & 30.89 \\
          & $(0.2, 0.4]$ & 96.35 & 96.78 & 91.03 & 73.67 & \textbf{36.84}\\
          & \cellcolor{gray!20}$(0.4, 0.6]$ & \cellcolor{gray!20}97.09  & \cellcolor{gray!20}\textbf{97.04} & \cellcolor{gray!20}\textbf{92.15} & \cellcolor{gray!20}\textbf{74.74} & \cellcolor{gray!20}36.66 \\
          & $(0.6, 0.8]$ & \textbf{97.14} & 96.65 & 91.58 & 74.40 & 35.81 \\
          & $(0.8, 1.0]$ & 96.83 & 96.38 & 90.38 & 72.91 & 34.83 \\
    \midrule
    \multicolumn{1}{c|}{\multirow{5}[1]{*}{ \makecell{Few-shot \\ { $\mathtiny{(K_n=2,K_a=1)}$}}  }} & $(0, 0.2]$ & 97.85 & 98.53 & 90.86 & 84.44 & 49.34 \\
          & $(0.2, 0.4]$ & \textbf{98.69} & 98.71  & 91.92 & 85.89 & 50.85 \\
          & \cellcolor{gray!20}$(0.4, 0.6]$ & \cellcolor{gray!20}98.56 & \cellcolor{gray!20}\textbf{98.90} & \cellcolor{gray!20}\textbf{92.48} & \cellcolor{gray!20}\textbf{86.85} & \cellcolor{gray!20}51.49 \\
          & $(0.6, 0.8]$ & 98.42 & 98.84 & 92.18 & 86.54 & \textbf{51.55}\\
          & $(0.8, 1.0]$ & 98.21 & 98.77 & 88.95 & 85.03 & 50.40\\
    \bottomrule
    \end{tabular}%
    }
  \label{tab:table11}%
\end{table}%

\begin{figure*}[t]
\centering
\includegraphics[width=\linewidth]{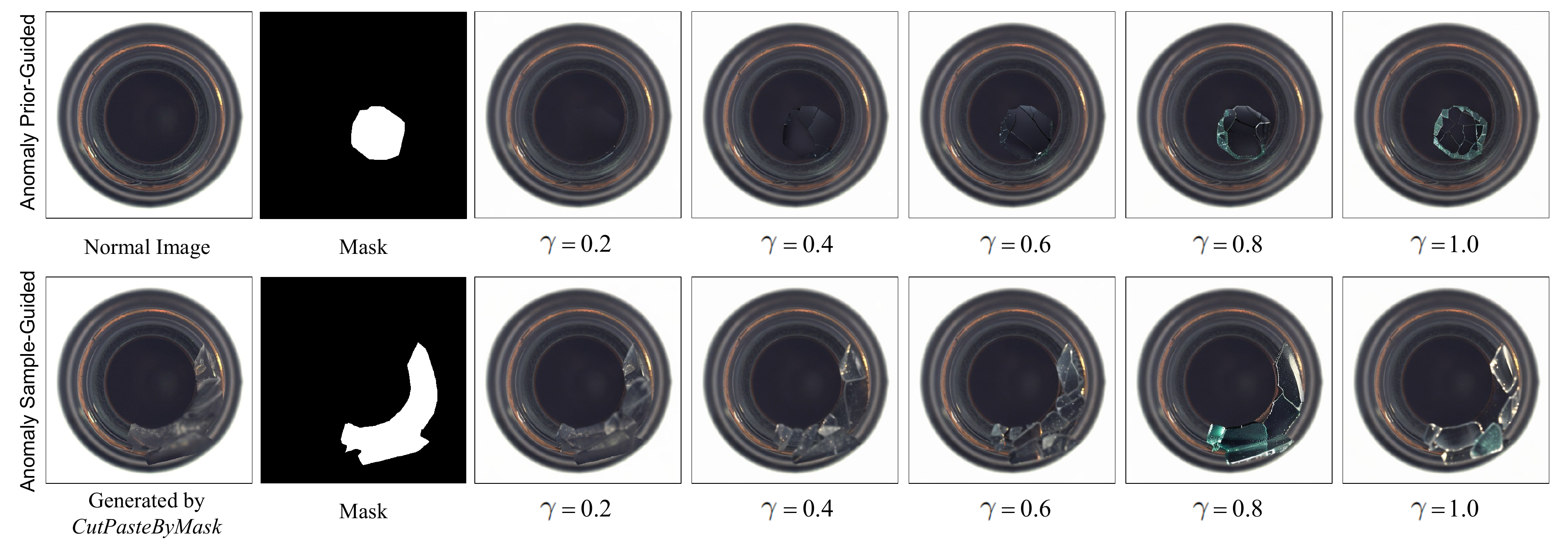}
\caption{Examples of synthetic anomaly images generated by UniADC across different noise factors, guided by anomaly prior or anomaly sample, respectively.}
\label{fig:fig9}
\end{figure*}

\begin{table}[t]
  \centering
   \renewcommand\arraystretch{0.75}
   \caption{Ablation study of the mini-batch size for category consistency selection on the MVTec-FS dataset.}
   \resizebox{0.7\linewidth}{!}{
    \footnotesize    
    \begin{tabular}{c|c|ccccc}
    \toprule
    \multicolumn{1}{c|}{Setting} & \multicolumn{1}{c|}{$\quad B \quad$} & \multicolumn{1}{c}{I-AUC} & \multicolumn{1}{c}{P-AUC} & \multicolumn{1}{c}{PRO} & \multicolumn{1}{c}{Acc} & \multicolumn{1}{c}{mIoU} \\
    \midrule
    \multicolumn{1}{c|}{\multirow{4}[1]{*}{\makecell{Zero-shot \\ $\mathtiny{(K_n=2)}$}}} & 8     & 95.93 & 96.55 & 90.55 & 71.90 & 33.17 \\
          & 16    & 96.29 & 96.80 & 90.38 & 73.12 & 35.14 \\
          & \cellcolor{gray!20}32    & \cellcolor{gray!20}\textbf{97.09} & \cellcolor{gray!20}\textbf{97.04} & \cellcolor{gray!20}92.15 & \cellcolor{gray!20}74.74 & \cellcolor{gray!20}36.66 \\
          & 64    & 97.07 & 96.98 & \textbf{92.47} & \textbf{74.91} & \textbf{36.98} \\
    \midrule
    \multicolumn{1}{c|}{\multirow{4}[1]{*}{\makecell{Few-shot \\ { $\mathtiny{(K_n=2,K_a=1)}$}}}} & 8     & 97.97 & 98.67 & 90.20 &	84.72  &  50.17\\
          & 16    & 98.04 & 98.72 & 90.49 & 85.94 &  50.63\\
          & \cellcolor{gray!20}32    & \cellcolor{gray!20}98.56 & \cellcolor{gray!20}\textbf{98.90} & \cellcolor{gray!20}92.48 & \cellcolor{gray!20}86.85 &  \cellcolor{gray!20}\textbf{51.49} \\
          & 64    & \textbf{99.02} & 98.70 & \textbf{92.81} & \textbf{86.93} &  51.12 \\
    \bottomrule
    \end{tabular}%
    }
  \label{tab:table12}%
\end{table}%

\begin{table}[t]
  \centering
    \renewcommand\arraystretch{0.75}
    \caption{Ablation study of the number of synthetic anomaly samples on the MVTec-FS dataset.}
   \resizebox{0.7\linewidth}{!}{
    \begin{tabular}{c|c|ccccc}
    \toprule
    \multicolumn{1}{c|}{Setting} & \multicolumn{1}{c|}{\makecell{Number of \\ Synthetic Samples}} & \multicolumn{1}{c}{I-AUC} & \multicolumn{1}{c}{P-AUC} & \multicolumn{1}{c}{PRO} & \multicolumn{1}{c}{Acc} & \multicolumn{1}{c}{mIoU} \\
    \midrule
    \multicolumn{1}{c|}{\multirow{5}[1]{*}{\makecell{Zero-shot \\ $\mathtiny{(K_n=2)}$}}} & 2     & 93.25 &   95.89  & 86.10 & 64.19 & 30.37 \\
          & 4 &  95.99 & 96.13 & 88.75 & 69.63 & 33.20\\
          & 8    & 96.34 & 96.85 & 91.73 & 73.55 & 35.79 \\
          & \cellcolor{gray!20}16    &  \cellcolor{gray!20}97.09   & \cellcolor{gray!20}\textbf{97.04} & \cellcolor{gray!20}\textbf{92.15} & \cellcolor{gray!20}74.74 & \cellcolor{gray!20}36.66\\
          & 32    & \textbf{97.16} & 96.93 & 92.04 & \textbf{74.82} &  \textbf{36.86} \\
    \midrule
    \multicolumn{1}{c|}{\multirow{5}[1]{*}{\makecell{Few-shot \\ { $\mathtiny{(K_n=2,K_a=1)}$}}}} & 2     & 97.25 & 98.05 & 87.90 & 83.05 & 46.70 \\
          & 4     & 98.09 & 98.05 & 91.02 & 84.94 & 49.34 \\
          & 8     & 98.27 & 98.16 & 91.53 & 85.59 & 51.03 \\
          & \cellcolor{gray!20}16    & \cellcolor{gray!20}98.56 & \cellcolor{gray!20}\textbf{98.90} & \cellcolor{gray!20}\textbf{92.48} & \cellcolor{gray!20}86.85 & \cellcolor{gray!20}51.49 \\
          & 32    & \textbf{98.72} & 98.75  & 92.20 & \textbf{86.94} & \textbf{51.80} \\
    \bottomrule
    \end{tabular}%
    }
  \label{tab:table13}%
\end{table}%

\textit{3) Ablation Study on Noise Factor.} Table \ref{tab:table11} presents the impact of the noise factor $\gamma$ on the performance of UniADC. A larger $\gamma$ increases the diversity of synthetic anomaly samples while also causing greater deviation from the original image distribution, as shown in Fig. \ref{fig:fig9}, which may undermine UniADC's ability to detect hard anomaly cases. Conversely, a smaller $\gamma$ may result in synthetic anomaly samples that do not align well with the provided anomaly description, leading to false positive anomalies or limited diversity. In our experiments, we uniformly sample $\gamma$ between 0.4 and 0.6 to ensure a balance between quality and diversity.

\begin{figure}[t]
  \centering
   \includegraphics[width=0.75\linewidth]{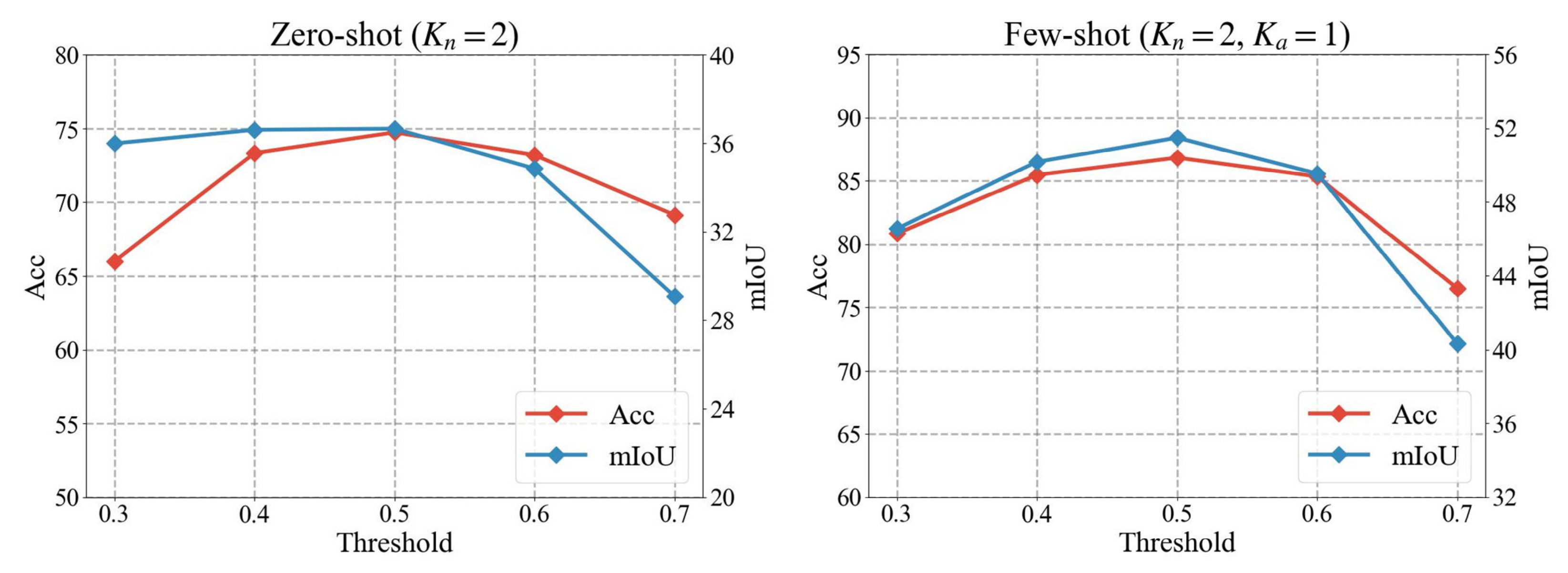}
   \caption{Ablation study of the anomaly score threshold on the MVTec-FS dataset.}
   \label{fig:fig10}
\end{figure}

\textit{4) Ablation Study on Sampling Hyperparameters.} Tables \ref{tab:table12} and \ref{tab:table13} respectively present the ablation study results on the mini-batch size in category consistency selection and the number of synthetic anomaly samples. Increasing the mini-batch size and the number of synthetic samples can improve the quality and diversity of training data, thereby enhancing the performance of UniADC. However, this also significantly increases the training cost. To ensure the performance of UniADC while reducing computational cost, we set the mini-batch size and the number of synthetic anomaly samples to 32 and 16, respectively. 

\textit{5) Ablation Study on Anomaly Score Threshold.} Fig. \ref{fig:fig10} illustrates the effect of varying the anomaly score threshold on the anomaly classification performance of UniADC. Compared to other methods \cite{jeong2023winclip, li2024promptad}, UniADC demonstrates improved robustness to threshold variations, consistently achieving reliable performance when the threshold is set between 0.4 and 0.6, which simplifies the threshold selection process and improves practical feasibility. In our experiments, we simply fixed the threshold at 0.5.

\section{Conclusion}

This paper integrates image anomaly detection and classification into a unified task and proposes UniADC, an innovative framework capable of anomaly synthesis, detection, and classification. UniADC consists of a training-free controllable inpainting network and an implicit discriminator. The former can synthesize high-quality anomaly samples of specific categories guided by anomaly priors or anomaly samples, while the latter mitigates the class imbalance between normal and anomalous pixels through implicit modeling of the normal class, achieving unified anomaly detection and classification. We conducted extensive experiments on four anomaly detection and classification benchmarks. The results demonstrate that UniADC achieves superior performance across various task settings, including zero-shot, few-shot, full-shot normal samples, as well as open-set anomaly detection. Furthermore, it features high inference efficiency and low training overhead, highlighting its potential for practical industrial scenarios.

\begin{figure}[t]
  \centering
   \includegraphics[width=0.55\linewidth]{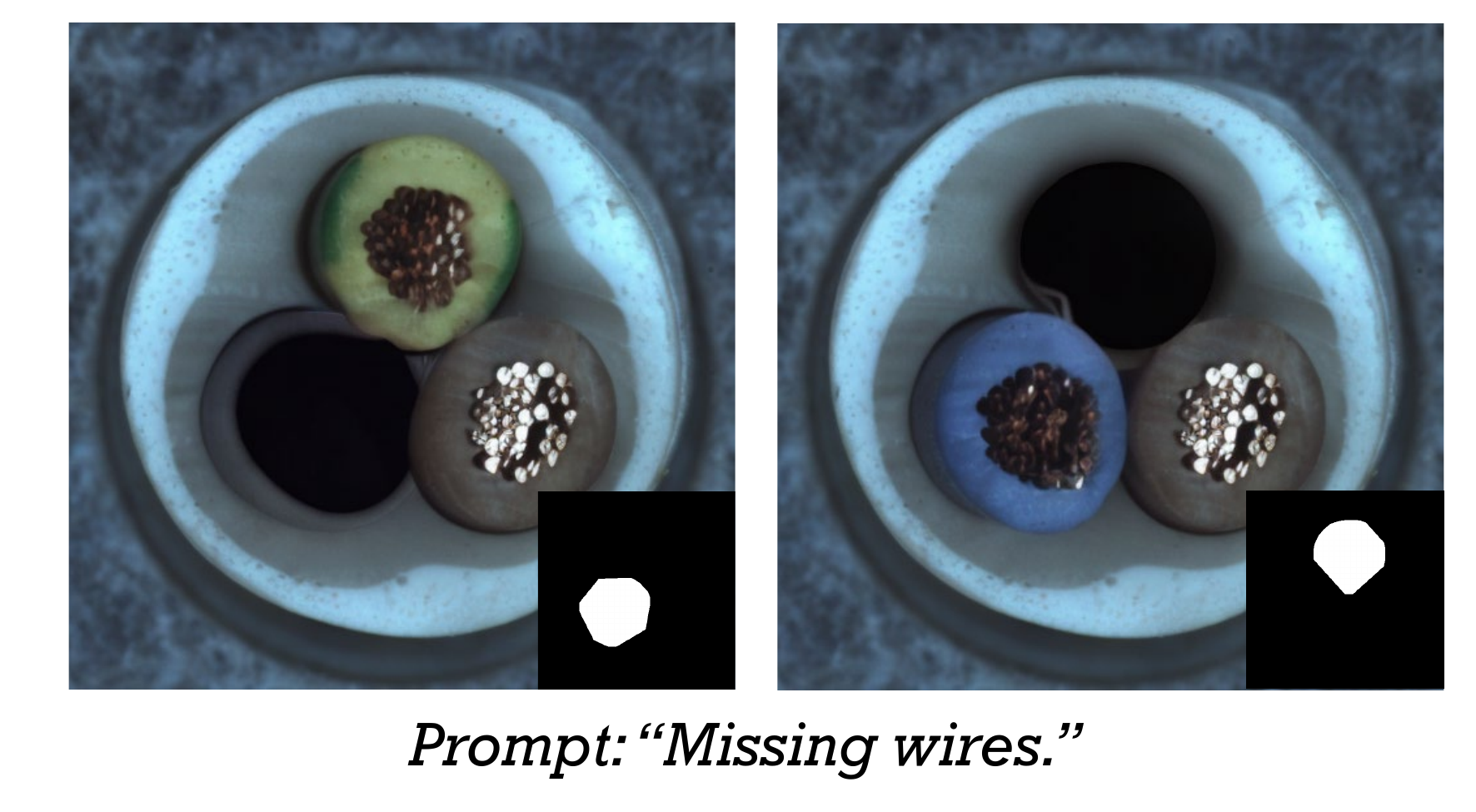}
   \caption{Representative failure cases generated by UniADC. }
   \label{fig:fig11}
\end{figure}

\textbf{Limitations and Future Work.} While UniADC employs various techniques to ensure the class fidelity of synthesized anomaly images, it may still produce erroneous results for certain semantically similar anomaly classes. Fig. \ref{fig:fig11} illustrates a representative example where UniADC confuses "Missing Wire" with "Missing Cable", which potentially hinders the discriminative model's ability to effectively distinguish between these two defects. Additionally, while GAP-Lib provides a more diverse range of geometric prototypes compared to existing zero-shot synthesis methods (which often rely on rudimentary masks like simple rectangles or Perlin noise), it may still struggle to accurately simulate anomalies with highly complex or irregular silhouettes. In future work, we aim to further enhance the synthesis accuracy and quality of our method while extending its applicability to broader domains such as medical image anomaly detection and classification.

\section*{Acknowledgements}
This work was funded in part by the National Natural Science Foundation of China under Grants 62576049 and 61972046.

\bibliography{references}

\clearpage

\renewcommand*{\thefigure}{S\arabic{figure}}
\renewcommand*{\thetable}{S\arabic{table}}
\renewcommand*{\theequation}{S\arabic{equation}}
\renewcommand*{\thealgorithm}{S\arabic{algorithm}}
\renewcommand\thesection{\Alph{section}}
\setcounter{table}{0}
\setcounter{figure}{0}
\setcounter{equation}{0}
\setcounter{section}{0}
\setcounter{algorithm}{0}

\begin{table*}[t]
  \centering
  \renewcommand\arraystretch{1.0}
  \caption{Anomaly prior list provided for each anomaly category.}
     \resizebox{\linewidth}{!}{
     \Large
    \begin{tabular}{c|c|c|l|c|c!{\vrule width 1pt}c|c|c|l|c|c}
    \bottomrule
    Dataset & \multicolumn{1}{c|}{\makecell{Image \\ Class}} & \multicolumn{1}{c|}{\makecell{Anomaly \\ Category}} & Description & Shape & Size  & Dataset & \multicolumn{1}{c|}{\makecell{Image \\ Class}} & \multicolumn{1}{c|}{\makecell{Anomaly \\ Category}} & Description & Shape & Size \\
    \hline
    \multicolumn{1}{c|}{\multirow{44}[80]{*}{MVTec-FS}} & \multicolumn{1}{c|}{\multirow{3}[1]{*}{Bottle}} & Broken large & Badly broken bottle &       & \multicolumn{1}{c!{\vrule width 1pt }}{Large} & \multicolumn{1}{c|}{\multirow{28}[1]{*}{MVTec-FS}} & \multicolumn{1}{c|}{\multirow{5}[1]{*}{Screw}} & \multicolumn{1}{c|}{\makecell{Manipulated \\ front}} & \multicolumn{1}{l|}{Damaged screw tip} &       & \multicolumn{1}{c}{   } \\
\cline{3-6}\cline{9-12}          &       & Broken small & Slightly broken bottle &       & \multicolumn{1}{c!{\vrule width 1pt }}{Small} & \multicolumn{1}{c|}{} &       & \multicolumn{1}{c|}{Scratch head} & \multicolumn{1}{l|}{Damaged screw head} &       &  \\
\cline{3-6}\cline{9-12}          &       & Contamination & Foreign matter in the bottle &       &       & \multicolumn{1}{c|}{} &       & \multicolumn{1}{c|}{Scratch neck} & \multicolumn{1}{l|}{Damaged screw neck} &       &  \\
\cline{2-6}\cline{9-12}          & \multicolumn{1}{c|}{\multirow{7}[1]{*}{Cable}} & Bent wire & Bending copper wires &       &       & \multicolumn{1}{c|}{} &       & \multicolumn{1}{c|}{Thread side} & \multicolumn{1}{l|}{Screw thread stripping} &       &  \\
\cline{3-6}\cline{9-12}          &       & Cable swap & Cable swap & \multicolumn{1}{c|}{Foreground} &       & \multicolumn{1}{c|}{} &       & \multicolumn{1}{c|}{Thread top} & \multicolumn{1}{l|}{Damaged screw thread} &       &  \\
\cline{3-6}\cline{8-12}          &       & \multicolumn{1}{c|}{\makecell{Cut inner \\ insulation}} & \makecell[l]{Cracks in the internal \\ insulation} &       &       & \multicolumn{1}{c|}{} & \multicolumn{1}{c|}{\multirow{5}[1]{*}{Tile}} & \multicolumn{1}{c|}{Crack} & \multicolumn{1}{l|}{Cracks on the tile} &       &  \\
\cline{3-6}\cline{9-12}          &       & \makecell{Cut outer \\ insulation} & \makecell[l]{Cracks in the external \\ insulation} &       &       & \multicolumn{1}{c|}{} &       & \multicolumn{1}{c|}{Glue strip} & \multicolumn{1}{l|}{Transparent glue} &       &  \\
\cline{3-6}\cline{9-12}          &       & Missing cable & Missing cables & \multicolumn{1}{c|}{Foreground} &       & \multicolumn{1}{c|}{} &       & \multicolumn{1}{c|}{Gray stroke} & \multicolumn{1}{l|}{Gray-black stain} &       & \multicolumn{1}{l}{Large} \\
\cline{3-6}\cline{9-12}          &       & Missing wire & Missing wires &       &       & \multicolumn{1}{c|}{} &       & \multicolumn{1}{c|}{Oil} & \multicolumn{1}{l|}{Oil stain} &       &  \\
\cline{3-6}\cline{9-12}          &       & Poke insulation & Puncture-induced holes &       &       & \multicolumn{1}{c|}{} &       & \multicolumn{1}{c|}{Rough} & \multicolumn{1}{l|}{Rough white tile} &       & \multicolumn{1}{l}{Large} \\
\cline{2-6}\cline{8-12}          & \multicolumn{1}{c|}{\multirow{5}[1]{*}{Capsule}} & Crack & Cracked capsule &       &       & \multicolumn{1}{c|}{} & \multicolumn{1}{c|}{Toothbrush} & \multicolumn{1}{c|}{Defective} & \multicolumn{1}{l|}{\makecell[l]{ Missing bristles,\\ Bending bristle, \\ Foreign matter, \\ A piece of thread}} &       &  \\
\cline{3-6}\cline{8-12}          &       & Faulty imprint & Font misprint, Font error &       &       & \multicolumn{1}{c|}{} & \multicolumn{1}{c|}{\multirow{4}[1]{*}{Transistor}} & \multicolumn{1}{c|}{Bent lead} & \multicolumn{1}{l|}{Bending metal pins} &       &  \\
\cline{3-6}\cline{9-12}          &       & Poke  & A tiny hole on the capsule & \multicolumn{1}{c|}{Ellipse} & \multicolumn{1}{c!{\vrule width 1pt }}{Small} & \multicolumn{1}{c|}{} &       & \multicolumn{1}{c|}{Cut lead} & \multicolumn{1}{l|}{Broken metal pins} &       &  \\
\cline{3-6}\cline{9-12}          &       & Scratch & Slight scratches on the capsule &       &       & \multicolumn{1}{c|}{} &       & \multicolumn{1}{c|}{Damaged case} & \multicolumn{1}{l|}{Damaged transistor} &       &  \\
\cline{3-6}\cline{9-12}          &       & Squeeze & \makecell[l]{A capsule deformed by \\ squeezing} &       & \multicolumn{1}{c!{\vrule width 1pt }}{Large} & \multicolumn{1}{c|}{} &       & \multicolumn{1}{c|}{Misplaced} & \multicolumn{1}{l|}{Misplaced transistor} & \multicolumn{1}{c|}{Foreground} &  \\
\cline{2-6}\cline{8-12}          & \multicolumn{1}{c|}{\multirow{6}[1]{*}{Carpet}} & Color & Red stain, Black stain &       &       & \multicolumn{1}{c|}{} & \multicolumn{1}{c|}{\multirow{4}[1]{*}{Wood}} & \multicolumn{1}{c|}{Color} & \multicolumn{1}{l|}{Red stain, Black stain} &       &  \\
\cline{3-6}\cline{9-12}          &       & Cut   & A tiny cut in carpet &       &       & \multicolumn{1}{c|}{} &       & \multicolumn{1}{c|}{Hole} & \multicolumn{1}{l|}{Hole in wood} & \multicolumn{1}{c|}{Ellipse} &  \\
\cline{3-6}\cline{9-12}          &       & Hole  & Hole  & \multicolumn{1}{c|}{Ellipse} &       & \multicolumn{1}{c|}{} &       & \multicolumn{1}{c|}{Liquid} & \multicolumn{1}{l|}{Transparent liquid stains} &       &  \\
\cline{3-6}\cline{9-12}          &       & \makecell{Metal \\ contamination} & \makecell[l]{Metal impurities,\\ Metal foreign matter} &       &       & \multicolumn{1}{c|}{} &       & \multicolumn{1}{c|}{Scratch} & \multicolumn{1}{l|}{Scratches in wood} &       &  \\
\cline{3-6}\cline{8-12}          &       & Thread & A piece of thread & \multicolumn{1}{c|}{\makecell{  Line, \\ Hollow \\Ellipse}} & \multicolumn{1}{c!{\vrule width 1pt}}{\makecell{Small, \\ Medium}} & \multicolumn{1}{c|}{} & \multicolumn{1}{c|}{\multirow{7}[1]{*}{Zipper}} & \multicolumn{1}{c|}{Broken teeth} & \multicolumn{1}{l|}{Damage to the zipper} &       &  \\
\cline{2-6}\cline{9-12}          & \multicolumn{1}{c|}{\multirow{6}[1]{*}{Grid}} & Bent  & Bending metal grid &       &       & \multicolumn{1}{c|}{} &       & \multicolumn{1}{c|}{Fabric border} & \multicolumn{1}{l|}{Border damage of fabric} &       &  \\
\cline{3-6}\cline{9-12}          &       & Broken & Broken defect &       &       & \multicolumn{1}{c|}{} &       & \multicolumn{1}{c|}{Fabric interior} & \multicolumn{1}{l|}{Interior damage of fabric} &       &  \\
\cline{3-6}\cline{9-12}          &       & Glue  & Transparent glue &       &       & \multicolumn{1}{c|}{} &       & \multicolumn{1}{c|}{Rough} & \multicolumn{1}{l|}{Wear on the zipper} &       &  \\
\cline{3-6}\cline{9-12}          &       & \makecell{Metal \\ contamination} & \makecell[l]{Metal impurities, \\ Metal foreign matter} &       &       & \multicolumn{1}{c|}{} &       & \multicolumn{1}{c|}{Split teeth} & \multicolumn{1}{l|}{Unzipped zipper} &       &  \\
\cline{3-6}\cline{9-12}          &       & Thread & A piece of thread & \multicolumn{1}{c|}{\makecell{ Line, \\ Hollow \\ Ellipse}} & \multicolumn{1}{c!{\vrule width 1pt }}{\makecell{Small,\\  Medium}} & \multicolumn{1}{c|}{} &       & \multicolumn{1}{c|}{Squeezed teeth} & \multicolumn{1}{l|}{Squeezed zipper} &       &  \\
\cline{2-12}          & \multicolumn{1}{c|}{\multirow{4}[1]{*}{Hazelnut}} & Crack & A hazelnut with cracks &       &       & \multicolumn{1}{c|}{\multirow{13}[1]{*}{WFDD}} & \multicolumn{1}{c|}{\multirow{4}[1]{*}{Grey Cloth}} & \multicolumn{1}{c|}{Contaminated} & \multicolumn{1}{l|}{Black stain} &       &  \\
\cline{3-6}\cline{9-12}          &       & Cut   & A hazelnut with cutting marks &       &       & \multicolumn{1}{c|}{} &       & \multicolumn{1}{c|}{Flecked} & \multicolumn{1}{l|}{White spot} & \multicolumn{1}{c|}{Ellipse} &  Small \\
\cline{3-6}\cline{9-12}          &       & Hole  & A hazelnut with a hole & \multicolumn{1}{c|}{Ellipse} &       & \multicolumn{1}{c|}{} &       & \multicolumn{1}{c|}{Line} & \multicolumn{1}{l|}{Black linear stain} & \multicolumn{1}{c|}{Line} & \multicolumn{1}{c}{Small} \\
\cline{3-6}\cline{9-12}          &       & Print & A hazelnut with printed text &       &       & \multicolumn{1}{c|}{} &       & \multicolumn{1}{c|}{String} & \multicolumn{1}{l|}{Fine line} &       & \multicolumn{1}{c}{Small} \\
\cline{2-6}\cline{8-12}          & \multicolumn{1}{c|}{\multirow{5}[1]{*}{Leather}} & Color & Red stain, Black stain &       &       & \multicolumn{1}{c|}{} & \multicolumn{1}{c|}{\multirow{3}[1]{*}{Grid Cloth}} & \multicolumn{1}{c|}{Fold} & \multicolumn{1}{l|}{Fabric with fold marks} & \multicolumn{1}{c|}{Line} &  \\
\cline{3-6}\cline{9-12}          &       & Cut   & A tiny cut in leather &       &       & \multicolumn{1}{c|}{} &       & \multicolumn{1}{c|}{Flecked} & \multicolumn{1}{l|}{White spot} & \multicolumn{1}{c|}{Ellipse} & Small \\
\cline{3-6}\cline{9-12}          &       & Fold  & Fold marks in leather &       &       & \multicolumn{1}{c|}{} &       & \multicolumn{1}{c|}{String} & \multicolumn{1}{l|}{Fine line} &       & \multicolumn{1}{c}{Small} \\
\cline{3-6}\cline{8-12}          &       & Glue  & Transparent glue &       &       & \multicolumn{1}{c|}{} & \multicolumn{1}{c|}{\multirow{3}[1]{*}{Pink Flower}} & \multicolumn{1}{c|}{Hole} & \multicolumn{1}{l|}{Hole} & \multicolumn{1}{c|}{Ellipse} &  \\
\cline{3-6}\cline{9-12}          &       & Poke  & A puncture-induced hole & \multicolumn{1}{c|}{Ellipse} & \multicolumn{1}{c!{\vrule width 1pt }}{Small} & \multicolumn{1}{c|}{} &       & \multicolumn{1}{c|}{Stain} & \multicolumn{1}{l|}{Small stain} &       & \multicolumn{1}{c}{Small} \\
\cline{2-6}\cline{9-12}          & \multicolumn{1}{c|}{\multirow{4}[1]{*}{Metal Nut}} & Bent  & Cracked metal nut &       &       & \multicolumn{1}{c|}{} &       & \multicolumn{1}{c|}{Tear} & \multicolumn{1}{l|}{Zipper-like tear} &       &  \\
\cline{3-6}\cline{8-12}          &       & Color & Red stain, Black stain &       &       & \multicolumn{1}{c|}{} & \multicolumn{1}{c|}{\multirow{3}[1]{*}{Yellow Cloth}} & \multicolumn{1}{c|}{Fold} & \multicolumn{1}{l|}{Fabric with fold marks} & \multicolumn{1}{c|}{Line} &  \\
\cline{3-6}\cline{9-12}          &       & Flip  & \makecell[l]{Rotating metal nut with \\ motion blur} & \multicolumn{1}{c|}{Foreground} &       & \multicolumn{1}{c|}{} &       & \multicolumn{1}{c|}{Stain} & \multicolumn{1}{l|}{Small stain} &       & \multicolumn{1}{c}{Small} \\
\cline{3-6}\cline{9-12}          &       & Scratch & Scratches on metal nut &       &       & \multicolumn{1}{c|}{} &       & \multicolumn{1}{c|}{String} & \multicolumn{1}{l|}{Fine line} &       & \multicolumn{1}{c}{Small} \\
\cline{2-12}          & \multicolumn{1}{c|}{\multirow{6}[1]{*}{Pill}} & Color & Red stain, Black stain &       &       & \multicolumn{2}{c|}{\multirow{6}[1]{*}{MTD}} & \multicolumn{1}{c|}{Blowhole} & \multicolumn{1}{l|}{Black hole} & \multicolumn{1}{c|}{Ellipse} & \multicolumn{1}{c}{Small} \\
\cline{3-6}\cline{9-12}          &       & Contamination & Foreign matter on the pill &       &       & \multicolumn{2}{c|}{} & \multicolumn{1}{c|}{Break} & \multicolumn{1}{l|}{Surface crumpling} &       & \multicolumn{1}{c}{Small} \\
\cline{3-6}\cline{9-12}          &       & Crack & Edge damage of the pill &       &       & \multicolumn{2}{c|}{} & \multicolumn{1}{c|}{Crack} & \multicolumn{1}{l|}{Surface cracking} &       & \multicolumn{1}{c}{Small} \\
\cline{3-6}\cline{9-12}          &       & Faulty imprint & Font misprint, Font error &       &       & \multicolumn{2}{c|}{} & \multicolumn{1}{c|}{Fray} & \multicolumn{1}{l|}{Surface abrasion} &       & \multicolumn{1}{c}{Large} \\
\cline{3-6}\cline{9-12}          &       & Pill type & Mold growth on the pill & \multicolumn{1}{c|}{Foreground} &       & \multicolumn{2}{c|}{} & \multicolumn{1}{c|}{\multirow{2}[1]{*}{Uneven}} & \multicolumn{1}{l|}{\multirow{2}[1]{*}{\makecell[l]{Surface unevenness with \\ lighting variation}}} & \multirow{2}[4]{*}{} & \multicolumn{1}{c}{\multirow{2}[1]{*}{Large}} \\
\cline{3-6}          &       & Scratch & Scratches on the pill &       &       & \multicolumn{2}{c|}{} &       &       &       &  \\
    \toprule
    \end{tabular}%
    }
  \label{tab:tables1}%
\end{table*}%

\begin{figure*}[t]
  \centering
   \includegraphics[width=\linewidth]{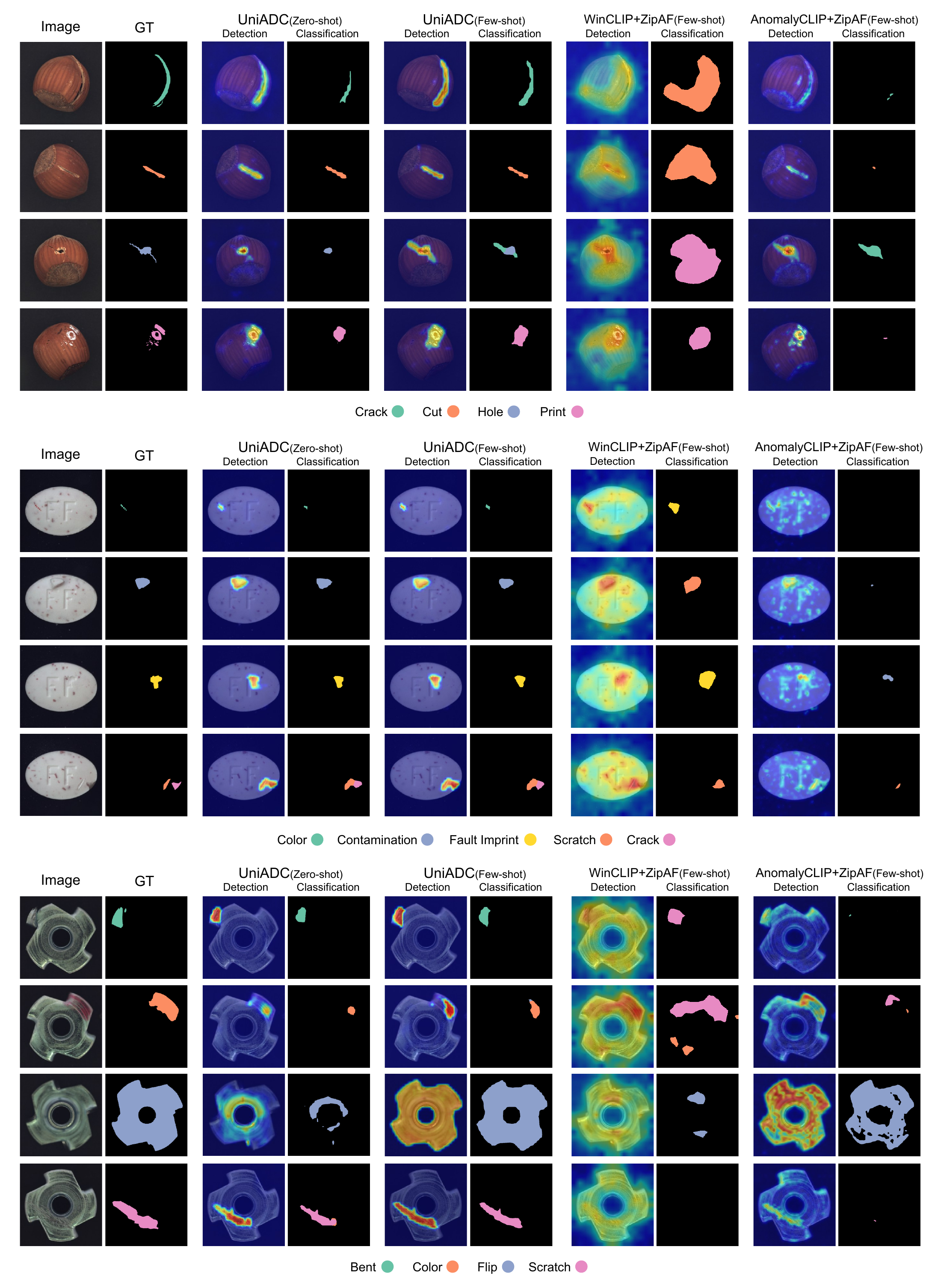}
   \vspace{-1.cm}
   \caption{Qualitative comparison of UniADC and other methods on the \textit{Hazelnut}, \textit{Pill}, and \textit{Metal Nut} classes of the MVTec-FS dataset.}
   \label{fig:figs1}
\end{figure*}

\begin{figure*}[t]
  \centering
   \includegraphics[width=\linewidth]{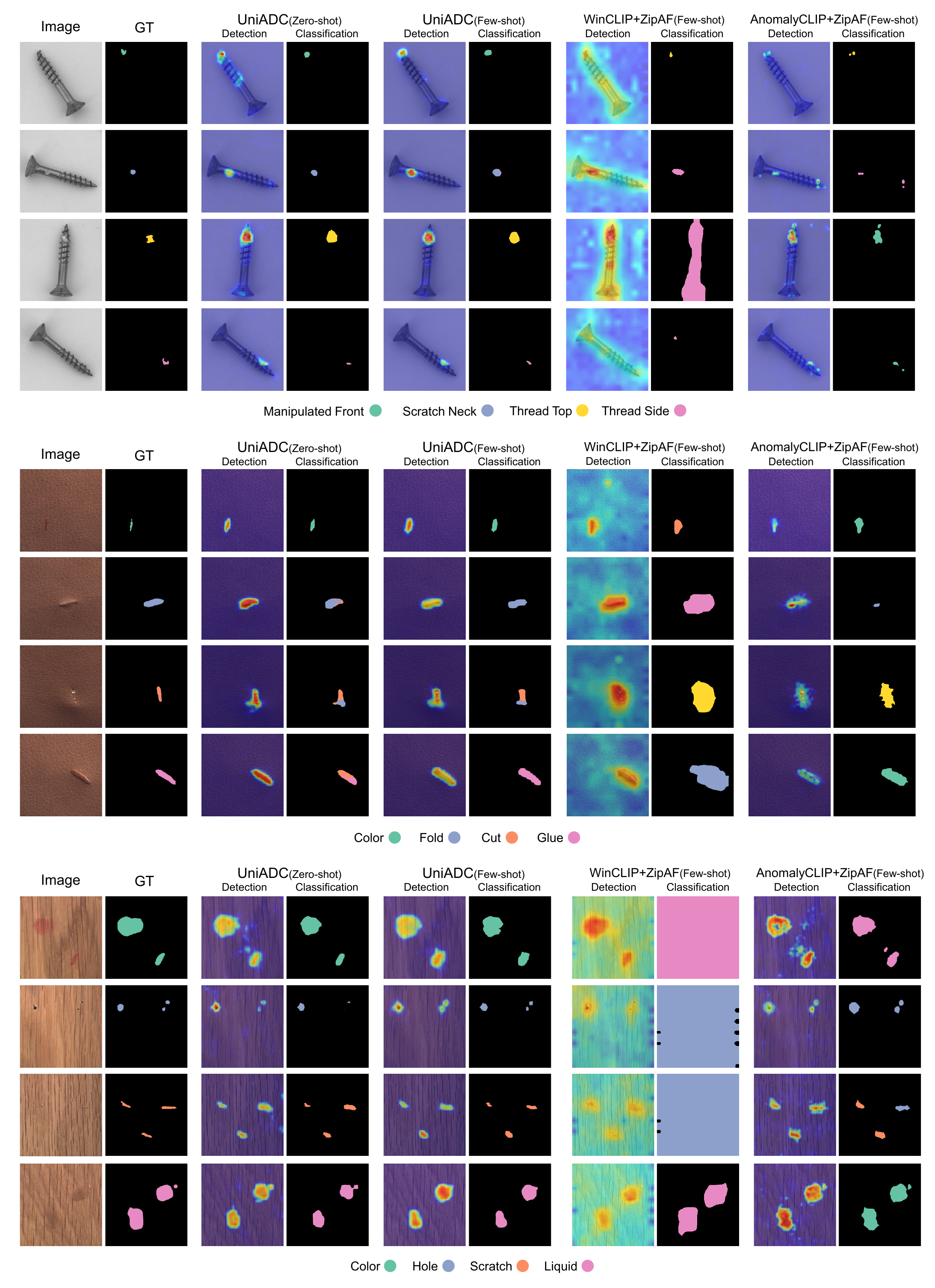}
   \vspace{-1.cm}
   \caption{Qualitative comparison of UniADC and other methods on the \textit{Screw}, \textit{Leather}, and \textit{Wood} classes of the MVTec-FS dataset.}
   \label{fig:figs2}
\end{figure*}

\begin{figure*}[t]
  \centering
   \includegraphics[width=\linewidth]{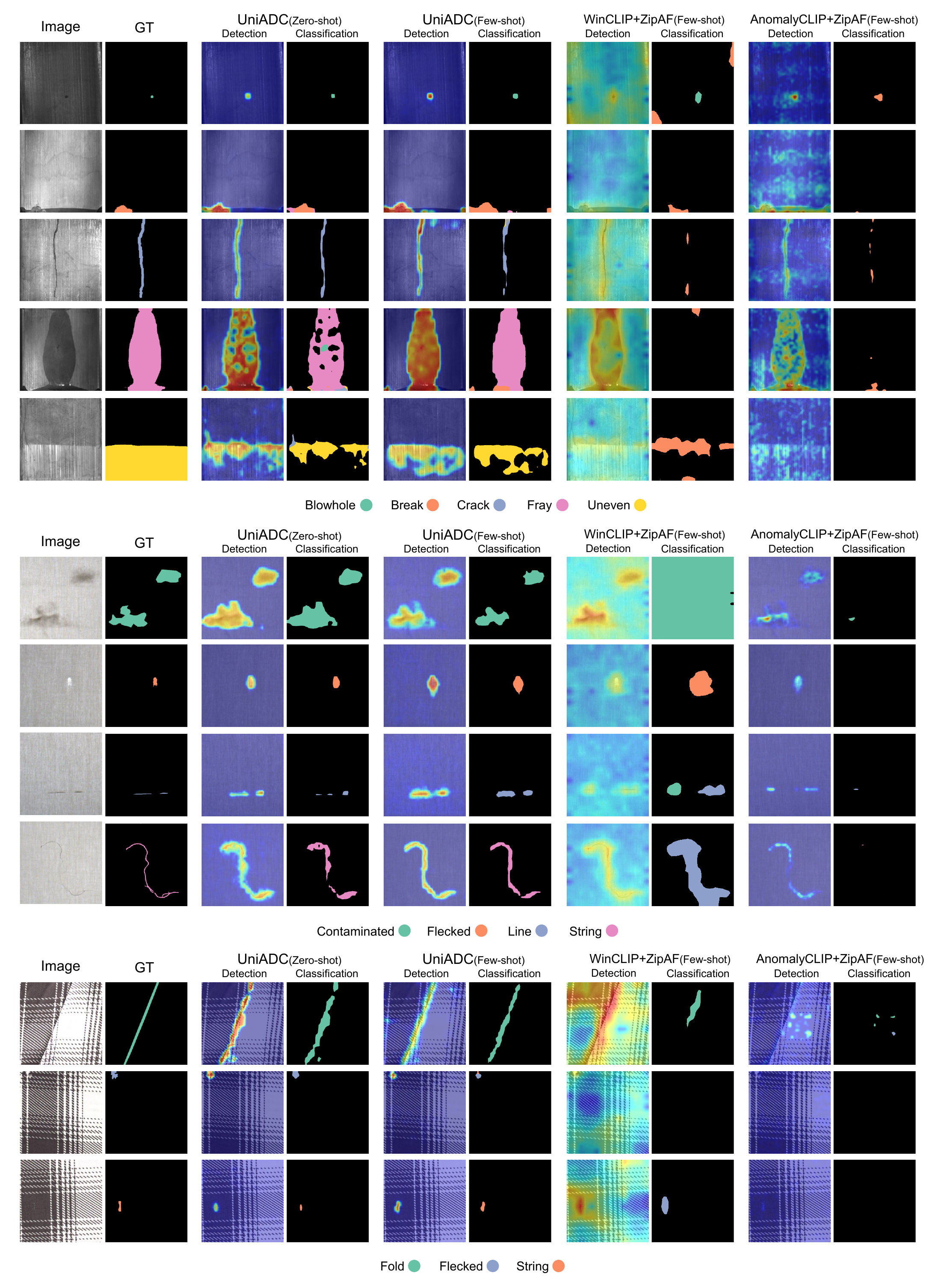}
      \vspace{-1.cm}
   \caption{Qualitative comparison of UniADC and other methods on the MTD and WFDD datasets.}
   \label{fig:figs3}
\end{figure*}

\begin{table}[t]
  \centering
  \renewcommand\arraystretch{0.6}
  \caption{Anomaly detection and classification results of \textbf{UniADC{\fontsize{7}{7}\selectfont \texttt{(CLIP)}}} on the MVTec-FS dataset under the zero-shot $(K_n=1)$ setting.}
     \resizebox{0.8\linewidth}{!}{
    \begin{tabular}{c|ccccc}
    \toprule
    \multicolumn{1}{c|}{Category} & \multicolumn{1}{c}{I-AUC} & \multicolumn{1}{c}{P-AUC} & \multicolumn{1}{c}{PRO} & \multicolumn{1}{c}{Acc} & \multicolumn{1}{c}{mIoU} \\
    \midrule
    Bottle & 99.52 & 93.85 & 85.24 & 72.55 & 42.41 \\
    Cable & 93.67 & 94.53 & 84.56 & 62.64 & 33.59 \\
    Capsule & 90.98 & 97.12 & 92.80  & 43.42 & 24.16 \\
    Carpet & 99.57 & 99.09 & 97.74 & 70.00    & 30.37 \\
    Grid  & 99.12 & 98.73 & 94.31 & 62.50  & 24.53 \\
    Hazelnut & 99.93 & 98.42 & 95.85 & 75.67 & 29.92 \\
    Leather & 100.0   & 99.85 & 98.86 & 77.63 & 39.34 \\
    Metal Nut & 99.60  & 96.69 & 78.57 & 79.10  & 40.59 \\
    Pill  & 98.30  & 98.40  & 96.27 & 56.98 & 28.41 \\
    Screw & 77.12 & 97.68 & 90.90  & 35.35 & 17.92 \\
    Tile  & 99.63 & 99.17 & 94.58 & 82.43 & 39.87 \\
    Toothbrush & 95.00    & 97.05 & 69.96 & 92.59 & 62.76 \\
    Transistor & 80.33 & 78.81 & 70.79 & 60.00    & 21.26 \\
    Wood  & 97.37 & 97.92 & 88.78 & 73.81 & 40.89 \\
    Zipper & 95.37 & 97.63 & 92.08 & 46.34 & 17.52 \\
    \midrule
    \textbf{AVG} & 95.03 & 96.33 & 88.75 & 66.07 & 32.90 \\
    \bottomrule
    \end{tabular}%
    }
  \label{tab:metrics1}%
\end{table}%

\begin{table}[b]
  \centering
    \renewcommand\arraystretch{0.6}
    \caption{Anomaly detection and classification results of \textbf{UniADC{\fontsize{7}{7}\selectfont \texttt{(CLIP)}}} on the MVTec-FS dataset under the zero-shot $(K_n=2)$ setting.}
    \resizebox{0.8\linewidth}{!}{
    \begin{tabular}{c|ccccc}
    \toprule
    \multicolumn{1}{c|}{Category} & \multicolumn{1}{c}{I-AUC} & \multicolumn{1}{c}{P-AUC} & \multicolumn{1}{c}{PRO} & \multicolumn{1}{c}{Acc} & \multicolumn{1}{c}{mIoU} \\
    \midrule
    Bottle & 98.06 & 96.54 & 84.09 & 76.47 & 39.18 \\
    Cable & 93.06 & 94.98 & 86.98 & 66.33 & 27.81 \\
    Capsule & 89.01 & 94.83 & 90.08 & 61.84 & 26.84 \\
    Carpet & 100.0   & 98.74 & 96.86 & 71.43 & 29.19 \\
    Grid  & 99.82 & 98.71 & 95.49 & 70.83 & 30.34 \\
    Hazelnut & 99.56 & 99.18 & 94.88 & 85.14 & 44.63 \\
    Leather & 100.0   & 99.87 & 97.17 & 75.00    & 35.64 \\
    Metal Nut & 99.70  & 97.55 & 84.20  & 83.58 & 41.15 \\
    Pill  & 96.83 & 98.25 & 97.20  & 65.12 & 26.20 \\
    Screw & 80.91 & 97.48 & 88.24 & 45.45 & 20.81 \\
    Tile  & 100.0   & 99.16 & 94.53 & 81.08 & 48.93 \\
    Toothbrush & 95.56 & 98.16 & 67.10  & 85.19 & 63.61 \\
    Transistor & 79.75 & 77.45 & 60.85 & 75.00    & 22.31 \\
    Wood  & 98.12 & 98.84 & 95.82 & 80.95 & 50.89 \\
    Zipper & 98.65 & 98.23 & 94.36 & 51.22 & 19.24 \\
    \midrule
    \textbf{AVG} & 95.27 & 96.53 & 88.52 & 71.64 & 35.12 \\
    \bottomrule
    \end{tabular}%
    }
  \label{tab:metrics2}%
\end{table}%

\begin{table}[t]
  \centering
  \renewcommand\arraystretch{0.6}
  \caption{Anomaly detection and classification results of \textbf{UniADC{\fontsize{7}{7}\selectfont \texttt{(CLIP)}}} on the MVTec-FS dataset under the zero-shot $(K_n=4)$ setting.}
     \resizebox{0.8\linewidth}{!}{
    \begin{tabular}{c|ccccc}
    \toprule
    \multicolumn{1}{c|}{Category} & \multicolumn{1}{c}{I-AUC} & \multicolumn{1}{c}{P-AUC} & \multicolumn{1}{c}{PRO} & \multicolumn{1}{c}{Acc} & \multicolumn{1}{c}{mIoU} \\
    \midrule
    Bottle & 99.20  & 94.69 & 87.26 & 74.51 & 40.20 \\
    Cable & 91.99 & 93.31 & 82.95 & 74.49 & 30.52 \\
    Capsule & 96.23 & 97.55 & 94.14 & 63.16 & 28.08 \\
    Carpet & 100.0  & 98.82 & 97.59 & 72.86 & 30.27 \\
    Grid  & 98.41 & 99.04 & 95.48 & 68.75 & 23.19 \\
    Hazelnut & 99.85 & 98.79 & 95.48 & 81.08 & 39.17 \\
    Leather & 99.15 & 99.65 & 99.05 & 76.32 & 39.82 \\
    Metal Nut & 97.79 & 96.22 & 90.50  & 82.09 & 40.20 \\
    Pill  & 98.64 & 98.93 & 95.62 & 63.95 & 26.92 \\
    Screw & 86.42 & 98.57 & 91.65 & 49.49 & 20.35 \\
    Tile  & 100.0   & 98.60  & 95.79 & 85.14 & 45.41 \\
    Toothbrush & 97.78 & 95.18 & 75.49 & 92.59 & 62.41 \\
    Transistor & 83.50  & 80.84 & 78.72 & 75.00   & 22.92 \\
    Wood  & 98.50  & 98.88 & 95.30  & 88.10  & 57.81 \\
    Zipper & 95.26 & 97.93 & 92.59 & 52.44 & 19.76 \\
    \midrule
    \textbf{AVG} & 96.18 & 96.47 & 91.17 & 73.33 & 35.14 \\
    \bottomrule
    \end{tabular}%
    }
  \label{tab:metrics3}%
\end{table}%

\begin{table}[b]
  \centering
  \renewcommand\arraystretch{0.6}
  \caption{Anomaly detection and classification results of \textbf{UniADC{\fontsize{7}{7}\selectfont \texttt{(CLIP)}}} on the MVTec-FS dataset under the few-shot $(K_n=1, K_a=1)$ setting.}
     \resizebox{0.8\linewidth}{!}{
    \begin{tabular}{c|ccccc}
    \toprule
    \multicolumn{1}{c|}{Category} & \multicolumn{1}{c}{I-AUC} & \multicolumn{1}{c}{P-AUC} & \multicolumn{1}{c}{PRO} & \multicolumn{1}{c}{Acc} & \multicolumn{1}{c}{mIoU} \\
    \midrule
    Bottle & 99.03 & 99.53 & 90.68 & 84.31 & 61.97 \\
    Cable & 98.89 & 96.78 & 90.21 & 93.88 & 52.38 \\
    Capsule & 97.70  & 98.24 & 94.09 & 71.05 & 32.34 \\
    Carpet & 100.0   & 99.66 & 91.38 & 90.00    & 48.08 \\
    Grid  & 99.82 & 99.46 & 97.57 & 85.42 & 36.86 \\
    Hazelnut & 100.0   & 98.50  & 88.62 & 100.0   & 55.76 \\
    Leather & 100.0   & 99.85 & 95.80  & 88.16 & 53.67 \\
    Metal Nut & 100.0   & 99.75 & 90.97 & 88.06 & 65.96 \\
    Pill  & 95.76 & 99.62 & 84.66 & 65.12 & 38.28 \\
    Screw & 83.77 & 97.79 & 87.61 & 46.46 & 23.47 \\
    Tile  & 100.0   & 99.42 & 93.79 & 100.0   & 77.38 \\
    Toothbrush & 95.00    & 97.10  & 71.59 & 92.59 & 59.24 \\
    Transistor & 96.50  & 93.75 & 74.14 & 87.50  & 24.65 \\
    Wood  & 99.62 & 98.99 & 96.54 & 92.86 & 51.61 \\
    Zipper & 99.30  & 99.54 & 98.14 & 85.37 & 47.66 \\
    \midrule
    \textbf{AVG} & 97.69 & 98.53 & 89.72 & 84.72 & 48.62 \\
    \bottomrule
    \end{tabular}%
    }
  \label{tab:metrics4}%
\end{table}%

\begin{table}[t]
  \centering
  \renewcommand\arraystretch{0.6}
  \caption{Anomaly detection and classification results of \textbf{UniADC{\fontsize{7}{7}\selectfont \texttt{(CLIP)}}} on the MVTec-FS dataset under the few-shot $(K_n=2, K_a=1)$ setting.}
     \resizebox{0.8\linewidth}{!}{
    \begin{tabular}{c|ccccc}
    \toprule
    \multicolumn{1}{c|}{Category} & \multicolumn{1}{c}{I-AUC} & \multicolumn{1}{c}{P-AUC} & \multicolumn{1}{c}{PRO} & \multicolumn{1}{c}{Acc} & \multicolumn{1}{c}{mIoU} \\
    \midrule
    Bottle & 99.35 & 99.51 & 84.85 & 84.31 & 63.35 \\
    Cable & 98.08 & 97.06 & 86.59 & 94.90  & 51.09 \\
    Capsule & 98.11 & 98.54 & 95.52 & 65.79 & 35.10 \\
    Carpet & 100.0  & 99.50  & 98.93 & 87.14 & 46.97 \\
    Grid  & 100.0  & 99.52 & 94.52 & 85.42 & 34.83 \\
    Hazelnut & 100.0   & 98.16 & 87.54 & 98.65 & 57.38 \\
    Leather & 100.0   & 99.85 & 97.10  & 88.16 & 49.87 \\
    Metal Nut & 100.0   & 99.67 & 87.34 & 89.55 & 63.53 \\
    Pill  & 97.45 & 99.46 & 91.53 & 62.79 & 38.66 \\
    Screw & 87.89 & 98.40  & 86.62 & 60.61 & 31.29 \\
    Tile  & 100.0   & 99.45 & 93.00    & 100.0   & 78.11 \\
    Toothbrush & 98.33 & 96.91 & 85.50 & 92.59 & 64.82 \\
    Transistor & 96.50  & 94.70  & 83.74 & 86.25 & 27.10 \\
    Wood  & 99.81 & 99.18 & 95.58 & 92.86 & 49.89 \\
    Zipper & 99.73 & 99.54 & 91.30  & 89.02 & 41.97 \\
    \midrule
    \textbf{AVG} & 98.35 & 98.63 & 90.64 & 85.20  & 48.93 \\
    \bottomrule
    \end{tabular}%
    }
  \label{tab:metrics5}%
\end{table}%

\begin{table}[b]
  \centering
  \renewcommand\arraystretch{0.6}
  \caption{Anomaly detection and classification results of \textbf{UniADC{\fontsize{7}{7}\selectfont \texttt{(CLIP)}}} on the MVTec-FS dataset under the few-shot $(K_n=2, K_a=2)$ setting.}
     \resizebox{0.8\linewidth}{!}{
    \begin{tabular}{c|ccccc}
    \toprule
    \multicolumn{1}{c|}{Category} & \multicolumn{1}{c}{I-AUC} & \multicolumn{1}{c}{P-AUC} & \multicolumn{1}{c}{PRO} & \multicolumn{1}{c}{Acc} & \multicolumn{1}{c}{mIoU} \\
    \midrule
    Bottle & 99.03 & 99.49 & 91.75 & 86.27 & 60.96 \\
    Cable & 98.39 & 96.76 & 93.63 & 93.88 & 51.90 \\
    Capsule & 97.29 & 98.83 & 95.44 & 78.95 & 36.03 \\
    Carpet & 100.0   & 99.50  & 95.08 & 92.86 & 49.76 \\
    Grid  & 99.82 & 99.39 & 97.45 & 87.50 & 38.92 \\
    Hazelnut & 100.0   & 99.71 & 95.57 & 100.0   & 75.10 \\
    Leather & 100.0   & 99.84 & 98.46 & 93.42 & 58.77 \\
    Metal Nut & 100.0   & 99.75 & 94.96 & 88.06 & 65.27 \\
    Pill  & 99.04 & 99.81 & 90.94 & 77.91 & 49.99 \\
    Screw & 95.04 & 99.13 & 95.06 & 72.73 & 36.34 \\
    Tile  & 100.0   & 99.41 & 94.54 & 100.0   & 79.37 \\
    Toothbrush & 96.11 & 97.39 & 78.95 & 92.59 & 65.9 \\
    Transistor & 97.50 & 95.51 & 85.46 & 92.50 & 36.52 \\
    Wood  & 99.62 & 99.29 & 97.20  & 97.62 & 63.92 \\
    Zipper & 99.35 & 99.42 & 95.10  & 93.90  & 55.24 \\
    \midrule
    \textbf{AVG} & 98.75 & 98.88 & 93.31 & 89.88 & 54.93 \\
    \bottomrule
    \end{tabular}%
    }
  \label{tab:metrics6}%
\end{table}%

\begin{table}[t]
  \centering
    \renewcommand\arraystretch{0.6}
  \caption{Anomaly detection and classification results of \textbf{UniADC{\fontsize{7}{7}\selectfont \texttt{(CLIP)}}} on the MVTec-FS dataset under the few-shot $(K_n=4, K_a=1)$ setting.}
     \resizebox{0.8\linewidth}{!}{
    \begin{tabular}{c|ccccc}
    \toprule
    \multicolumn{1}{c|}{Category} & \multicolumn{1}{c}{I-AUC} & \multicolumn{1}{c}{P-AUC} & \multicolumn{1}{c}{PRO} & \multicolumn{1}{c}{Acc} & \multicolumn{1}{c}{mIoU} \\
    \midrule
    Bottle & 100.0   & 99.55 & 93.76 & 82.35 & 65.97 \\
    Cable & 99.04 & 97.00    & 89.19 & 94.90  & 53.99 \\
    Capsule & 98.85 & 98.23 & 93.57 & 68.42 & 33.75 \\
    Carpet & 100.0   & 99.67 & 98.75 & 88.57 & 46.52 \\
    Grid  & 99.29 & 99.49 & 97.22 & 87.50  & 40.92 \\
    Hazelnut & 100.0   & 98.27 & 95.95 & 97.30  & 55.50 \\
    Leather & 100.0   & 99.85 & 95.94 & 86.84 & 54.37 \\
    Metal Nut & 100.0   & 99.72 & 90.39 & 88.06 & 64.02 \\
    Pill  & 98.08 & 99.62 & 89.39 & 70.93 & 41.00 \\
    Screw & 92.60  & 98.89 & 82.67 & 55.56 & 26.84 \\
    Tile  & 100.0   & 99.50  & 92.37 & 100.0   & 76.21 \\
    Toothbrush & 96.67 & 97.76 & 78.30  & 92.59 & 65.16 \\
    Transistor & 94.50  & 92.34 & 82.60  & 87.50  & 28.71 \\
    Wood  & 100.0   & 98.97 & 96.22 & 95.24 & 53.34 \\
    Zipper & 99.52 & 99.46 & 95.03 & 87.80  & 43.81 \\
    \midrule
    \textbf{AVG} & 98.57 & 98.55 & 91.42 & 85.57 & 50.01 \\
    \bottomrule
    \end{tabular}%
    }
  \label{tab:metrics7}%
\end{table}%

\begin{table}[b]
  \centering
  \renewcommand\arraystretch{0.6}
  \caption{Anomaly detection and classification results of \textbf{UniADC{\fontsize{7}{7}\selectfont \texttt{(CLIP)}}} on the MVTec-FS dataset with full-shot normal samples and $K_a=1$.}
     \resizebox{0.8\linewidth}{!}{
    \begin{tabular}{c|ccccc}
    \toprule
    \multicolumn{1}{c|}{Category} & \multicolumn{1}{c}{I-AUC} & \multicolumn{1}{c}{P-AUC} & \multicolumn{1}{c}{PRO} & \multicolumn{1}{c}{Acc} & \multicolumn{1}{c}{mIoU} \\
    \midrule
    Bottle & 100.0   & 99.68 & 92.40  & 82.35 & 65.17 \\
    Cable & 99.35 & 96.77 & 94.87 & 94.90  & 55.04 \\
    Capsule & 99.51 & 99.00    & 97.08 & 71.05 & 34.34 \\
    Carpet & 100.0   & 99.34 & 95.58 & 88.57 & 47.77 \\
    Grid  & 99.65 & 99.39 & 96.92 & 89.58 & 42.56 \\
    Hazelnut & 100.0   & 98.02 & 92.78 & 98.65 & 57.54 \\
    Leather & 100.0   & 99.83 & 97.24 & 88.16 & 52.82 \\
    Metal Nut & 100.0   & 99.74 & 94.11 & 94.03 & 66.48 \\
    Pill  & 99.04 & 99.68 & 92.54 & 72.09 & 45.28 \\
    Screw & 91.34 & 98.92 & 90.67 & 62.63 & 30.26 \\
    Tile  & 100.0   & 99.44 & 96.61 & 100.0   & 76.79 \\
    Toothbrush & 96.11 & 98.81 & 83.80  & 92.59 & 66.97 \\
    Transistor & 98.25 & 94.12 & 87.52 & 90.00    & 28.23 \\
    Wood  & 100.0   & 98.65 & 96.34 & 95.24 & 58.06 \\
    Zipper & 99.46 & 99.54 & 98.00    & 85.37 & 46.49 \\
    \midrule
    \textbf{AVG} & 98.85 & 98.73 & 93.76 & 87.01 & 51.59 \\
    \bottomrule
    \end{tabular}%
    }
  \label{tab:metrics8}%
\end{table}%

\begin{table}[t]
  \centering
  \renewcommand\arraystretch{0.6}
  \caption{Anomaly detection and classification results of \textbf{UniADC{\fontsize{7}{7}\selectfont \texttt{(DINO)}}} on the MVTec-FS dataset under the zero-shot $(K_n=1)$ setting.}
     \resizebox{0.8\linewidth}{!}{
    \begin{tabular}{c|ccccc}
    \toprule
    \multicolumn{1}{c|}{Category} & \multicolumn{1}{c}{I-AUC} & \multicolumn{1}{c}{P-AUC} & \multicolumn{1}{c}{PRO} & \multicolumn{1}{c}{Acc} & \multicolumn{1}{c}{mIoU} \\
    \midrule
    Bottle & 99.03 & 93.73 & 80.66 & 80.39 & 47.82 \\
    Cable & 94.29 & 94.46 & 88.32 & 55.10  & 22.53 \\
    Capsule & 94.26 & 98.02 & 93.16 & 59.21 & 26.57 \\
    Carpet & 100.0   & 99.29 & 98.43 & 85.71 & 39.34 \\
    Grid  & 100.0   & 99.61 & 98.45 & 62.50  & 25.09 \\
    Hazelnut & 99.19 & 98.24 & 80.86 & 75.68 & 38.12 \\
    Leather & 100.0   & 99.80  & 96.53 & 80.26 & 39.12 \\
    Metal Nut & 94.65 & 97.16 & 96.34 & 55.22 & 37.30 \\
    Pill  & 96.83 & 92.32 & 95.57 & 50.00    & 20.50 \\
    Screw & 81.96 & 97.51 & 87.82 & 31.31 & 20.34 \\
    Tile  & 99.63 & 97.83 & 90.58 & 78.38 & 35.43 \\
    Toothbrush & 95.00    & 98.87 & 64.09 & 92.59 & 70.73 \\
    Transistor & 92.75 & 77.29 & 72.72 & 81.25 & 24.50 \\
    Wood  & 98.31 & 98.38 & 97.27 & 85.71 & 60.13 \\
    Zipper & 99.68 & 99.09 & 96.53 & 51.22 & 18.35 \\
    \midrule
    \textbf{AVG} & 96.37 & 96.11 & 89.16 & 68.30  & 35.06 \\
    \bottomrule
    \end{tabular}%
    }
  \label{tab:metrics9}%
\end{table}%

\begin{table}[b]
  \centering
  \renewcommand\arraystretch{0.6}
  \caption{Anomaly detection and classification results of \textbf{UniADC{\fontsize{7}{7}\selectfont \texttt{(DINO)}}} on the MVTec-FS dataset under the zero-shot $(K_n=2)$ setting.}
     \resizebox{0.8\linewidth}{!}{
    \begin{tabular}{c|ccccc}
    \toprule
    \multicolumn{1}{c|}{Category} & \multicolumn{1}{c}{I-AUC} & \multicolumn{1}{c}{P-AUC} & \multicolumn{1}{c}{PRO} & \multicolumn{1}{c}{Acc} & \multicolumn{1}{c}{mIoU} \\
    \midrule
    Bottle & 98.39 & 97.81 & 93.78 & 82.35 & 54.97 \\
    Cable & 92.53 & 96.26 & 87.31 & 65.31 & 23.65 \\
    Capsule & 98.52 & 98.78 & 97.59 & 61.84 & 32.78 \\
    Carpet & 99.66 & 98.71 & 97.98 & 81.43 & 34.54 \\
    Grid  & 100.0   & 99.61 & 96.57 & 72.92 & 31.05 \\
    Hazelnut & 99.63 & 97.53 & 91.29 & 85.14 & 38.84 \\
    Leather & 99.93 & 99.52 & 98.39 & 82.89 & 41.56 \\
    Metal Nut & 99.39 & 98.05 & 97.08 & 79.10  & 42.24 \\
    Pill  & 95.98 & 88.81 & 96.12 & 58.14 & 23.38 \\
    Screw & 88.98 & 98.25 & 91.98 & 55.56 & 26.93 \\
    Tile  & 99.48 & 98.66 & 94.62 & 78.38 & 34.05 \\
    Toothbrush & 99.44 & 97.19 & 79.21 & 96.30  & 68.10 \\
    Transistor & 92.92 & 89.11 & 69.53 & 77.50  & 21.62 \\
    Wood  & 93.61 & 98.95 & 96.26 & 88.10  & 56.52 \\
    Zipper & 97.95 & 98.36 & 94.50  & 56.10  & 19.73 \\
    \midrule
    \textbf{AVG} & 97.09 & 97.04 & 92.15 & 74.74 & 36.66 \\
    \bottomrule
    \end{tabular}%
    }
  \label{tab:metrics10}%
\end{table}%

\begin{table}[t]
  \centering
  \renewcommand\arraystretch{0.6}
  \caption{Anomaly detection and classification results of \textbf{UniADC{\fontsize{7}{7}\selectfont \texttt{(DINO)}}} on the MVTec-FS dataset under the zero-shot $(K_n=4)$ setting.}
     \resizebox{0.8\linewidth}{!}{
    \begin{tabular}{c|ccccc}
    \toprule
    \multicolumn{1}{c|}{Category} & \multicolumn{1}{c}{I-AUC} & \multicolumn{1}{c}{P-AUC} & \multicolumn{1}{c}{PRO} & \multicolumn{1}{c}{Acc} & \multicolumn{1}{c}{mIoU} \\
    \midrule
    Bottle & 99.35 & 96.82 & 94.11 & 82.35 & 56.14 \\
    Cable & 92.07 & 95.40  & 90.47 & 71.43 & 29.28 \\
    Capsule & 97.95 & 97.00    & 93.41 & 63.16 & 31.50 \\
    Carpet & 100.0   & 99.63 & 95.21 & 72.86 & 30.93 \\
    Grid  & 100.0   & 99.60  & 97.57 & 66.67 & 28.54 \\
    Hazelnut & 99.56 & 98.62 & 95.24 & 85.14 & 36.31 \\
    Leather & 100.0   & 99.78 & 99.63 & 90.78 & 49.25 \\
    Metal Nut & 99.90 & 98.36 & 94.95 & 82.09 & 37.84 \\
    Pill  & 97.79 & 96.11 & 97.90  & 56.98 & 21.66 \\
    Screw & 88.44 & 97.86 & 91.35 & 55.56 & 20.66 \\
    Tile  & 98.45 & 98.07 & 95.64 & 86.49 & 43.62 \\
    Toothbrush & 100.0   & 96.90  & 73.31 & 100.0   & 63.42 \\
    Transistor & 95.00    & 88.84 & 79.31 & 86.25 & 26.01 \\
    Wood  & 98.12 & 99.16 & 97.04 & 88.10  & 59.55 \\
    Zipper & 98.06 & 98.27 & 95.05 & 57.32 & 23.80 \\
    \midrule
    \textbf{AVG} & 97.65 & 97.36 & 92.68 & 76.35 & 37.23 \\
    \bottomrule
    \end{tabular}%
    }
  \label{tab:metrics11}%
\end{table}%

\begin{table}[b]
  \centering
  \renewcommand\arraystretch{0.6}
  \caption{Anomaly detection and classification results of \textbf{UniADC{\fontsize{7}{7}\selectfont \texttt{(DINO)}}} on the MVTec-FS dataset under the few-shot $(K_n=1, K_a=1)$ setting.}
     \resizebox{0.8\linewidth}{!}{
    \begin{tabular}{c|ccccc}
    \toprule
    \multicolumn{1}{c|}{Category} & \multicolumn{1}{c}{I-AUC} & \multicolumn{1}{c}{P-AUC} & \multicolumn{1}{c}{PRO} & \multicolumn{1}{c}{Acc} & \multicolumn{1}{c}{mIoU} \\
    \midrule
    Bottle & 99.52 & 99.71 & 88.05 & 82.35 & 61.95 \\
    Cable & 96.74 & 96.77 & 90.09 & 92.86 & 54.27 \\
    Capsule & 99.51 & 99.26 & 96.03 & 76.32 & 36.23 \\
    Carpet & 100.0   & 99.81 & 98.04 & 81.43 & 45.19 \\
    Grid  & 100.0   & 99.51 & 96.77 & 91.67 & 41.38 \\
    Hazelnut & 100.0   & 98.36 & 93.94 & 100.0   & 54.67 \\
    Leather & 100.0   & 99.84 & 96.65 & 93.42 & 50.08 \\
    Metal Nut & 99.90  & 99.74 & 94.22 & 88.06 & 65.78 \\
    Pill  & 95.48 & 99.67 & 79.47 & 66.28 & 40.39 \\
    Screw & 85.58 & 99.17 & 88.31 & 54.55 & 30.12 \\
    Tile  & 100.0   & 99.44 & 96.31 & 98.65 & 74.52 \\
    Toothbrush & 100.0   & 97.73 & 81.59 & 100.0   & 67.76 \\
    Transistor & 100.0   & 96.68 & 92.18 & 91.25 & 30.78 \\
    Wood  & 99.62 & 99.03 & 94.94 & 92.86 & 57.39 \\
    Zipper & 99.95 & 99.72 & 97.24 & 91.46 & 58.72 \\
    \midrule
    \textbf{AVG} & 98.42 & 98.96 & 92.26 & 86.74 & 51.28 \\
    \bottomrule
    \end{tabular}%
    }
  \label{tab:metrics12}%
\end{table}%

\begin{table}[t]
  \centering
  \renewcommand\arraystretch{0.6}
  \caption{Anomaly detection and classification results of \textbf{UniADC{\fontsize{7}{7}\selectfont \texttt{(DINO)}}} on the MVTec-FS dataset under the few-shot $(K_n=2, K_a=1)$ setting.}
     \resizebox{0.8\linewidth}{!}{
    \begin{tabular}{c|ccccc}
    \toprule
    \multicolumn{1}{c|}{Category} & \multicolumn{1}{c}{I-AUC} & \multicolumn{1}{c}{P-AUC} & \multicolumn{1}{c}{PRO} & \multicolumn{1}{c}{Acc} & \multicolumn{1}{c}{mIoU} \\
    \midrule
    Bottle & 100.0   & 99.69 & 90.82 & 82.35 & 61.38 \\
    Cable & 97.28 & 97.22 & 85.47 & 93.88 & 56.36 \\
    Capsule & 98.28 & 99.18 & 95.61 & 75.00    & 38.77 \\
    Carpet & 100.0   & 99.41 & 98.48 & 81.43 & 43.57 \\
    Grid  & 100.0   & 99.58 & 96.91 & 91.67 & 37.38 \\
    Hazelnut & 100.0   & 99.12 & 95.51 & 98.65 & 54.04 \\
    Leather & 100.0   & 99.83 & 96.21 & 93.42 & 50.59 \\
    Metal Nut & 100.0   & 99.77 & 93.91 & 89.55 & 65.82 \\
    Pill  & 97.17 & 99.60  & 87.16 & 68.60  & 40.80 \\
    Screw & 88.10  & 98.98 & 86.88 & 62.63 & 31.49 \\
    Tile  & 100.0   & 99.31 & 96.79 & 98.65 & 75.22 \\
    Toothbrush & 98.33 & 97.13 & 87.94 & 92.59 & 68.74 \\
    Transistor & 99.92 & 96.25 & 86.5  & 91.25 & 29.50 \\
    Wood  & 99.25 & 98.73 & 94.72 & 92.86 & 57.74 \\
    Zipper & 100.0   & 99.72 & 94.34 & 90.24 & 60.95 \\
    \midrule
    \textbf{AVG} & 98.56 & 98.90  & 92.48 & 86.85 & 51.49 \\
    \bottomrule
    \end{tabular}%
    }
  \label{tab:metrics13}%
\end{table}%

\begin{table}[b]
  \centering
  \renewcommand\arraystretch{0.6}
  \caption{Anomaly detection and classification results of \textbf{UniADC{\fontsize{7}{7}\selectfont \texttt{(DINO)}}} on the MVTec-FS dataset under the few-shot $(K_n=2, K_a=2)$ setting.}
     \resizebox{0.8\linewidth}{!}{
    \begin{tabular}{c|ccccc}
    \toprule
    \multicolumn{1}{c|}{Category} & \multicolumn{1}{c}{I-AUC} & \multicolumn{1}{c}{P-AUC} & \multicolumn{1}{c}{PRO} & \multicolumn{1}{c}{Acc} & \multicolumn{1}{c}{mIoU} \\
    \midrule
    Bottle & 100.0   & 99.67 & 94.18 & 84.31 & 55.23 \\
    Cable & 98.05 & 97.93 & 89.26 & 87.76 & 45.81 \\
    Capsule & 98.93 & 99.10  & 92.47 & 77.63 & 40.49 \\
    Carpet & 99.66 & 99.70  & 98.77 & 88.57 & 50.76 \\
    Grid  & 100.0   & 99.63 & 96.01 & 87.50  & 41.06 \\
    Hazelnut & 100.0   & 99.78 & 95.58 & 97.30  & 70.42 \\
    Leather & 100.0   & 99.82 & 96.77 & 93.42 & 52.75 \\
    Metal Nut & 100.0   & 99.76 & 95.22 & 88.06 & 66.13 \\
    Pill  & 99.21 & 99.81 & 93.64 & 73.26 & 51.18 \\
    Screw & 89.96 & 99.13 & 95.36 & 72.73 & 36.34 \\
    Tile  & 100.0   & 99.30  & 98.23 & 100.0   & 77.04 \\
    Toothbrush & 100.0   & 97.30  & 83.60  & 96.30  & 66.82 \\
    Transistor & 99.92 & 96.76 & 85.19 & 91.25 & 38.44 \\
    Wood  & 100.0   & 99.12 & 94.82 & 97.62 & 60.69 \\
    Zipper & 100.0   & 99.67 & 94.12 & 95.12 & 59.93 \\
    \midrule
    \textbf{AVG} & 99.05 & 99.10  & 93.55 & 88.72 & 54.21 \\
    \bottomrule
    \end{tabular}%
    }
  \label{tab:metrics14}%
\end{table}%

\begin{table}[t]
  \centering
  \renewcommand\arraystretch{0.6}
  \caption{Anomaly detection and classification results of \textbf{UniADC{\fontsize{7}{7}\selectfont \texttt{(DINO)}}} on the MVTec-FS dataset under the few-shot $(K_n=4, K_a=1)$ setting.}
     \resizebox{0.8\linewidth}{!}{
    \begin{tabular}{c|ccccc}
    \toprule
    \multicolumn{1}{c|}{Category} & \multicolumn{1}{c}{I-AUC} & \multicolumn{1}{c}{P-AUC} & \multicolumn{1}{c}{PRO} & \multicolumn{1}{c}{Acc} & \multicolumn{1}{c}{mIoU} \\
    \midrule
    Bottle & 99.68 & 99.65 & 92.64 & 82.35 & 64.09 \\
    Cable & 97.82 & 96.17 & 86.18 & 92.86 & 56.49 \\
    Capsule & 99.10  & 99.24 & 92.04 & 76.32 & 39.26 \\
    Carpet & 100.0   & 99.23 & 98.07 & 84.29 & 45.52 \\
    Grid  & 100.0   & 99.59 & 96.08 & 91.67 & 38.76 \\
    Hazelnut & 100.0   & 99.03 & 98.33 & 95.95 & 55.39 \\
    Leather & 100.0   & 99.83 & 96.81 & 89.47 & 54.12 \\
    Metal Nut & 100.0   & 99.79 & 92.50  & 86.57 & 65.63 \\
    Pill  & 97.51 & 99.66 & 92.65 & 69.77 & 43.59 \\
    Screw & 87.55 & 98.97 & 89.40  & 57.58 & 27.68 \\
    Tile  & 100.0   & 98.55 & 97.52 & 100.0   & 74.99 \\
    Toothbrush & 100.0   & 98.26 & 86.35 & 100.0   & 69.02 \\
    Transistor & 99.83 & 97.00    & 89.72 & 91.25 & 33.87 \\
    Wood  & 99.06 & 98.97 & 93.59 & 92.86 & 58.14 \\
    Zipper & 100.0   & 99.74 & 98.44 & 91.46 & 59.02 \\
    \midrule
    \textbf{AVG} & 98.70  & 98.91 & 93.35 & 86.83 & 52.37 \\
    \bottomrule
    \end{tabular}%
    }
  \label{tab:metrics15}%
\end{table}%

\begin{table}[htbp]
  \centering
  \renewcommand\arraystretch{0.6}
  \caption{Anomaly detection and classification results of \textbf{UniADC{\fontsize{7}{7}\selectfont \texttt{(DINO)}}} on the MVTec-FS dataset with full-shot normal samples and $K_a=1$.}
     \resizebox{0.8\linewidth}{!}{
    \begin{tabular}{c|ccccc}
    \toprule
    \multicolumn{1}{c|}{Category} & \multicolumn{1}{c}{I-AUC} & \multicolumn{1}{c}{P-AUC} & \multicolumn{1}{c}{PRO} & \multicolumn{1}{c}{Acc} & \multicolumn{1}{c}{mIoU} \\
    \midrule
    Bottle & 100.0   & 99.71 & 94.61 & 78.43 & 58.28 \\
    Cable & 97.66 & 96.48 & 90.91 & 94.90  & 56.25 \\
    Capsule & 99.26 & 99.33 & 95.31 & 75.00    & 42.60 \\
    Carpet & 99.92 & 99.23 & 96.55 & 82.86 & 42.77 \\
    Grid  & 100.0   & 99.59 & 96.71 & 85.42 & 42.41 \\
    Hazelnut & 100.0   & 98.65 & 92.32 & 97.30  & 55.91 \\
    Leather & 100.0   & 99.85 & 96.04 & 92.11 & 52.04 \\
    Metal Nut & 100.0   & 99.78 & 94.00    & 89.55 & 66.52 \\
    Pill  & 99.32 & 99.62 & 89.46 & 72.09 & 44.18 \\
    Screw & 95.21 & 99.00   & 90.42 & 74.75 & 31.86 \\
    Tile  & 100.0   & 99.34 & 97.99 & 100.0   & 75.35 \\
    Toothbrush & 100.0   & 97.10  & 94.60  & 100.0   & 67.21 \\
    Transistor & 100.0   & 97.87 & 89.15 & 90.00    & 34.11 \\
    Wood  & 99.81 & 98.99 & 97.04 & 95.24 & 53.83 \\
    Zipper & 100.0   & 99.68 & 95.79 & 90.24 & 55.70\\
    \midrule
    \textbf{AVG} & 99.41 & 98.95 & 94.06 & 87.86 & 51.93 \\
    \bottomrule
    \end{tabular}%
    }
  \label{tab:metrics16}%
\end{table}%

\begin{table}[htbp]
  \centering
  \renewcommand\arraystretch{0.6}
  \caption{Anomaly detection and classification results of \textbf{UniADC{\fontsize{7}{7}\selectfont \texttt{(CLIP)}}} on the WFDD dataset under the zero-shot $(K_n=1)$ setting.}
     \resizebox{0.8\linewidth}{!}{
    \begin{tabular}{c|ccccc}
    \toprule
    \multicolumn{1}{c|}{Category} & \multicolumn{1}{c}{I-AUC} & \multicolumn{1}{c}{P-AUC} & \multicolumn{1}{c}{PRO} & \multicolumn{1}{c}{Acc} & \multicolumn{1}{c}{mIoU} \\
    \midrule
    Grey Cloth & 99.89 & 99.08 & 85.75 & 81.25 & 38.05 \\
    Grid Cloth & 94.51 & 98.10  & 78.98 & 90.00    & 44.08 \\
    Pink Flower & 95.18 & 99.90  & 90.85 & 85.71 & 55.14 \\
    Yellow Cloth & 98.46 & 98.92 & 92.54 & 89.82 & 44.93 \\
    \midrule
    \textbf{AVG} & 97.01 & 99.00    & 87.03 & 86.70  & 45.55 \\
    \bottomrule
    \end{tabular}%
    }
  \label{tab:metrics17}%
\end{table}%

\begin{table}[htbp]
  \centering
  \renewcommand\arraystretch{0.6}
  \caption{Anomaly detection and classification results of \textbf{UniADC{\fontsize{7}{7}\selectfont \texttt{(CLIP)}}} on the WFDD dataset under the zero-shot $(K_n=2)$ setting.}
     \resizebox{0.8\linewidth}{!}{
    \begin{tabular}{c|ccccc}
    \toprule
    \multicolumn{1}{c|}{Category} & \multicolumn{1}{c}{I-AUC} & \multicolumn{1}{c}{P-AUC} & \multicolumn{1}{c}{PRO} & \multicolumn{1}{c}{Acc} & \multicolumn{1}{c}{mIoU} \\
    \midrule
    Grey Cloth & 99.11 & 98.67 & 84.78 & 92.19 & 42.70 \\
    Grid Cloth & 95.43 & 98.56 & 79.80  & 86.00    & 41.23 \\
    Pink Flower & 97.27 & 99.90  & 91.49 & 89.29 & 62.02 \\
    Yellow Cloth & 98.39 & 98.79 & 92.47 & 90.42 & 46.64 \\
    \midrule
    \textbf{AVG} & 97.55 & 98.98 & 87.14 & 89.48 & 48.15 \\
    \bottomrule
    \end{tabular}%
    }
  \label{tab:metrics18}%
\end{table}%

\begin{table}[htbp]
  \centering
  \renewcommand\arraystretch{0.6}
  \caption{Anomaly detection and classification results of \textbf{UniADC{\fontsize{7}{7}\selectfont \texttt{(CLIP)}}} on the WFDD dataset under the zero-shot $(K_n=4)$ setting.}
     \resizebox{0.8\linewidth}{!}{
    \begin{tabular}{c|ccccc}
    \toprule
    \multicolumn{1}{c|}{Category} & \multicolumn{1}{c}{I-AUC} & \multicolumn{1}{c}{P-AUC} & \multicolumn{1}{c}{PRO} & \multicolumn{1}{c}{Acc} & \multicolumn{1}{c}{mIoU} \\
    \midrule
    Grey Cloth & 99.89 & 98.95 & 90.16 & 95.32 & 42.74 \\
    Grid Cloth & 94.42 & 98.51 & 75.61 & 83.00    & 41.52 \\
    Pink Flower & 98.96 & 99.67 & 96.86 & 92.86 & 62.84 \\
    Yellow Cloth & 99.34 & 99.22 & 90.51 & 92.22 & 45.85 \\
    \midrule
    \textbf{AVG} & 98.15 & 99.09 & 88.29 & 90.85 & 48.24 \\
    \bottomrule
    \end{tabular}%
    }
  \label{tab:metrics19}%
\end{table}%

\begin{table}[htbp]
  \centering
  \renewcommand\arraystretch{0.6}
  \caption{Anomaly detection and classification results of \textbf{UniADC{\fontsize{7}{7}\selectfont \texttt{(CLIP)}}} on the WFDD dataset under the few-shot $(K_n=1, K_a=1)$ setting.}
     \resizebox{0.8\linewidth}{!}{
    \begin{tabular}{c|ccccc}
    \toprule
    \multicolumn{1}{c|}{Category} & \multicolumn{1}{c}{I-AUC} & \multicolumn{1}{c}{P-AUC} & \multicolumn{1}{c}{PRO} & \multicolumn{1}{c}{Acc} & \multicolumn{1}{c}{mIoU} \\
    \midrule
    Grey Cloth & 99.78 & 98.08 & 88.76 & 92.19 & 39.84 \\
    Grid Cloth & 95.76 & 98.68 & 92.87 & 89.00    & 40.55 \\
    Pink Flower & 98.70  & 99.97 & 95.56 & 96.43 & 68.70 \\
    Yellow Cloth & 99.77 & 98.60  & 94.63 & 96.41 & 45.93 \\
    \midrule
    \textbf{AVG} & 98.50  & 98.83 & 92.96 & 93.51 & 48.76 \\
    \bottomrule
    \end{tabular}%
    }
  \label{tab:metrics20}%
\end{table}%

\begin{table}[htbp]
  \centering
  \renewcommand\arraystretch{0.6}
  \caption{Anomaly detection and classification results of \textbf{UniADC{\fontsize{7}{7}\selectfont \texttt{(CLIP)}}} on the WFDD dataset under the few-shot $(K_n=2, K_a=1)$ setting.}
     \resizebox{0.8\linewidth}{!}{
    \begin{tabular}{c|ccccc}
    \toprule
    \multicolumn{1}{c|}{Category} & \multicolumn{1}{c}{I-AUC} & \multicolumn{1}{c}{P-AUC} & \multicolumn{1}{c}{PRO} & \multicolumn{1}{c}{Acc} & \multicolumn{1}{c}{mIoU} \\
    \midrule
    Grey Cloth & 99.67 & 98.80  & 95.53 & 92.19 & 42.92 \\
    Grid Cloth & 97.41 & 98.62 & 94.95 & 89.00   & 41.42 \\
    Pink Flower & 98.96 & 99.98 & 93.74 & 96.43 & 68.46 \\
    Yellow Cloth & 99.85 & 99.23 & 94.35 & 97.01 & 46.04 \\
    \midrule
    \textbf{AVG} & 98.97 & 99.16 & 94.64 & 93.66 & 49.71 \\
    \bottomrule
    \end{tabular}%
    }
  \label{tab:metrics21}%
\end{table}%

\begin{table}[htbp]
  \centering
  \renewcommand\arraystretch{0.6}
  \caption{Anomaly detection and classification results of \textbf{UniADC{\fontsize{7}{7}\selectfont \texttt{(CLIP)}}} on the WFDD dataset under the few-shot $(K_n=2, K_a=2)$ setting.}
     \resizebox{0.8\linewidth}{!}{
    \begin{tabular}{c|ccccc}
    \toprule
    \multicolumn{1}{c|}{Category} & \multicolumn{1}{c}{I-AUC} & \multicolumn{1}{c}{P-AUC} & \multicolumn{1}{c}{PRO} & \multicolumn{1}{c}{Acc} & \multicolumn{1}{c}{mIoU} \\
    \midrule
    Grey Cloth & 99.11 & 98.89 & 95.70  & 93.75 & 42.85 \\
    Grid Cloth & 98.75 & 99.11 & 93.07 & 90.00    & 43.64 \\
    Pink Flower & 98.70  & 99.97 & 95.06 & 96.43 & 67.73 \\
    Yellow Cloth & 100.0   & 99.40  & 95.55 & 99.40  & 50.02 \\
    \midrule
    \textbf{AVG} & 99.14 & 99.34 & 94.85 & 94.90  & 51.06 \\
    \bottomrule
    \end{tabular}%
    }
  \label{tab:metrics22}%
\end{table}%

\begin{table}[htbp]
  \centering
  \renewcommand\arraystretch{0.6}
  \caption{Anomaly detection and classification results of \textbf{UniADC{\fontsize{7}{7}\selectfont \texttt{(CLIP)}}} on the WFDD dataset under the few-shot $(K_n=4, K_a=1)$ setting.}
     \resizebox{0.8\linewidth}{!}{
    \begin{tabular}{c|ccccc}
    \toprule
    \multicolumn{1}{c|}{Category} & \multicolumn{1}{c}{I-AUC} & \multicolumn{1}{c}{P-AUC} & \multicolumn{1}{c}{PRO} & \multicolumn{1}{c}{Acc} & \multicolumn{1}{c}{mIoU} \\
    \midrule
    Grey Cloth & 99.78 & 98.67 & 92.77 & 92.19 & 42.24 \\
    Grid Cloth & 98.06 & 98.89 & 94.81 & 90.00    & 42.79 \\
    Pink Flower & 99.61 & 99.97 & 95.55 & 96.43 & 67.80 \\
    Yellow Cloth & 99.94 & 99.24 & 94.80  & 97.01 & 46.92 \\
    \midrule
    \textbf{AVG} & 99.35 & 99.19 & 94.48 & 93.91 & 49.94 \\
    \bottomrule
    \end{tabular}%
    }
  \label{tab:metrics23}%
\end{table}%

\begin{table}[htbp]
  \centering
  \renewcommand\arraystretch{0.6}
  \caption{Anomaly detection and classification results of \textbf{UniADC{\fontsize{7}{7}\selectfont \texttt{(CLIP)}}} on the WFDD dataset with full-shot normal samples and $(K_a=1)$.}
     \resizebox{0.8\linewidth}{!}{
    \begin{tabular}{c|ccccc}
    \toprule
    \multicolumn{1}{c|}{Category} & \multicolumn{1}{c}{I-AUC} & \multicolumn{1}{c}{P-AUC} & \multicolumn{1}{c}{PRO} & \multicolumn{1}{c}{Acc} & \multicolumn{1}{c}{mIoU} \\
    \midrule
    Grey Cloth & 99.67 & 98.91 & 94.10 & 92.19 & 42.21 \\
    Grid Cloth & 98.30 & 98.55 & 95.79 & 90.00 &  44.99\\
    Pink Flower & 99.87 & 99.98 & 95.34 & 98.21 & 67.60 \\
    Yellow Cloth & 99.94 & 99.22 &  96.88 & 98.20 & 47.46 \\
    \midrule
    \textbf{AVG} & 99.45 & 99.17 & 95.53 & 94.65 & 50.57 \\
    \bottomrule
    \end{tabular}%
    }
  \label{tab:metrics24}%
\end{table}%

\begin{table}[htbp]
  \centering
  \renewcommand\arraystretch{0.6}
  \caption{Anomaly detection and classification results of \textbf{UniADC{\fontsize{7}{7}\selectfont \texttt{(DINO)}}} on the WFDD dataset under the zero-shot $(K_n=1)$ setting.}
     \resizebox{0.8\linewidth}{!}{
    \begin{tabular}{c|ccccc}
    \toprule
    \multicolumn{1}{c|}{Category} & \multicolumn{1}{c}{I-AUC} & \multicolumn{1}{c}{P-AUC} & \multicolumn{1}{c}{PRO} & \multicolumn{1}{c}{Acc} & \multicolumn{1}{c}{mIoU} \\
    \midrule
    Grey Cloth & 99.11 & 99.11 & 85.80  & 71.88 & 36.15 \\
    Grid Cloth & 95.76 & 99.29 & 92.48 & 92.00    & 52.53 \\
    Pink Flower & 97.66 & 99.94 & 93.57 & 94.64 & 66.77 \\
    Yellow Cloth & 99.77 & 99.49 & 96.65 & 97.01 & 53.24 \\
    \midrule
    \textbf{AVG} & 98.08 & 99.46 & 92.13 & 88.88 & 52.17 \\
    \bottomrule
    \end{tabular}%
    }
  \label{tab:metrics25}%
\end{table}%

\begin{table}[htbp]
  \centering
  \renewcommand\arraystretch{0.6}
  \caption{Anomaly detection and classification results of \textbf{UniADC{\fontsize{7}{7}\selectfont \texttt{(DINO)}}} on the WFDD dataset under the zero-shot $(K_n=2)$ setting.}
     \resizebox{0.8\linewidth}{!}{
    \begin{tabular}{c|ccccc}
    \toprule
    \multicolumn{1}{c|}{Category} & \multicolumn{1}{c}{I-AUC} & \multicolumn{1}{c}{P-AUC} & \multicolumn{1}{c}{PRO} & \multicolumn{1}{c}{Acc} & \multicolumn{1}{c}{mIoU} \\
    \midrule
    Grey Cloth & 99.89 & 99.22 & 84.57 & 73.44 & 38.96 \\
    Grid Cloth & 94.91 & 99.15 & 93.18 & 90.00    & 51.90 \\
    Pink Flower & 98.18 & 99.91 & 94.48 & 96.43 & 69.34 \\
    Yellow Cloth & 99.84 & 99.57 & 97.15 & 97.01 & 53.33 \\
    \midrule
    \textbf{AVG} & 98.21 & 99.46 & 92.35 & 89.22 & 53.38 \\
    \bottomrule
    \end{tabular}%
    }
  \label{tab:metrics26}%
\end{table}%

\begin{table}[htbp]
  \centering
  \renewcommand\arraystretch{0.6}
  \caption{Anomaly detection and classification results of \textbf{UniADC{\fontsize{7}{7}\selectfont \texttt{(DINO)}}} on the WFDD dataset under the zero-shot $(K_n=4)$ setting.}
     \resizebox{0.8\linewidth}{!}{
    \begin{tabular}{c|ccccc}
    \toprule
    \multicolumn{1}{c|}{Category} & \multicolumn{1}{c}{I-AUC} & \multicolumn{1}{c}{P-AUC} & \multicolumn{1}{c}{PRO} & \multicolumn{1}{c}{Acc} & \multicolumn{1}{c}{mIoU} \\
    \midrule
    Grey Cloth & 100.0   & 99.33 & 90.00    & 73.44 & 40.04 \\
    Grid Cloth & 94.38 & 98.72 & 95.09 & 92.00   & 51.85 \\
    Pink Flower & 100.0   & 99.96 & 95.67 & 100.0   & 74.90 \\
    Yellow Cloth & 100.0   & 99.56 & 98.26 & 98.80  & 54.98 \\
    \midrule
    \textbf{AVG} & 98.60  & 99.39 & 94.76 & 91.06 & 55.44 \\
    \bottomrule
    \end{tabular}%
    }
  \label{tab:metrics27}%
\end{table}%

\begin{table}[htbp]
  \centering
  \renewcommand\arraystretch{0.6}
  \caption{Anomaly detection and classification results of \textbf{UniADC{\fontsize{7}{7}\selectfont \texttt{(DINO)}}} on the WFDD dataset under the few-shot $(K_n=1, K_a=1)$ setting.}
     \resizebox{0.8\linewidth}{!}{
    \begin{tabular}{c|ccccc}
    \toprule
    \multicolumn{1}{c|}{Category} & \multicolumn{1}{c}{I-AUC} & \multicolumn{1}{c}{P-AUC} & \multicolumn{1}{c}{PRO} & \multicolumn{1}{c}{Acc} & \multicolumn{1}{c}{mIoU} \\
    \midrule
    Grey Cloth & 99.67 & 98.74 & 89.51 & 92.19 & 42.52 \\
    Grid Cloth & 99.72 & 99.41 & 96.11 & 94.00    & 45.38 \\
    Pink Flower & 100.0   & 99.98 & 95.19 & 100.0   & 68.18 \\
    Yellow Cloth & 99.99 & 99.35 & 95.48 & 98.20  & 46.03 \\
    \midrule
    \textbf{AVG} & 99.85 & 99.37 & 94.07 & 96.10  & 50.53 \\
    \bottomrule
    \end{tabular}%
    }
  \label{tab:metrics28}%
\end{table}%

\begin{table}[htbp]
  \centering
  \renewcommand\arraystretch{0.6}
  \caption{Anomaly detection and classification results of \textbf{UniADC{\fontsize{7}{7}\selectfont \texttt{(DINO)}}} on the WFDD dataset under the few-shot $(K_n=2, K_a=1)$ setting.}
     \resizebox{0.8\linewidth}{!}{
    \begin{tabular}{c|ccccc}
    \toprule
    \multicolumn{1}{c|}{Category} & \multicolumn{1}{c}{I-AUC} & \multicolumn{1}{c}{P-AUC} & \multicolumn{1}{c}{PRO} & \multicolumn{1}{c}{Acc} & \multicolumn{1}{c}{mIoU} \\
    \midrule
    Grey Cloth & 99.78 & 99.03 & 90.87 & 96.88 & 42.32 \\
    Grid Cloth & 99.72 & 99.45 & 97.17 & 95.00    & 49.43 \\
    Pink Flower & 100.0   & 99.97 & 92.74 & 98.21 & 68.35 \\
    Yellow Cloth & 99.97 & 99.46 & 95.68 & 98.80  & 47.02 \\
    \midrule
    \textbf{AVG} & 99.87 & 99.48 & 94.12 & 97.22 & 51.78 \\
    \bottomrule
    \end{tabular}%
    }
  \label{tab:metrics29}%
\end{table}%

\begin{table}[htbp]
  \centering
  \renewcommand\arraystretch{0.6}
  \caption{Anomaly detection and classification results of \textbf{UniADC{\fontsize{7}{7}\selectfont \texttt{(DINO)}}} on the WFDD dataset under the few-shot $(K_n=2, K_a=2)$ setting.}
     \resizebox{0.8\linewidth}{!}{
    \begin{tabular}{c|ccccc}
    \toprule
    \multicolumn{1}{c|}{Category} & \multicolumn{1}{c}{I-AUC} & \multicolumn{1}{c}{P-AUC} & \multicolumn{1}{c}{PRO} & \multicolumn{1}{c}{Acc} & \multicolumn{1}{c}{mIoU} \\
    \midrule
    Grey Cloth & 99.89 & 99.00    & 94.57 & 98.44 & 46.38 \\
    Grid Cloth & 99.84 & 99.44 & 96.82 & 95.00    & 46.76 \\
    Pink Flower & 100.0   & 99.98 & 91.60  & 100.0   & 67.73 \\
    Yellow Cloth & 100.0   & 99.54 & 93.77 & 97.60  & 46.78 \\
    \midrule
    \textbf{AVG} & 99.93 & 99.49 & 94.19 & 97.76 & 51.91 \\
    \bottomrule
    \end{tabular}%
    }
  \label{tab:metrics30}%
\end{table}%

\begin{table}[htbp]
  \centering
  \renewcommand\arraystretch{0.6}
  \caption{Anomaly detection and classification results of \textbf{UniADC{\fontsize{7}{7}\selectfont \texttt{(DINO)}}} on the WFDD dataset under the few-shot $(K_n=4, K_a=1)$ setting.}
     \resizebox{0.8\linewidth}{!}{
    \begin{tabular}{c|ccccc}
    \toprule
    \multicolumn{1}{c|}{Category} & \multicolumn{1}{c}{I-AUC} & \multicolumn{1}{c}{P-AUC} & \multicolumn{1}{c}{PRO} & \multicolumn{1}{c}{Acc} & \multicolumn{1}{c}{mIoU} \\
    \midrule
    Grey Cloth & 99.78 & 98.91 & 94.05 & 96.88 & 43.22 \\
    Grid Cloth & 99.68 & 99.53 & 94.24 & 95.00   & 46.77 \\
    Pink Flower & 100.0   & 99.97 & 94.17 & 100.0   & 69.40 \\
    Yellow Cloth & 100.0   & 99.38 & 95.88 & 98.80  & 47.79 \\
    \midrule
    \textbf{AVG} & 99.87 & 99.45 & 94.59 & 97.67 & 51.80 \\
    \bottomrule
    \end{tabular}%
    }
  \label{tab:metrics31}%
\end{table}%

\begin{table}[htbp]
  \centering
  \renewcommand\arraystretch{0.6}
  \caption{Anomaly detection and classification results of \textbf{UniADC{\fontsize{7}{7}\selectfont \texttt{(DINO)}}} on the WFDD dataset with full-shot normal samples and $(K_a=1)$.}
     \resizebox{0.8\linewidth}{!}{
    \begin{tabular}{c|ccccc}
    \toprule
    \multicolumn{1}{c|}{Category} & \multicolumn{1}{c}{I-AUC} & \multicolumn{1}{c}{P-AUC} & \multicolumn{1}{c}{PRO} & \multicolumn{1}{c}{Acc} & \multicolumn{1}{c}{mIoU} \\
    \midrule
    Grey Cloth & 100.0   & 98.82 & 95.95 & 97.14 & 40.78 \\
    Grid Cloth & 99.96 & 99.51 & 94.90  & 95.00    & 48.75 \\
    Pink Flower & 100.0  & 99.97 & 95.92 & 100.0 & 70.59 \\
    Yellow Cloth & 99.97 & 99.44 & 94.58 & 98.80  & 49.85 \\
    \midrule
    \textbf{AVG} & 99.98 & 99.44 & 95.34 & 97.74 & 52.49 \\
    \bottomrule
    \end{tabular}%
    }
  \label{tab:metrics32}%
\end{table}%

\begin{table}[htbp]
  \centering
  \renewcommand\arraystretch{0.75}
  \caption{Anomaly detection and classification results of \textbf{UniADC{\fontsize{7}{7}\selectfont \texttt{(CLIP)}}} on the Real-IAD dataset under the few-shot $(K_n=2, K_a=1)$ setting.}
  \resizebox{0.8\linewidth}{!}{
    \begin{tabular}{c|ccccc}
    \toprule
    Category & \multicolumn{1}{c}{I-AUC} & \multicolumn{1}{c}{P-AUC} & \multicolumn{1}{c}{PRO} & \multicolumn{1}{c}{Acc} & \multicolumn{1}{c}{mIoU} \\
    \midrule
    Audiojack & 76.83 & 97.84 & 88.42 & 68.74 & 23.61 \\
    Bottle Cap & 90.87 & 98.75 & 87.82 & 78.36 & 31.33 \\
    Button Battery & 91.22 & 98.79 & 90.65 & 63.42 & 31.40 \\
    End Cap & 77.20  & 96.97 & 87.25 & 51.19 & 25.27 \\
    Eraser & 92.23 & 99.59 & 92.55 & 81.57 & 28.56 \\
    Fire Hood & 95.94 & 98.69 & 94.97 & 89.10  & 41.41 \\
    Mint  & 75.01 & 95.41 & 90.26 & 48.08 & 30.55 \\
    Mounts & 89.13 & 97.87 & 85.52 & 75.68 & 40.49 \\
    Pcb   & 68.35 & 90.64 & 84.51 & 39.97 & 21.47 \\
    Phone Battery & 87.44 & 98.99 & 87.14 & 62.31 & 28.82 \\
    Plastic Nut & 81.26 & 97.40  & 85.83 & 81.79 & 24.85 \\
    Plastic Plug & 90.04 & 98.59 & 89.91 & 76.67 & 36.16 \\
    Porcelain Doll & 93.48 & 98.89 & 92.85 & 80.60  & 35.24 \\
    Regulator & 88.30  & 98.94 & 92.34 & 89.50  & 39.77 \\
    Rolled Strip Base & 95.08 & 98.90  & 95.92 & 68.77 & 43.14 \\
    Sim Card Set & 97.93 & 99.50  & 96.54 & 67.90  & 39.44 \\
    Switch & 80.04 & 91.69 & 84.69 & 51.95 & 34.88 \\
    Tape  & 97.83 & 99.75 & 94.47 & 84.08 & 39.51 \\
    Terminalblock & 92.58 & 98.89 & 85.86 & 68.90  & 43.22 \\
    Toothbrush & 92.02 & 98.65 & 89.36 & 75.69 & 41.38 \\
    Toy   & 84.07 & 88.15 & 85.45 & 50.53 & 29.39 \\
    Toy Brick & 88.22 & 97.33 & 90.35 & 82.09 & 44.96 \\
    Transistor1 & 87.16 & 94.42 & 87.70  & 58.04 & 36.92 \\
    U Block & 82.76 & 99.47 & 87.33 & 80.96 & 29.58 \\
    Usb   & 85.75 & 97.49 & 86.00   & 69.51 & 30.41 \\
    Usb Adaptor & 81.54 & 96.15 & 90.55 & 61.31 & 26.74 \\
    Vcpill & 87.75 & 99.30  & 86.61 & 83.17 & 48.09 \\
    Wooden Beads & 89.24 & 96.79 & 94.28 & 64.08 & 32.27 \\
    Woodstick & 92.88 & 98.71 & 96.84 & 84.23 & 35.78 \\
    Zipper & 95.58 & 99.14 & 89.29 & 68.93 & 40.13 \\
    \midrule
    \textbf{AVG} & 87.59 & 97.39 & 89.71 & 70.24 & 34.49 \\
    \bottomrule
    \end{tabular}%
    }
  \label{tab:metrics33}%
\end{table}%

\begin{table}[htbp]
  \centering
  \renewcommand\arraystretch{0.75}
  \caption{Anomaly detection and classification results of \textbf{UniADC{\fontsize{7}{7}\selectfont \texttt{(DINO)}}} on the Real-IAD dataset under the few-shot $(K_n=2, K_a=1)$ setting.}
  \resizebox{0.8\linewidth}{!}{
    \begin{tabular}{c|ccccc}
    \toprule
    Category & \multicolumn{1}{c}{I-AUC} & \multicolumn{1}{c}{P-AUC} & \multicolumn{1}{c}{PRO} & \multicolumn{1}{c}{Acc} & \multicolumn{1}{c}{mIoU} \\
    \midrule
       Audiojack & 76.35 & 98.62 & 90.16 & 73.65 & 28.35 \\
    Bottle Cap & 95.50  & 99.40  & 92.23 & 80.41 & 34.86 \\
    Button Battery & 84.90  & 99.06 & 92.33 & 64.12 & 36.65 \\
    End Cap & 80.47 & 97.69 & 89.44 & 47.69 & 30.60 \\
    Eraser & 92.50  & 99.71 & 92.63 & 82.37 & 32.25 \\
    Fire Hood & 97.47 & 99.63 & 97.60  & 89.95 & 42.51 \\
    Mint  & 77.33 & 97.08 & 92.26 & 51.13 & 32.67 \\
    Mounts & 89.37 & 99.76 & 88.69 & 74.88 & 37.29 \\
    Pcb   & 70.57 & 95.17 & 85.96 & 40.66 & 28.46 \\
    Phone Battery & 91.79 & 99.59 & 85.71 & 73.54 & 34.83 \\
    Plastic Nut & 90.34 & 99.37 & 87.70  & 83.04 & 28.23 \\
    Plastic Plug & 88.38 & 98.99 & 92.41 & 81.11 & 35.08 \\
    Porcelain Doll & 92.94 & 99.03 & 91.85 & 84.78 & 38.74 \\
    Regulator & 91.83 & 99.82 & 94.64 & 81.16 & 46.50 \\
    Rolled Strip Base & 95.71 & 99.34 & 93.08 & 69.83 & 46.77 \\
    Sim Card Set & 96.21 & 97.77 & 89.29 & 57.35 & 30.67 \\
    Switch & 81.21 & 92.62 & 85.07 & 48.99 & 34.72 \\
    Tape  & 97.61 & 99.68 & 91.05 & 76.85 & 37.30 \\
    Terminalblock & 94.29 & 98.94 & 88.22 & 65.05 & 43.00 \\
    Toothbrush & 91.87 & 97.66 & 91.42 & 67.72 & 39.49 \\
    Toy   & 87.28 & 88.69 & 85.30  & 60.34 & 30.49 \\
    Toy Brick & 89.36 & 99.29 & 92.96 & 83.52 & 46.33 \\
    Transistor1 & 92.52 & 98.25 & 93.79 & 58.31 & 34.09 \\
    U Block & 85.24 & 99.75 & 89.24 & 82.74 & 31.76 \\
    Usb   & 86.10  & 98.63 & 88.98 & 69.21 & 31.62 \\
    Usb Adaptor & 87.77 & 98.49 & 89.24 & 67.71 & 27.02 \\
    Vcpill & 90.34 & 99.51 & 90.17 & 78.55 & 42.21 \\
    Wooden Beads & 87.79 & 98.70 & 86.94 & 55.03 & 30.63 \\
    Woodstick & 92.01 & 98.91 & 90.09 & 87.10  & 38.10 \\
    Zipper & 99.92 & 99.64 & 91.62 & 76.27 & 35.98 \\
    \midrule
    \textbf{AVG} & 89.17 & 98.29 & 90.34 & 70.44 & 35.57 \\
    \bottomrule
    \end{tabular}%
    }
  \label{tab:metrics34}%
\end{table}%

\end{document}